\documentclass[runningheads]{llncs}
\pdfoutput=1
 
\usepackage{eccv}

\usepackage{eccvabbrv}

\usepackage{graphicx}
\usepackage{booktabs}
\usepackage{amsmath}
\usepackage{amssymb}
\usepackage{booktabs}
\usepackage{verbatim}
\usepackage[dvipsnames]{xcolor}
\usepackage{multirow}
\usepackage{makecell}
\usepackage[font=small,skip=3pt]{caption}
\usepackage[accsupp]{axessibility}  

\usepackage[pagebackref,breaklinks,colorlinks]{hyperref}

\usepackage{orcidlink}

\begin{document}

\title{Always Clear Days: Degradation Type and Severity Aware All-In-One Adverse Weather Removal} 

\titlerunning{Degradation Type and Severity Aware All-In-One Adverse Weather Removal}

\author{Yu-Wei Chen\inst{1}\orcidlink{0000-0001-9127-6536} \and
Soo-Chang Pei\inst{1,2}\orcidlink{0000-0003-2448-4196}}

\authorrunning{Chen and Pei.}

\institute{Graduate Institute of Communication Engineering, National Taiwan University \and Department of Electrical Engineering, National Taiwan University\\
\email{\{r09942066, peisc\}@ntu.edu.tw}}

\maketitle

\begin{abstract}
  All-in-one adverse weather removal is an emerging topic on image restoration, which aims to restore multiple weather degradations in an unified model, and the challenge are twofold. First, discover and handle the property of multi-domain in target distribution formed by multiple weather conditions. Second, design efficient and effective operations for different degradations. To resolve this problem, most prior works focus on the multi-domain caused by different weather types. Inspired by inter\&intra-domain adaptation literature, we observe that not only weather type but also weather severity introduce multi-domain within each weather type domain, which is ignored by previous methods, and further limit their performance. To this end, we propose a degradation type and severity aware model, called \textbf{UtilityIR}, for blind all-in-one bad weather image restoration. To extract weather information from single image, we propose a novel Marginal Quality Ranking Loss (MQRL) and utilize Contrastive Loss (CL) to guide weather severity and type extraction, and leverage a bag of novel techniques such as Multi-Head Cross Attention (MHCA) and Local-Global Adaptive Instance Normalization (LG-AdaIN) to efficiently restore spatial varying weather degradation. The proposed method can outperform the state-of-the-art methods subjectively and objectively on different weather removal tasks with a large margin, and enjoy less model parameters. Proposed method even can restore \textbf{unseen} combined multiple degradation images, and modulate restoration level. Implementation code and pre-trained weights will be available at \url{https://github.com/fordevoted/UtilityIR}
\end{abstract}
\vspace{-0.4cm}
\section{Introduction}
\label{sec:intro}
\begin{figure}[t!]
  \centering
    \includegraphics[width=0.5\linewidth]{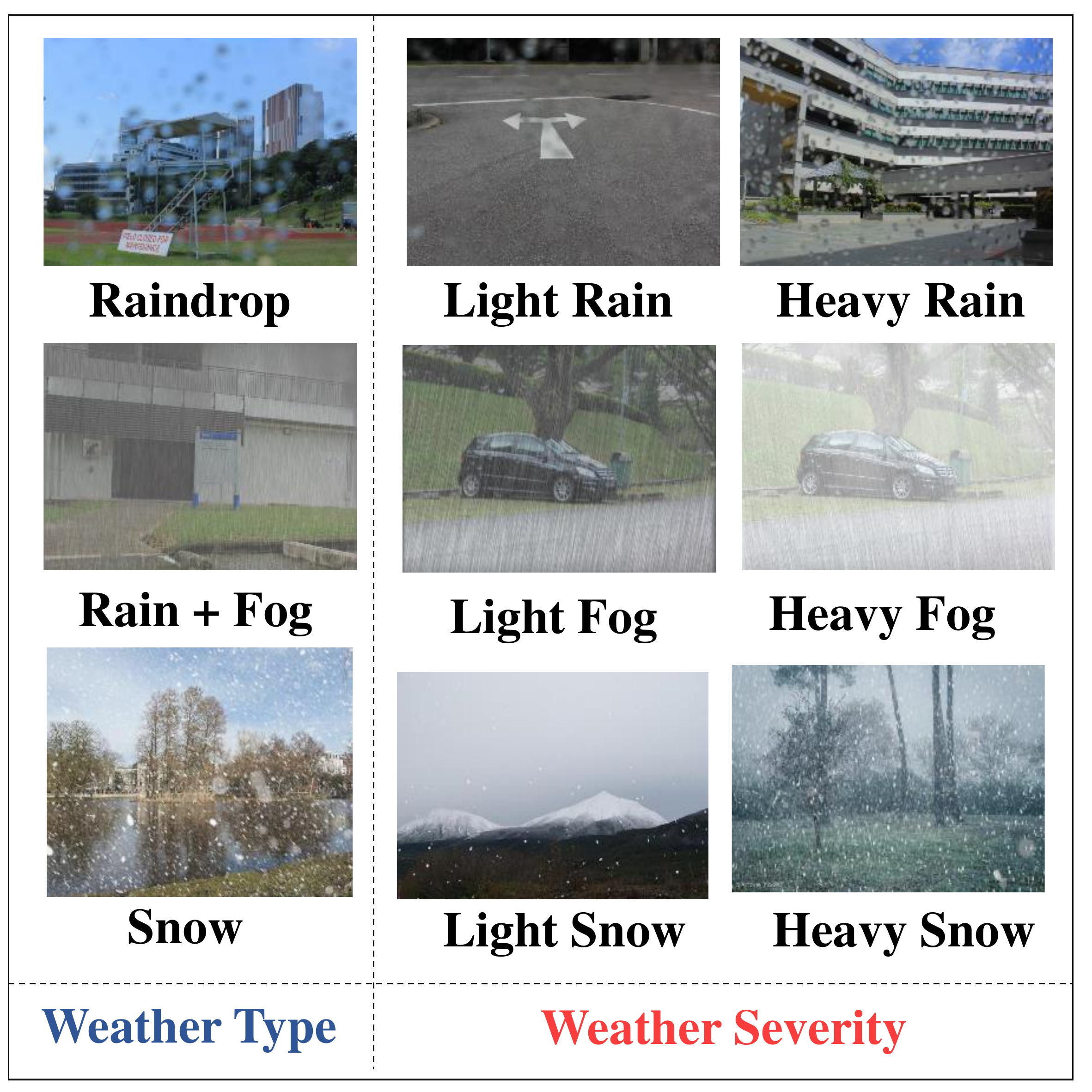}
   \caption{Illustration of challenge of handling various weather degraded images. Not only weather type result in multi-domain and different degradation appearance, weather severity also cause diverse visual appearance and inra-domain gap, which is ignored by previous all-in-one bad weather removal works.}
   \label{fig:intro_type_severity}
\end{figure}
Despite the fact that learning based methods usher in a dramatic growth and success on computer vision such as image classification and segmentation in last decade, the performance of many high-level vision algorithms usually degrade while applied in real-world adverse environments, e.g. underwater, adverse illumination, and bad weather conditions. To restore degraded images, most of previous works design task specific model for each adverse environment such as underwater \cite{uiess, dauie}, low-light \cite{dln}, rain \cite{prenet, didmdn, jorder, twostep_derain, deraindrop, durn, heavyrain}, haze \cite{zeroshotdehaze, twostep_dehaze, aecrnet, dcp, aodnet, dehazenet, hardgan}, and snow \cite{HDCWnet, JSTASR, ddmsnet, desnownet}. However, deploying multiple models for different scenario is resource in-efficiency for practical usage. All-in-one bad weather removal \cite{all-in-one, transweather, unified, weatherdiff, wgwsnet, gridformer} is thus emerging topic to resolve this issue, which aims to learn an unified model to restore image degraded by different adverse weather conditions. The challenge of all-in-one adverse weather removal (or broadly, all-in-one image restoration) are twofold:
\begin{enumerate}
    \item Effectively and efficiently integrate different operations suitable for different weather degradations.
    \item The target distribution formed by different weather conditions is with the property of multi-domain, naively learn the mapping from all degraded images to clean images would result in learning from a large variance uni-domain distribution, and lead to sub-optimal performance. Accurately and implicitly handle the property of multi-domain at test-time would be challenging.
\end{enumerate}

To confront the first challenge, previous works adopt various techniques for effective and efficient architecture, e.g. NAS \cite{all-in-one}, ViT \cite{transweather, gridformer}, knowledge distillation \cite{unified}, deformable convolution and feature affine \cite{airNet}, FAIG \cite{adms}, weather general\&specific operation \cite{wgwsnet}, and diffusion model \cite{weatherdiff}, etc. As the second challenge, some prior works \cite{unified, airNet} utilize contrastive learning \cite{cl} to separate different weather type feature and learn the multi-domain of target distribution, another techniques such as classifier \cite{adms}, multiple weather specific operation/encoder \cite{wgwsnet, all-in-one}, and learnable query \cite{transweather} are also adopted. It is worthy to note that in some works \cite{all-in-one, wgwsnet}, the weather type label is available while testing, namely \textit{non-blind} all-in-one image restoration. A more challenging setting would be contrary situation \cite{transweather, unified, adms, airNet} that input image is unknown degradation while inferencing, which is \textit{blind} all-in-one image restoration, and we focus on the latter one in this paper. 

Inspired by inter\&intra-domain adaptation literature for image enhancement \cite{dauie}, we view the problem from data domain perspective and observe that multi-domain obstacles are not only exist between different weather types, the diverse weather severity also introduce multi-domain in intra-domain of weather type, which is ignored by most of previous works, and further limit their performance. As shown in Fig. \ref{fig:intro_type_severity}, not only weather type result in different degradation, the diverse weather severity also cause various appearance and lead to intra-domain gap.

To this end, we propose a degradation type and severity aware all-in-one adverse weather removal network based on the image quality ranker with proposed Marginal Quality Ranking Loss (MQRL) and a bag of techniques to effectively and efficiently restore diverse degraded images. Specifically, to extract type and severity information, we design a Degradation Information Encoder (DIE) to perform two branches multi-task feature extraction. For the weather type branch, we impose Contrastive Loss (CL) to diminish the distance of feature extracted by same weather type, and enlarge the distance between different weather types. As severity, we develop based on the intuitive observation that the more severely degraded image produce worse Image Quality Assessment (IQA) score, and IQA score is positively related to restoration level that required to apply. Motivated by previous works \cite{dauie, rankersrgan} that demonstrate the effectiveness of learning an image quality ranker with Marginal Ranking Loss (MRL) to benefit the following image restoration, we train a ranker to predict weather severity. Nevertheless, the standard MRL only consider the ranking information between input image pairs, that make the ranker prone to predict incorrect IQA score interval, and lead to apply inappropriate restoration level while use the predicted IQA score as restoration level signal. To be the remedy, we further proposed an interval-aware MRL, i.e. MQRL, to better extract the severity information. After obtaining the weather information, they will be injected into model through Degradation Information Local-Global AdaIN (DI-LGAdaIN) and Degradation-guided Cross Attention (DGCA) for type and severity aware global-local degradation removal. The proposed method can outperform the state-of-the-art methods on real-world and synthetic dataset \cite{all-in-one, transweather} subjectively and objectively, and can restore the combined multiple weather degradation \textbf{without} training these type data, but enjoy less parameters compared with other blind all-in-one methods. 
Our contributions are summarized as follow:
\begin{enumerate}
    \item To the best of our knowledge, UtilityIR is the first type and severity aware blind all-in-one image restoration method with single image input, that can modulate restoration level.
    \item We proposed MQRL to address the insufficiency of standard MRL in scenario of degradation severity estimation. We also modify and exploit a bag of off-the-shelf techniques to better estimate weather information and restore diverse degradation type images.
    \item Proposed method significantly outperform the SOTA methods on different weather removal tasks subjectively and objectively with less parameters, and even can restore unseen combined multiple weather images.
\end{enumerate}

\begin{figure*}[t]
  \centering
  \subfloat[]{
  \includegraphics[width=0.9\textwidth]{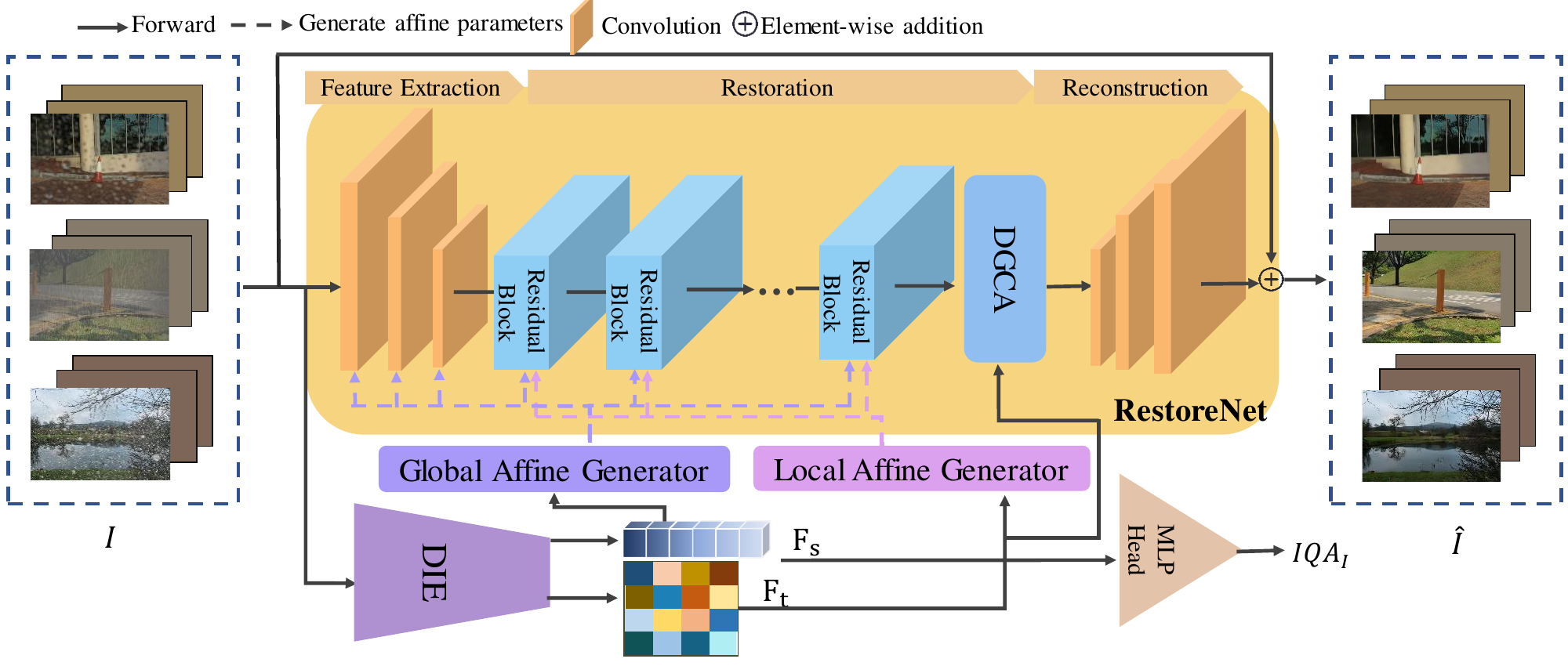}
  }
  
  \subfloat[]{
  \includegraphics[width=0.3\linewidth]{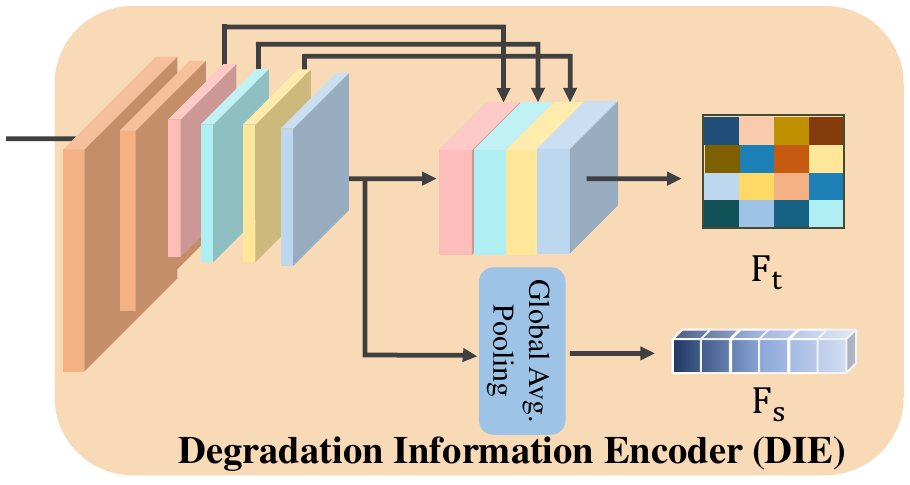}
  }
  \subfloat[]{
  \includegraphics[width=0.25\linewidth]{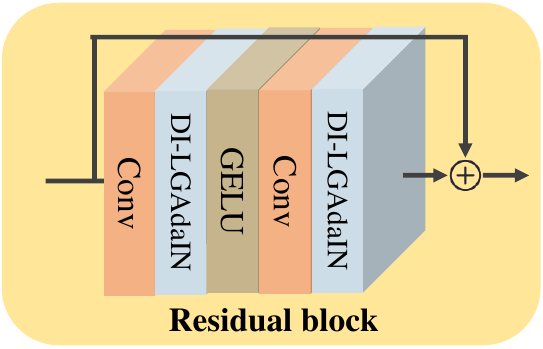}
  }
  \subfloat[]{
  \includegraphics[width=0.4\linewidth]{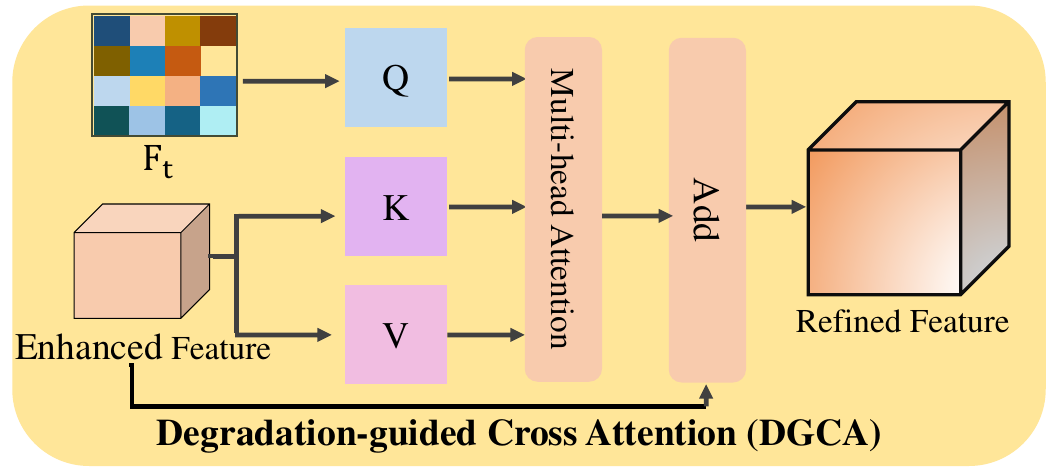}
  }
   \caption{(a) Overview of proposed UtilityIR. Input arbitrary weather image, Degradation Information Encoder (DIE) shown in (b) will first be applied to extract weather type and severity, then inject these information through Degradation Information Local-Global AdaIN (DI-LGAdaIN) in residual block shown in (c) and Degradation-guided Cross Attention (DGCA) shown in (d).}
   \label{fig:method_model}
   \vspace{-0.2cm}
\end{figure*}

\vspace{-0.3cm}
\section{Related works}
\paragraph{Single Image Restoration for Adverse Weather}
  To restore adverse weather images, many learning based methods are designed by weather specific characteristics or physical model. For image deraining \cite{durn, prenet, jorder, didmdn, heavyrain, deraindrop, twostep_derain}, JORDER \cite{jorder} is based on additive rain model to jointly learn the rain streak detection and removal. Follow the formulation of raindrop image, Qian \etal~\cite{deraindrop} remove raindrop through learning raindrop mask attention. For image dehazing \cite{dehazenet, zeroshotdehaze, twostep_dehaze, aecrnet, aodnet, hardgan}, DehazeNet\cite{dehazenet} learn to estimate transmission map. ZID \cite{zeroshotdehaze} leverage atmosphere scatter model to learn a dehazing model without ground-truth. To remove snow from single image \cite{HDCWnet, JSTASR, ddmsnet, desnownet}, Chen \etal~\cite{HDCWnet} proposed HDCWNet based on dual-tree complex wavelet transform. Another trend of research lie in task-agnostic model for general image restoration \cite{swinir, restormer, mprnet, maxim, nafnet, grl, mirnet}. SwinIR \cite{swinir} based on Swin Transformer, which combine the advantages of the local processing of convolution and the long-range dependency modeling of Transformer. MAXIM \cite{maxim} use multi-axis gated design based on MLP with local and global branches. Nevertheless, these methods can only restore one type of degradation once for single model and weights, which limit the practical usage in the real-world.

\paragraph{All-in-one Image Restoration}
All-in-one image restoration is a new trend on image restoration in recent years, which aims to remove different degradation types with an unified model. For non-blind all-in-one image restoration \cite{all-in-one, wgwsnet, ipt}, All-in-one network \cite{all-in-one} first proposed task-specific encoders and common decoder based on NAS. WGWS-Net \cite{wgwsnet} observed that model trained by different weather types exist general and specific filters, and further design a weather general and specific network. To achieve blind all-in-one image restoration \cite{transweather, unified, airNet, adms, weatherdiff, idr, vrdir},  Transweather \cite{transweather} based on transformer and learnable weather query. Unified model \cite{unified} and AirNet \cite{airNet} apply CL to separate different degradation type features. Zhang \etal~\cite{idr} proposed an ingredient-oriented degradation reformulation framework that can integrate to any Transformer based model. ADMS \cite{adms} utilize FAIG to obtain task specific discriminative filter and synergize with a degradation classifier. Weatherdiff \cite{weatherdiff} remove multiple weather degradation based on diffusion model. However, above approaches only consider the multi-domain caused by weather type, but not explicitly handle weather severity, which also produce different difficulty for image restoration, and limit their performance.

\paragraph{Severity Aware Image Restoration}
Apart from degradation type, degradation severity also bring various appearance and difficulty to image restoration, many prior works \cite{didmdn,  dauie, twostep_derain, JSTASR, QGCN, cbdnet} have been demonstrated the effectiveness of taking degradation severity into account for image restoration. DID-MDN \cite{didmdn} proposed a rain density aware model and divided rain level into light, medium, and heavy, a similar work is also done by \cite{twostep_derain}. JSTASR \cite{JSTASR} estimates snow size and transparency for snow removal. Yi \etal~\cite{twostep_dehaze} utilize a domain adaptation framework to mitigate the Syn2real and intra-domain gap due to haze severity for image dehazing. Li \etal~\cite{QGCN} proposed a quantization-aware JPEG artifact removal. CBDNet \cite{cbdnet} estimate noise level and jointly feed noise level and noisy image into model for image denosing. In this work, we aim to deal with the diverse degradation appearance caused by different degradation type and severity for adverse weather removal in a more general form.

\section{Proposed Method}
\label{sec:method}
\subsection{Overview}
The overview of proposed UtilityIR is depicted in Fig. \ref{fig:method_model}. Input arbitrary weather image $I \in \mathbb{R}^{H\times W \times C}$, the Degradation Information Encoder (DIE) will be applied to extract weather information, include type $F_t \in \mathbb{R}^{\frac{H}{S} \times \frac{W}{S}}$ and severity $F_s \in \mathbb{R}^{D}$, where $S$ is the down sampling ratio and $D$ is the feature dimension. The process can be presented by:
\begin{equation}
    F_t, F_s = DIE(I)
\end{equation}
Note that $F_t$ will keep as 2D matrix and $F_s$ will be encoded into a 1D vector. The intuition behind our design is that different weather types give rise to spatial varying degradation, e.g. raindrop usually bring on local degradation, and fog bring on global degradation, which make keep spatial information being crucial while extract representative weather type feature. On the other hand, using scalar or vector to represent degradation severity is enough, i.e. IQA score or noise level, we adopt the encoded feature vector before IQA score regressor as the severity rather than predicted IQA score for richer information. 

\begin{figure}[t]
  \centering
  \includegraphics[width=0.75\linewidth]{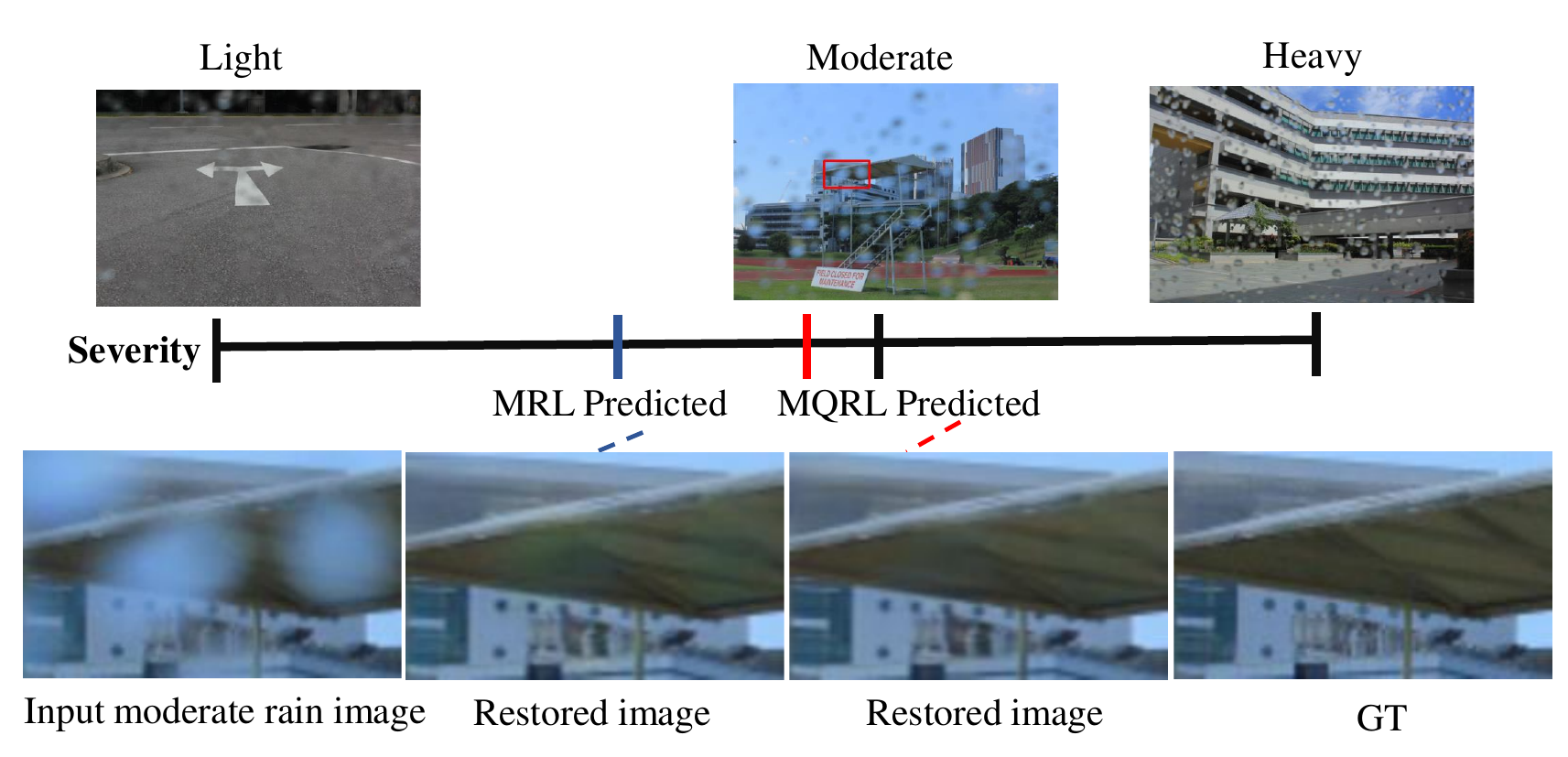}
   \caption{Illustration of insufficient of guided by standard MRL. The incorrect IQA interval prediction lead to apply inappropriate restoration level. Both cases achieve 0 in MRL.}
   \label{fig:MRL_cons}
   \vspace{-0.4cm}
\end{figure}

Based on the intuition of severe weather image result in worse IQA score, we follow prior works \cite{rankersrgan, dauie} to learn a image quality ranker to boost following image restoration. Different from RankSRGAN \cite{rankersrgan} utilize ranker to provide novel rank-content loss, and TUDA \cite{dauie} leverage ranker to initial model weights, our ranker is adopted to directly extract degradation severity for restoration level signal. Note that the ranker is usually guided by MRL \cite{stdrkl} in \cite{rankersrgan, dauie} rather than directly regress GT IQA score, since what we care is not exact IQA value but the ranking information\cite{rankersrgan}. However, the ranker guided by standard MRL do not consider the interval of predicting IQA score but only ranking information, which might lead to under-/over-restored while regard the severity feature as the restoration level signal to inject into restoration module. As the example illustrates in Fig. \ref{fig:MRL_cons}, the severity of moderate rainy image is actually closer to the heavy rain, but the severity predicted by MRL ranker is closer to the light one, that make model adopt insufficient restoration level and result in under-restored.

To tackle this issue, we further propose Marginal Quality Ranking Loss (MQRL) to guide weather severity extraction. We first introduce $diff_{in} = IQA_I -IQA_{I'}$ is predicted IQA score difference of degraded image pair, $diff_{gt} = \Phi(I, I_{gt}) - \Phi(I', I'_{gt})$ is GT IQA score difference of the image pair, and $diff = |diff_{gt} - diff_{in}|$ is the distance between $diff_{in}$ and $diff_{gt}$, where $I$ is input degraded image, $I'$ is random sample image with same weather type to $I$, $I_{gt}, I'_{gt}$ are corresponding GT images for $I$ and $I'$, respectively, and $\Phi(\cdot)$ is any IQA metric, we adopt PSNR for simplicity although others perceptual score can be adopted. Then, the MQRL can be formulated by:
\begin{equation}
    \begin{aligned}
        \begin{split}
        &\mathcal{L}_{mqrl}(IQA_I, IQA_{I'}, I, I', I_{gt}, I'_{gt}) = \\ & \begin{cases}
        diff& , \text{$if sgn(diff_{gt}) \neq sgn(diff_{en})$} \\ 
        max(0, diff - \epsilon)&, \text{$else $}
    \end{cases}
    \end{split}
    \end{aligned}
    \label{eq: mqrl}
\end{equation}

where $sgn(\cdot)$ is sign function, and $\epsilon$ is the margin. The aim behind formula is learning to rank weather severity first, then ease the burden while predicted severity is correctly ranked and the difference interval is closed enough to GT. Compared to directly regress IQA score, MQRL ease the burden of regressing the bias of GT distribution, which is trivial for severity feature; Compared to MRL, MQRL learning more about variance of GT distribution, which is more crucial for learning restoration level signal.
As for meeting the representative weather type feature, we follow to prior works \cite{unified, airNet} to impose CL \cite{cl} to improve discriminative of feature extracted from different weather types. After obtaining representative weather information, they will jointly input into model with input image $I$ to guide the restoration process, and obtain the high-quality restored result $\hat{I}$:
\begin{equation}
    \hat{I} = R(I, F_t, F_s)
\end{equation}
where $R$ is the restoration network. We will detail the architecture and how to inject weather information below. 

\subsection{Network Architecture}
\subsubsection{Degradation Information Encoder (DIE)}
As shown in Fig. \ref{fig:method_model} (b) , DIE consists with stack of convolution layers to extract different level features. We generate type and severity feature under multitasking manner using two branches architecture, and guided by different objective function. 

For weather type branch, the features extracted from different layers will be concatenated, then digest by another convolution layer to encode from low-level to high-level features for diverse degradation type. As for the weather severity branch, we perform global average pooling on input feature to obtain quality feature vector and regress IQA score by a simple two-layers MLP. Similar to mapping network in \cite{cvprw2022} and CBDE in \cite{airNet}, our DIE can be integrated to any task-specific image restoration backbone network, and extend to all-in-one image restoration fashion.

\subsubsection{RestoreNet}
The RestoreNet consist three stages, i.e. feature extraction, restoration, and reconstruction. We stack convolution layers with stride 2 as feature extraction block to extract degraded feature $F \in \mathbb{R}^{\frac{H}{S} \times \frac{W}{S} \times D}$, where $S$ is down sampling ratio, and $D$ is feature dimension.

For restoration stage, we stack $K$ residual blocks and a DGCA block to improve ability of modeling long-range dependency from convolution net. As shown in Fig. \ref{fig:method_model} (c), residual block is formed by classic conv-norm-activ block. Thanks to the success of AdaIN \cite{adain} and feature affine transform in image restoration literature \cite{adaFM, airNet, uiess, hardgan, psfrgan}, we inject the information into model through proposed Degradation Information Local-Global Adaptive Instance Normalization (DI-LGAdaIN), which is inspired by LG-AdaIN \cite{lgAdaIN}. The DI-LGAdaIN consist a local and global feature affine transform and an instance normalization. Considering the feature dimensionality and consisting spatial information, we use weather type to generate local, i.e. pixel-wise, affine transform parameters. Follow to \cite{adaFM}, which shows different restoration level can be modulating through channel-wise feature linear operation, the weather severity is used to perform global, i.e. channel-wise, affine transform, which is equivalent to kernel size set to $1\times1$ in \cite{adaFM}. The DI-LGAdaIN can be formulated as:
\begin{equation}
    F' = \alpha_g(\alpha_l\frac{F-\mu}{\sigma} +\beta_l)+ \beta_g
\end{equation}
where $\alpha_l, \beta_l \in \mathbb{R}^{\frac{H}{S} \times \frac{W}{S}}$ are local affine parameters, $\alpha_g, \beta_g \in \mathbb{R}^{D}$ are global affine parameters, and $\mu$ and $\sigma$ are mean and standard deviation of $F$. Then we consider the Multi-Head Self-Attention (MHSA) at the end of residual blocks to introduce global receptive field operation and refine the enhanced feature. However, MHSA tends to globally attend content of input feature \cite{restoreformer}, which might not be suitable for local degradation, considering degradation appearance mainly depend on weather type, we further utilize Multi-Head Cross-Attention (MHCA) \cite{restoreformer}, and take weather type as query to guide the self-attention feature refinement as \cite{transweather}. The DGCA is depicted in Fig. \ref{fig:method_model} (d). Finally, the restored feature will transform back to image domain through reconstruction module.

\subsection{Training Loss}
\label{sec:method_loss}
To extract weather information, we use proposed MQRL in Eq.\ref{eq: mqrl} to guide learning the weather severity, and CL to guide weather type extraction. The CL can be formulated as:
\begin{equation}
    \begin{split}
        & \mathcal{L}_{cl}(F_t, F^+_t, F^-_t) = \\ & -log[\frac{exp(\frac{\Psi(F_t, F^+_t)}{\tau})}{exp(\frac{\Psi(F_t, F_t^+)}{\tau})+ \Sigma^N_{j=1}  exp(\frac{\Psi(F_t, F^-_{t, j})}{\tau})}]\\ 
    \end{split}
\end{equation}
where $F_t$ is input weather type feature, $F^+_t$, $F^-_t$ are positive and negative feature examples, $\tau$ is temperature, and $\Psi(\cdot)$ is the cosine similarity function. For image restoration, we utilize L1 loss and SSIM loss as pixel fidelity loss, which can be formulated by:
\begin{equation}
    \mathcal{L}_{l1} = \| \hat{I} - Y||_1
\end{equation}
and
\begin{equation}
    \mathcal{L}_{ssim} = 1-SSIM(\hat{I}, Y)
\end{equation}
We also use perceptual loss to achieve more realistic result:
\begin{equation}
        \mathcal{L}_{per} = \mathop{\sum}\limits_{j} \frac{1}{c_j h_h w_j}\| \psi_j(\hat{I}) - \psi_j(Y)\|^2_2
\end{equation}
where $c_j h_h w_j$ is shape of $j^{th}$ feature map $\psi_j$ of pre-trained VGG-16 network. The total loss can be summarized as:
\begin{equation}
    \begin{split}
    &\mathcal{L} = \mathcal{L}_{mqrl} + \lambda_{cl}\mathcal{L}_{cl}  + \lambda_{l1}\mathcal{L}_{l1} + \\&
    \lambda_{ssim}\mathcal{L}_{ssim} + \lambda_{per}\mathcal{L}_{per}
    \end{split}
\end{equation}
where $\lambda$ are weightings to control the contribution of each loss term.

\section{Experiments}
We employ two dataset setting for experiments. For \textbf{Setting 1}, we align our experiment setting with \cite{all-in-one, transweather, weatherdiff} for fair comparison, that the model is trained and evaluated on All-Weather dataset \cite{transweather}, which contain three kind of weather datasets, i.e. Raindrop dataset \cite{deraindrop} for raindrop, Outdoor-Rain \cite{heavyrain} for rain+fog, and Snow100K-L \cite{desnownet} for snow. We select fourteen state-of-the-art methods for comparison, include four paradigms, i.e. single task specific (AttentiveGAN \cite{deraindrop} and DuRN \cite{durn} for raindrop. pix2pix \cite{pix2pix} and HRGAN \cite{heavyrain} for rain+fog. JSTASR \cite{JSTASR} and DDMSNet \cite{ddmsnet} for snow), single task agnostic (MPRNet \cite{mprnet}, Restormer \cite{restormer} and SwinIR \cite{swinir}), non-blind all-in-one (All-in-one network \cite{all-in-one} and WGWS-Net \cite{wgwsnet}), and blind all-in-one (Transweather \cite{transweather}, Unified model \cite{unified} and Weatherdiff \cite{weatherdiff}). As for \textbf{Setting 2}, we follow to previous works \cite{unified, wgwsnet} training model on Rain1400 \cite{rain1400} for rain streak, RESIDE \cite{reside} for haze, and Snow100K-L \cite{desnownet} for snow. We use pre-trained weights if it is available, otherwise, we use author's official code with default setting or minimal modification to overcome limitation of CUDA memory for training the models.

\subsection{Implementation Details}
We randomize crop the images with patch size 256$\times$256 and normalize to [0, 1], use AdamW optimizer with betas 0.5 and 0.999, and set the batch size to 2. The model will first train by full loss for 40 epochs, then finetune by smaller lr and only pixel fidelity loss, which is more crucial for image restoration, for 30 epochs. The learning rate is set to $1e^{-4}$ and will gradually decay after epoch 18.

\begin{table*}[!t]
\captionsetup{justification=raggedright,singlelinecheck=true}
\caption{Quantity evaluation for \textbf{Setting 1} on All-weather dataset with state-of-the-art methods.}
\vspace{-0.5cm}
\begin{center}
\scalebox{0.8}{
\begin{tabular}{|c|c|c|c|c|c|c|c|c|}
\hline
 \multicolumn{2}{|c|}{}& \multicolumn{2}{|c|}{\textbf{Snow}} &\multicolumn{2}{|c|}{\textbf{Rain + Fog}} &\multicolumn{2}{|c|}{\textbf{Raindrop}} &{\textbf{Model}} \\

\cline{3-4}
\cline{5-6}
\cline{7-8}
\multicolumn{2}{|c|}{} & \textbf{PSNR} & \textbf{SSIM}& \textbf{PSNR} & \textbf{SSIM}&
\textbf{PSNR}& \textbf{SSIM}& \textbf{param. (M)}\\
\hline

\multirow{6}{*}{Task Specific} &AttentiveGAN \cite{deraindrop}& -& - & - & - & 31.47 & 0.911 & 6.2  \\

&DuRN \cite{durn}& -& - & - & - & 31.10 & 0.916 & 10.2 \\

&pix2pix \cite{pix2pix}& -& - & 19.09 & 0.710 & - & - & 54.4 \\

&HRGAN \cite{heavyrain}& -& - & 21.56 & 0.856 & - & - & 25.1 \\

&JSTASR \cite{JSTASR}& 25.32 & 0.808 & - & - & - & - & 65.9 \\

&DDMSNet \cite{ddmsnet}& 28.85 & 0.877 & - & - & - & - & 229.4 \\

\hline
\multirow{3}{*}{Task Agnoistic} &SwinIR \cite{swinir}& 28.18 & 0.880 & 23.23 & 0.869 & 30.82 & 0.904 & 11.8 \\

&MPRNet \cite{mprnet}& 28.66 & 0.869 & 30.25 & 0.914 & 30.99 & 0.916 & 3.6 \\

&Restormer \cite{restormer}& 29.37 & 0.881 & 29.22 & 0.907 & 31.21 & 0.919 & 18.2 \\

\hline
\multirow{2}{*}{\makecell{Non-blind \\ All-in-one}} &All-in-one \cite{all-in-one}& 28.33 & 0.882 & 24.71 & 0.898 & 31.12 & 0.927 & 44.0 \\
&WGWS-Net \cite{wgwsnet}& 28.91 & 0.856 & 29.28 & 0.922 & 32.01 & 0.925 & 6.0\\
\hline
\multirow{4}{*}{\makecell{Blind \\ All-in-one}}&Transweather \cite{transweather}& 27.83 & 0.845 & 27.63 & 0.884 & 28.85 & 0.884 & 38.1 \\

&Unified Model \cite{unified}& 26.97 & 0.822 & 28.65 & 0.892 & 30.47 & 0.904 & 28.7 \\

&WeatherDiff \cite{weatherdiff}& 28.57 & 0.863 & 28.05 & 0.918 & 29.22 & 0.905 & 83.0 \\

& \textbf{UtilityIR (ours)} & \textbf{29.47} & \textbf{0.879} & \textbf{31.16} & \textbf{0.927} & \textbf{32.01} & \textbf{0.925} & \textbf{26.3}\\
\hline
\end{tabular}}
\vspace{-0.5cm}
\label{table: comp}
\end{center}
\end{table*}

\begin{table*}[!t]
\captionsetup{justification=raggedright,singlelinecheck=true}
\caption{Quantity evaluation for \textbf{Setting 2} on Rain1400, RESIDE OTS, and Snow100k-L dataset with state-of-the-art methods.}
\vspace{-0.5cm}
\begin{center}
\scalebox{0.8}{
\begin{tabular}{|c|c|c|c|c|c|c|}
\hline
 \multicolumn{1}{|c|}{}& \multicolumn{2}{|c|}{\textbf{Rain streak}} &\multicolumn{2}{|c|}{\textbf{Haze}}&\multicolumn{2}{|c|}{\textbf{Snow}}\\

\cline{1-3}
\cline{3-5}
\cline{5-7}
\multicolumn{1}{|c|}{} & \textbf{PSNR} & \textbf{SSIM}& \textbf{PSNR} & \textbf{SSIM}& \textbf{PSNR} & \textbf{SSIM}\\
\hline
Transweather \cite{transweather}& 28.53 & 0.877 & 27.12 & 0.914 &25.90&0.823\\
Unified Model \cite{unified}& 28.69 & 0.872 & 27.34 & 0.902& 25.49& 0.798 \\
WGWS-Net \cite{wgwsnet}& 29.10 & 0.880 & 26.54 & 0.901&28.48&0.859\\
\textbf{UtilityIR (ours)} & \textbf{30.15} & \textbf{0.905} & \textbf{28.37} & \textbf{0.922}&\textbf{29.35}&\textbf{0.886}\\
\hline
\end{tabular}}
\label{table: comp2}
\vspace{-0.8cm}
\end{center}
\end{table*}

\subsection{Comparisons with the State-of-the-art Methods}
We first compare the proposed method with the state-of-the-art methods quantitatively on three different weather removal tasks for setting 1 with PSNR and SSIM. As Tab. \ref{table: comp} shown, UtilityIR outperform same type methods, i.e. blind all-in-one, with large margins on three tasks for \textbf{+1.54 db} on de-raindrop, \textbf{+2.65 db} on rain+fog removal, and \textbf{+0.9 db} on desnowing. The model also achieve state-of-the-art performance compared with other type methods on raindrop and snow removal, and significantly better than all comparison models on challenging rain+fog dataset, the reason might be that the restoration level is fixed for whole image in our method, which is sufficient for global degradation that degradation level is consistent for whole image, but raindrop and snow images are more on local degradation, that might require different restoration level for each patch to further boost the performance. For setting 2 results, UtilityIR consistently outperform comparison methods as shown in Tab. \ref{table: comp2}, more visual results can be found in Supp.

\begin{figure}[t!]
  \centering
    \includegraphics[width=1.0\linewidth]{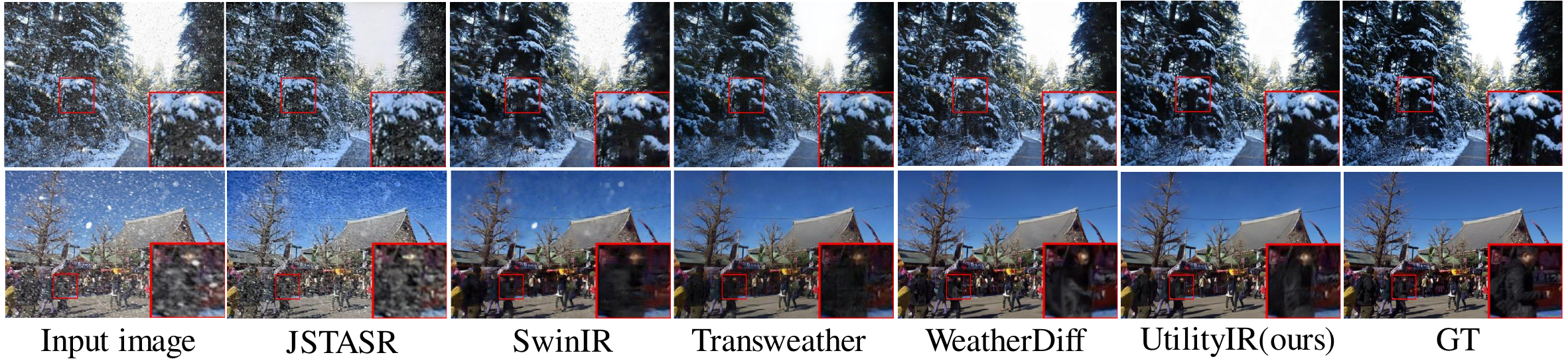}
   \caption{Visual results of desnowing on Snow100K-L dataset. Zoom in for best view.}
   \label{fig:exp_sw}
\end{figure}

\begin{figure}[t!]
  \centering
    \includegraphics[width=1.0\linewidth]{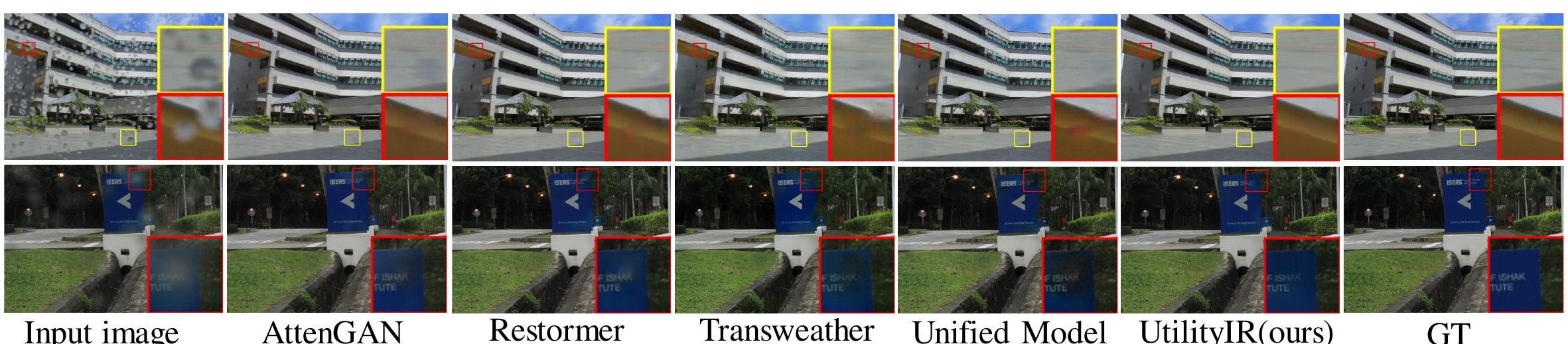}
   \caption{Visual results of deraindrop on Raindrop test-a dataset. Zoom in for best view.}
   \label{fig:exp_rd}
\end{figure}

\begin{figure}[t!]
  \centering
    \includegraphics[width=1.0\linewidth]{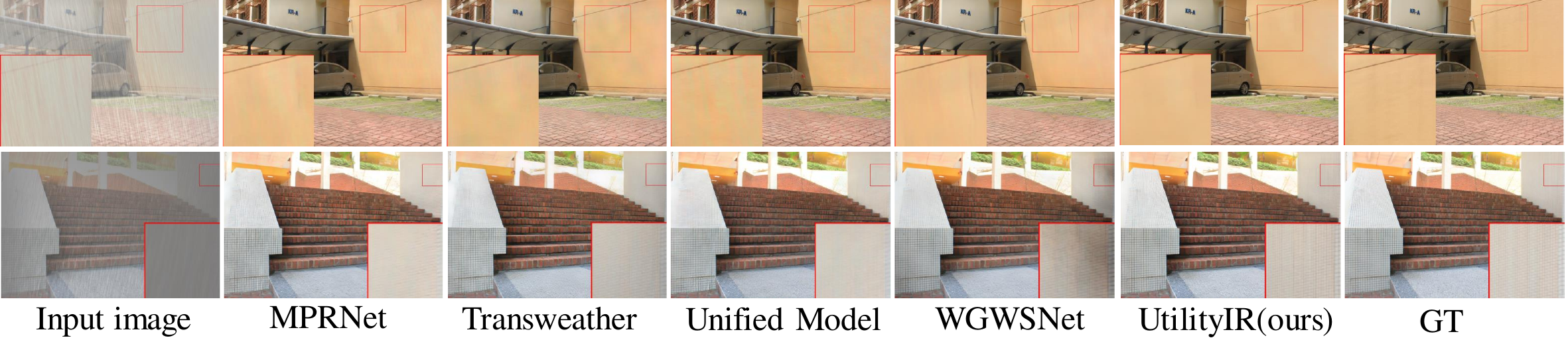}
   \caption{Visual results of derain streak \& defogging on Outdoor-Rain dataset. Zoom in for best view.}
   \label{fig:exp_rs}
\end{figure}

For the visual comparisons, we first demonstrate result on All-weather dataset. As depicted in Fig. \ref{fig:exp_sw}, UtilityIR can remove more snow degradation and generate clearer result compared with other methods. For the raindrop removal, UtilityIR can restore better straight line, while other methods produce twisted line as shown in top images of Fig. \ref{fig:exp_rd}. As presentation of bottom images of Fig. \ref{fig:exp_rd}, all comparison methods remain a black stain artifact after restoration, while UtilityIR can reduce the artifact and obtain cleaner result. For rain+fog removal, all comparison methods result uneven color artifact on the wall of top images in Fig. \ref{fig:exp_rs}, that UtilityIR successfully remove the rain+fog and generate visual pleasing result. As the red area of bottom images in Fig. \ref{fig:exp_rs} shown, our model can obtain clear and higher fidelity to GT. MPRNet and Unified model over-smooth, Transweather remain vertical line color artifact, and WGWS-Net produce a stain artifact. 

\begin{figure}[t!]
  \centering
    \includegraphics[width=1.0\linewidth]{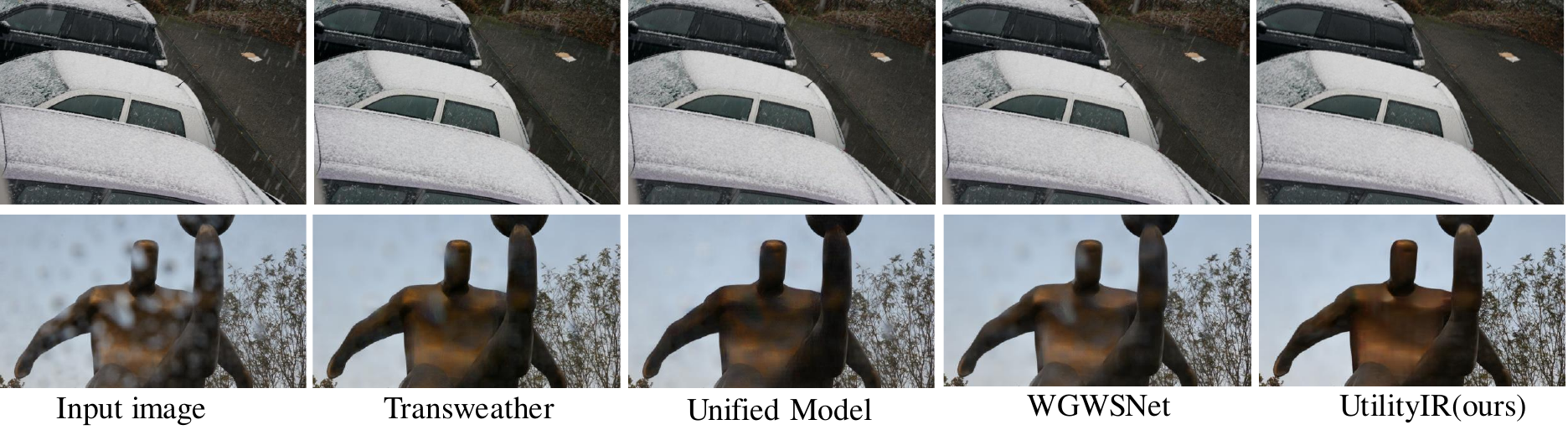}
   \caption{Visual result of real-world images. Zoom in for best view.}
   \label{fig:exp_real}
   \vspace{-0.7cm}
\end{figure}

We further validate generalization ability on real-world weather images. Fig. \ref{fig:exp_real} shows the real-world snow and raindrop results, as figures shown, UtilityIR can remove more degradation compared with other SOTA methods, demonstrate the superiority and excellent generalization ability to real-world images, more reuslt can be found in Supp. 

\begin{figure*}[t]
  \centering
  \includegraphics[width=1.0\linewidth]{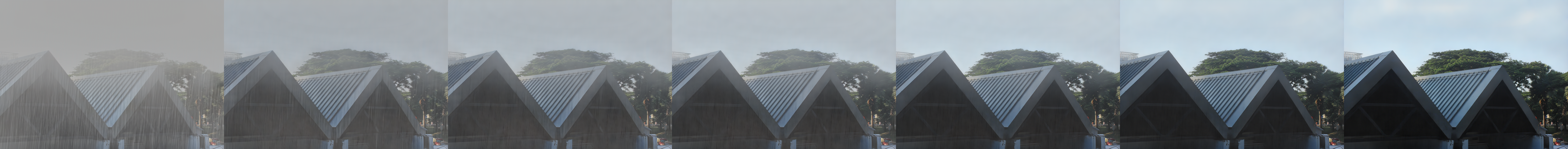}
   \caption{Visual result of restoration level modulation using latent space manipulation, which can achieve by tuning user input parameter.}
   \label{fig:manipulation}
\end{figure*}

\begin{figure}[t!]
  \centering
    \includegraphics[width=1.0\linewidth]{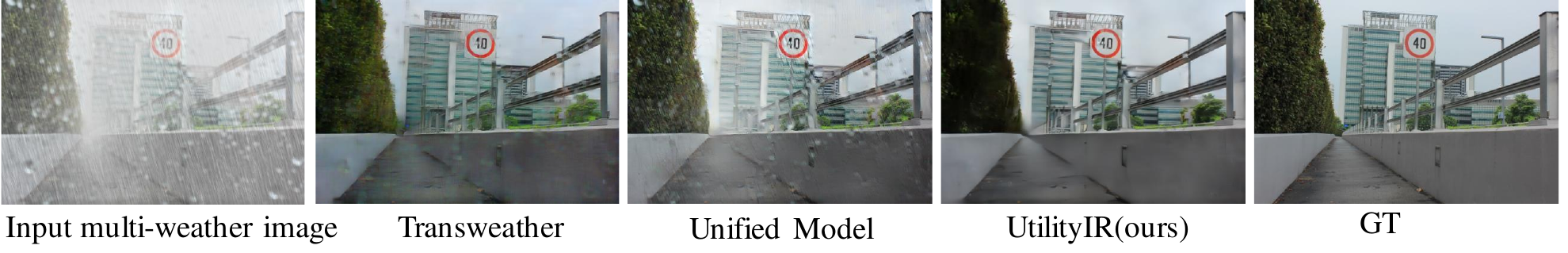}
   \caption{Visual result of combined multiple weather images, our model can perform better than others. Zoom in for the best view.}
   \label{fig:exp_combination}
\end{figure}

\begin{figure}[hptb!]
  \subfloat[]{\begin{minipage}[b][2cm][b]{0.25\linewidth}
              \includegraphics[width=1.0\linewidth]{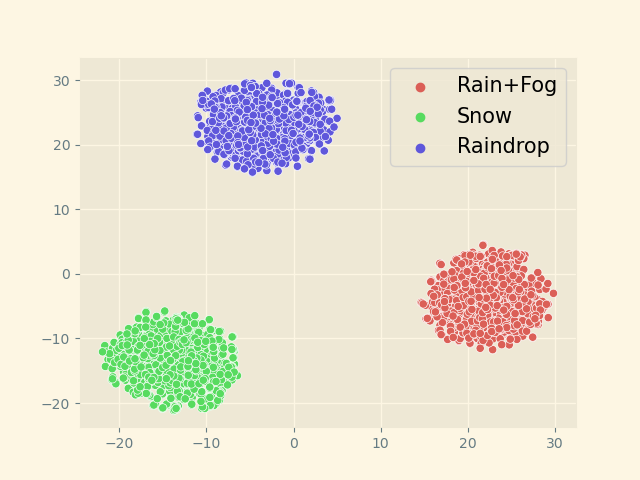}
        \end{minipage}}
  \subfloat[]{\begin{minipage}[b][2cm][b]{0.25\linewidth}
              \includegraphics[width=1.0\linewidth]{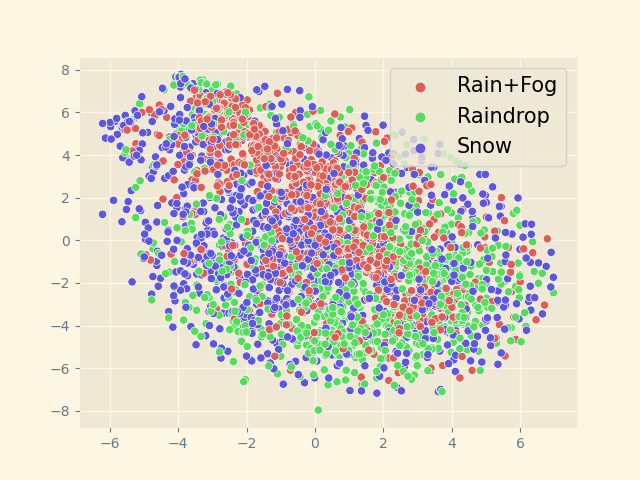}
        \end{minipage}}
  \subfloat[]{\begin{minipage}[b][2cm][b]{0.25\linewidth}
              \includegraphics[width=1.0\linewidth]{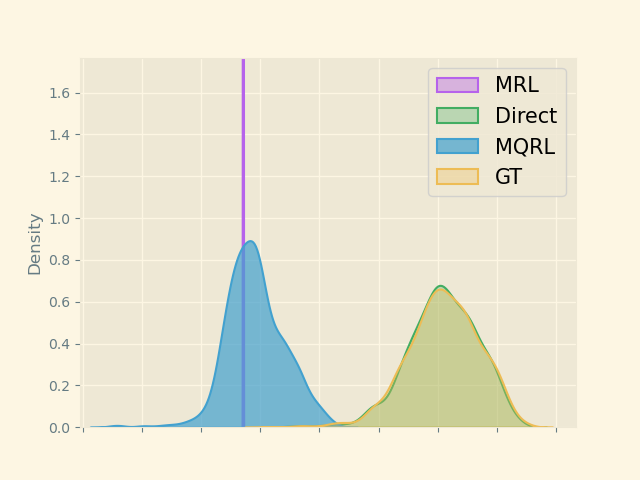}
        \end{minipage}}
    \hfill
    \subfloat[]{\begin{minipage}[b][2cm][b]{0.25\linewidth}
              \includegraphics[width=1.0\linewidth]{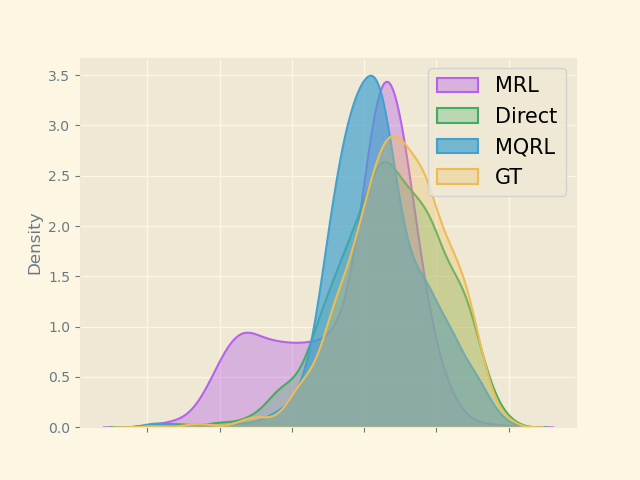}
        \end{minipage}}
    \hfill
   \caption{(a) T-SNE visualization of weather type feature training w/ CL (b) w/o CL. (c) The distribution of GT and predicted IQA score guided by different loss. (d) After applied min-max normalization.}
   \label{fig:exp_type_severity}
\end{figure}

\begin{table}[htbp]
\captionsetup{justification=raggedright,singlelinecheck=true}
\caption{Ablation study on proposed method, we evaluate training loss for weather type and severity to validate the effectiveness. Experiments are conducted on raindrop dataset.}
\begin{center}
\scalebox{0.9}{
\begin{tabular}{|c|c|c|c|}
\hline
\textbf{Losses} & Type & Severity & \textbf{PSNR} $\uparrow$ \\
\hline
None & &  &  30.90 \\
MQRL & & \checkmark & 31.57 \\
CL  & \checkmark &  & 31.65 \\
\hline
CL + direct IQA  & \checkmark & \checkmark & 31.48\\
CL + MRL & \checkmark & \checkmark & 31.60 \\
CL + MQRL & \checkmark & \checkmark & \textbf{31.85}\\
\hline
\end{tabular}}
\label{table:ab_method}
\vspace{-0.5cm}
\end{center}
\end{table}

\subsection{Ablation Study}
\paragraph{Proposed method} 
We evaluate the variant and effectiveness of the weather type and severity by removing corresponding loss function, and report result before finetuning stage. As shown in Tab. \ref{table:ab_method}, with addressing no matter weather type or severity by imposing CL or MQRL, the performance elevate significantly, especially for CL, the result illustrate the multi-domain existing in target distribution for both type and severity, and weather type is more important since it introduce more obviously different degradation and visual appearance. Then we visualize the weather type $F_t$ by T-SNE to evaluate representative. As Fig. \ref{fig:exp_type_severity} (a) and (b) present the T-SNE result of training with and without CL, the feature well gather in corresponding cluster as expectation while imposing CL, and mixed together when removing CL. We suppose the reason behind such perfect clusters result is that we separate the weather information into type and severity, in contrast to other works\cite{airNet, unified, transweather} encode whole weather information into single \textit{weather type} feature, which contain not only weather type but also severity information, we can extract the weather type with more correct semantic and result in better clusters while visualization. After jointly addressing weather type and severity with CL and MQRL, the performance is further improved.

\paragraph{Marginal Quality Ranking Loss} 
To learn more on severity feature, we validate the proposed MQRL by comparing with MRL and directly regress IQA score. We randomize sample 3000 image pairs from All-Weather dataset, and Fig. \ref{fig:exp_type_severity} (c) depict the GT and predicted IQA score distribution guided by each loss. MQRL and MRL predicted incorrect bias compared with GT. After min-max normalization as shown in Fig \ref{fig:exp_type_severity} (d), it can be seen that the ranker guided by MQRL can capture the correct variance, which is more related to interval of ground truth distribution, while ranker guided by MRL fail, and guided by directly regress IQA score result accurately predict bias and variance. However, note that as result shown in the bottom of Tab. \ref{table:ab_method}, severity extraction guided by directly regress IQA score actually deteriorate the image restoration performance because it waste too much effort to regress the bias of target distribution. Model guided by MRL perform better than directly regress IQA score since it provide the ranking information and alleviate the burden of learning exactly value of IQA score, but still slightly degrade the model since lack of learning the information of interval of IQA score. After utilizing MQRL, the model can obtain more informative guidance, and converge to better optima, that demonstrates the key information for learning weather severity is ranking and interval of severity signal, i.e. IQA score.

\subsection{Iterative Restoration}
\label{sec:iter}
\paragraph{Progressive Restoration}
Because of the type and severity aware handling, our model is able to dynamically distinguish severity and restoration level, and progressively improve the image quality while iteratively input the restored image into model. Visualization can be found in Supp.

\paragraph{Restoration Level Modulation}
Thanks to the success of progressive restoration, we further perform restoration level modulation to demonstrate the representative of extracted severity $F_s$. Since $F_s$ is only meaningful in latent space, we utilize latent space manipulation as \cite{mainpulation, uiess} to modulate restoration level. Specifically, we iteratively feed the degraded image $I$ into model to find the meaningful direction in latent space:
\begin{align*}
    &F_s, F_t = DIE(I), & I^{'} = R(I, F_s, F_t)\\
    & F_{s}^{'} = DIE(I^{'})
\end{align*}
where $F_s^{'}$ is predicted severity for restored image $I^{'}$ and represent the smaller restoration level compared with $F_s$, that the vector between $F_s$ and $F_s^{'}$ could form a meaningful direction of modulating restoration level in latent space. Then restoration level modulation can be performed by: 
\begin{align*}
    F_s^{''} &= F_s + \alpha \times (F_s^{'}-F_s) \\
    I_{mod} &= R(I, F_s^{''}, F_t)
\end{align*}
where $\alpha$ is modulating parameter to perform interpolation or extrapolation of two severity $F_s$ and $F_{s}^{'}$, and obtain different restoration level result $I_{mod}$. As shown in Fig.\ref{fig:manipulation}, by tuning the $\alpha$, the model can generate different restoration level result and demonstrate the controllability and representative of extracted severity.

\paragraph{Combination of Multiple Weather Removal}
To be an interesting finding, we observe the proposed method trained on multiple single degradation type data can restore combined type of multiple degradation, which is \textbf{unseen} type data for the trained model. Concretely, for image degraded by $N$ type adverse weather conditions, we can iteratively restore the image $N$ times to remove all degradation. Note that this approach can only perform by blind all-in-one methods since we observe that the order of input weather label does matter for non-blind all-in-one method, i.e. WGWS-Net \cite{wgwsnet}, which make it vulnerable in real-world application. Fig. \ref{fig:exp_combination} shown the comparison with SOTA models, UtilityIR can restore more visual pleasing result, and be more stable and robust compared with other methods, the reason might be our method can be better separate different type of degradation and explicitly handle different severity, and avoid the misuse of incorrect filter for the input image. More detail can be found in Supp. 

\subsection{Limitation}
 Despite that UtilityIR demonstrate superior performance, there are still some limitations, include the proposed MQRL is parameter sensitive, require high computation cost, and suffer from blur result for severe occlusion on high frequency area, which also fail for other SOTA methods, the visualization can be found in Supp., and we will explore these limitations in the future toward practical and efficient all-in-one image restoration.

\section{Conclusion}
In this paper, we proposed a type and severity aware method for blind all-in-one weather removal with CL and MQRL to guide the model extract representative weather information, and utilize a bag of techniques to inject the information into model. Proposed UtilityIR enjoy less parameters and achieve the state-of-the-art performance and can restore unseen combined multi-weather images, and modulate restoration level. Our future work include resolve current limitations and extend to robust unseen multiple degradation restoration toward blind weather removal in the wild.

\section*{Acknowledgements}
This work is supported by NSTC (National Science and Technology Council, Taiwan) 108-2221-E-002-040-MY3.    

{\small
\bibliographystyle{splncs04}
\bibliography{egbib}

\begin{thebibliography}{10}
\providecommand{\url}[1]{\texttt{#1}}
\providecommand{\urlprefix}{URL }
\providecommand{\doi}[1]{https://doi.org/#1}

\bibitem{dehazenet}
Cai, B., Xu, X., Jia, K., Qing, C., Tao, D.: Dehazenet: An end-to-end system for single image haze removal. IEEE transactions on image processing  \textbf{25}(11),  5187--5198 (2016)

\bibitem{twostep_derain}
Cao, M., Gao, Z., Ramesh, B., Mei, T., Cui, J.: A two-stage density-aware single image deraining method. IEEE Transactions on Image Processing  \textbf{30},  6843--6854 (2021)

\bibitem{psfrgan}
Chen, C., Li, X., Yang, L., Lin, X., Zhang, L., Wong, K.Y.K.: Progressive semantic-aware style transformation for blind face restoration. In: Proceedings of the IEEE/CVF conference on computer vision and pattern recognition. pp. 11896--11905 (2021)

\bibitem{ipt}
Chen, H., Wang, Y., Guo, T., Xu, C., Deng, Y., Liu, Z., Ma, S., Xu, C., Xu, C., Gao, W.: Pre-trained image processing transformer. In: Proceedings of the IEEE/CVF conference on computer vision and pattern recognition. pp. 12299--12310 (2021)

\bibitem{nafnet}
Chen, L., Chu, X., Zhang, X., Sun, J.: Simple baselines for image restoration. In: European Conference on Computer Vision. pp. 17--33. Springer (2022)

\bibitem{JSTASR}
Chen, W.T., Fang, H.Y., Ding, J.J., Tsai, C.C., Kuo, S.Y.: Jstasr: Joint size and transparency-aware snow removal algorithm based on modified partial convolution and veiling effect removal. In: European Conference on Computer Vision (2020)

\bibitem{HDCWnet}
Chen, W.T., Fang, H.Y., Hsieh, C.L., Tsai, C.C., Chen, I., Ding, J.J., Kuo, S.Y., et~al.: All snow removed: Single image desnowing algorithm using hierarchical dual-tree complex wavelet representation and contradict channel loss. In: Proceedings of the IEEE/CVF International Conference on Computer Vision. pp. 4196--4205 (2021)

\bibitem{unified}
Chen, W.T., Huang, Z.K., Tsai, C.C., Yang, H.H., Ding, J.J., Kuo, S.Y.: Learning multiple adverse weather removal via two-stage knowledge learning and multi-contrastive regularization: Toward a unified model. In: 2022 IEEE/CVF Conference on Computer Vision and Pattern Recognition (CVPR). pp. 17632--17641 (2022). \doi{10.1109/CVPR52688.2022.01713}

\bibitem{uiess}
Chen, Y.W., Pei, S.C.: Domain adaptation for underwater image enhancement via content and style separation. IEEE Access  \textbf{10},  90523--90534 (2022)

\bibitem{hardgan}
Deng, Q., Huang, Z., Tsai, C.C., Lin, C.W.: Hardgan: A haze-aware representation distillation gan for single image dehazing. In: European conference on computer vision. pp. 722--738. Springer (2020)

\bibitem{rain1400}
Fu, X., Huang, J., Zeng, D., Huang, Y., Ding, X., Paisley, J.: Removing rain from single images via a deep detail network. In: Proceedings of the IEEE conference on computer vision and pattern recognition. pp. 3855--3863 (2017)

\bibitem{cbdnet}
Guo, S., Yan, Z., Zhang, K., Zuo, W., Zhang, L.: Toward convolutional blind denoising of real photographs. In: Proceedings of the IEEE/CVF conference on computer vision and pattern recognition. pp. 1712--1722 (2019)

\bibitem{cl}
Hadsell, R., Chopra, S., LeCun, Y.: Dimensionality reduction by learning an invariant mapping. In: 2006 IEEE Computer Society Conference on Computer Vision and Pattern Recognition (CVPR'06). vol.~2, pp. 1735--1742. IEEE (2006)

\bibitem{adaFM}
He, J., Dong, C., Qiao, Y.: Modulating image restoration with continual levels via adaptive feature modification layers. In: The IEEE Conference on Computer Vision and Pattern Recognition (CVPR) (June 2019)

\bibitem{dcp}
He, K., Sun, J., Tang, X.: Single image haze removal using dark channel prior. IEEE Transactions on Pattern Analysis and Machine Intelligence  \textbf{33}(12),  2341--2353 (2011). \doi{10.1109/TPAMI.2010.168}

\bibitem{adain}
Huang, X., Belongie, S.: Arbitrary style transfer in real-time with adaptive instance normalization. In: Proceedings of the IEEE international conference on computer vision. pp. 1501--1510 (2017)

\bibitem{lgAdaIN}
Huang, X., Mallya, A., Wang, T.C., Liu, M.Y.: Multimodal conditional image synthesis with product-of-experts gans. In: Computer Vision--ECCV 2022: 17th European Conference, Tel Aviv, Israel, October 23--27, 2022, Proceedings, Part XVI. pp. 91--109. Springer (2022)

\bibitem{pix2pix}
Isola, P., Zhu, J.Y., Zhou, T., Efros, A.A.: Image-to-image translation with conditional adversarial networks. In: Proceedings of the IEEE conference on computer vision and pattern recognition. pp. 1125--1134 (2017)

\bibitem{aodnet}
Li, B., Peng, X., Wang, Z., Xu, J., Feng, D.: Aod-net: All-in-one dehazing network. In: 2017 IEEE International Conference on Computer Vision (ICCV). pp. 4780--4788 (2017). \doi{10.1109/ICCV.2017.511}

\bibitem{reside}
Li, B., Ren, W., Fu, D., Tao, D., Feng, D., Zeng, W., Wang, Z.: Benchmarking single-image dehazing and beyond. IEEE Transactions on Image Processing  \textbf{28}(1),  492--505 (2018)

\bibitem{zeroshotdehaze}
Li, B., Gou, Y., Liu, J.Z., Zhu, H., Zhou, J.T., Peng, X.: Zero-shot image dehazing. IEEE Transactions on Image Processing  \textbf{29},  8457--8466 (2020). \doi{10.1109/TIP.2020.3016134}

\bibitem{airNet}
Li, B., Liu, X., Hu, P., Wu, Z., Lv, J., Peng, X.: All-in-one image restoration for unknown corruption. In: Proceedings of the IEEE/CVF Conference on Computer Vision and Pattern Recognition. pp. 17452--17462 (2022)

\bibitem{QGCN}
Li, J., Wang, Y., Xie, H., Ma, K.K.: Learning a single model with a wide range of quality factors for jpeg image artifacts removal. IEEE Transactions on Image Processing  \textbf{29},  8842--8854 (2020)

\bibitem{heavyrain}
Li, R., Cheong, L.F., Tan, R.T.: Heavy rain image restoration: Integrating physics model and conditional adversarial learning. In: Proceedings of the IEEE/CVF conference on computer vision and pattern recognition. pp. 1633--1642 (2019)

\bibitem{all-in-one}
Li, R., Tan, R.T., Cheong, L.F.: All in one bad weather removal using architectural search. In: Proceedings of the IEEE/CVF conference on computer vision and pattern recognition. pp. 3175--3185 (2020)

\bibitem{grl}
Li, Y., Fan, Y., Xiang, X., Demandolx, D., Ranjan, R., Timofte, R., Van~Gool, L.: Efficient and explicit modelling of image hierarchies for image restoration. In: Proceedings of the IEEE/CVF Conference on Computer Vision and Pattern Recognition (CVPR). pp. 18278--18289 (June 2023)

\bibitem{swinir}
Liang, J., Cao, J., Sun, G., Zhang, K., Van~Gool, L., Timofte, R.: Swinir: Image restoration using swin transformer. In: Proceedings of the IEEE/CVF international conference on computer vision. pp. 1833--1844 (2021)

\bibitem{stdrkl}
Liu, X., Van De~Weijer, J., Bagdanov, A.D.: Rankiqa: Learning from rankings for no-reference image quality assessment. In: Proceedings of the IEEE international conference on computer vision. pp. 1040--1049 (2017)

\bibitem{durn}
Liu, X., Suganuma, M., Sun, Z., Okatani, T.: Dual residual networks leveraging the potential of paired operations for image restoration. In: Proceedings of the IEEE/CVF Conference on Computer Vision and Pattern Recognition. pp. 7007--7016 (2019)

\bibitem{desnownet}
Liu, Y.F., Jaw, D.W., Huang, S.C., Hwang, J.N.: Desnownet: Context-aware deep network for snow removal. IEEE Transactions on Image Processing  \textbf{27}(6),  3064--3073 (2018)

\bibitem{niqe}
Mittal, A., Soundararajan, R., Bovik, A.C.: Making a “completely blind” image quality analyzer. IEEE Signal processing letters  \textbf{20}(3),  209--212 (2012)

\bibitem{weatherdiff}
{\"O}zdenizci, O., Legenstein, R.: Restoring vision in adverse weather conditions with patch-based denoising diffusion models. IEEE Transactions on Pattern Analysis and Machine Intelligence  (2023)

\bibitem{adms}
Park, D., Lee, B.H., Chun, S.Y.: All-in-one image restoration for unknown degradations using adaptive discriminative filters for specific degradations. In: Proceedings of the IEEE/CVF Conference on Computer Vision and Pattern Recognition. pp. 5815--5824 (2023)

\bibitem{deraindrop}
Qian, R., Tan, R.T., Yang, W., Su, J., Liu, J.: Attentive generative adversarial network for raindrop removal from a single image. In: Proceedings of the IEEE conference on computer vision and pattern recognition. pp. 2482--2491 (2018)

\bibitem{prenet}
Ren, D., Zuo, W., Hu, Q., Zhu, P., Meng, D.: Progressive image deraining networks: A better and simpler baseline. In: Proceedings of the IEEE/CVF conference on computer vision and pattern recognition. pp. 3937--3946 (2019)

\bibitem{mainpulation}
Shen, Y., Gu, J., Tang, X., Zhou, B.: Interpreting the latent space of gans for semantic face editing. In: Proceedings of the IEEE/CVF conference on computer vision and pattern recognition. pp. 9243--9252 (2020)

\bibitem{cvprw2022}
Shin, W., Ahn, N., Moon, J.H., Sohn, K.A.: Exploiting distortion information for multi-degraded image restoration. In: Proceedings of the IEEE/CVF Conference on Computer Vision and Pattern Recognition. pp. 537--546 (2022)

\bibitem{maxim}
Tu, Z., Talebi, H., Zhang, H., Yang, F., Milanfar, P., Bovik, A., Li, Y.: Maxim: Multi-axis mlp for image processing. CVPR  (2022)

\bibitem{transweather}
Valanarasu, J.M.J., Yasarla, R., Patel, V.M.: Transweather: Transformer-based restoration of images degraded by adverse weather conditions. In: Proceedings of the IEEE/CVF Conference on Computer Vision and Pattern Recognition (CVPR). pp. 2353--2363 (June 2022)

\bibitem{dln}
Wang, L.W., Liu, Z.S., Siu, W.C., Lun, D.P.: Lightening network for low-light image enhancement. IEEE Transactions on Image Processing  \textbf{29},  7984--7996 (2020)

\bibitem{gridformer}
Wang, T., Zhang, K., Shao, Z., Luo, W., Stenger, B., Lu, T., Kim, T.K., Liu, W., Li, H.: Gridformer: Residual dense transformer with grid structure for image restoration in adverse weather conditions. arXiv preprint arXiv:2305.17863  (2023)

\bibitem{dauie}
Wang, Z., Shen, L., Xu, M., Yu, M., Wang, K., Lin, Y.: Domain adaptation for underwater image enhancement. IEEE Transactions on Image Processing  \textbf{32},  1442--1457 (2023)

\bibitem{restoreformer}
Wang, Z., Zhang, J., Chen, R., Wang, W., Luo, P.: Restoreformer: High-quality blind face restoration from undegraded key-value pairs. In: Proceedings of the IEEE/CVF Conference on Computer Vision and Pattern Recognition. pp. 17512--17521 (2022)

\bibitem{aecrnet}
Wu, H., Qu, Y., Lin, S., Zhou, J., Qiao, R., Zhang, Z., Xie, Y., Ma, L.: Contrastive learning for compact single image dehazing. In: Proceedings of the IEEE/CVF Conference on Computer Vision and Pattern Recognition. pp. 10551--10560 (2021)

\bibitem{jorder}
Yang, W., Tan, R.T., Feng, J., Liu, J., Guo, Z., Yan, S.: Deep joint rain detection and removal from a single image. In: Proceedings of the IEEE conference on computer vision and pattern recognition. pp. 1357--1366 (2017)

\bibitem{vrdir}
Yang, Z., Huang, J., Chang, J., Zhou, M., Yu, H., Zhang, J., Zhao, F.: Visual recognition-driven image restoration for multiple degradation with intrinsic semantics recovery. In: Proceedings of the IEEE/CVF Conference on Computer Vision and Pattern Recognition. pp. 14059--14070 (2023)

\bibitem{twostep_dehaze}
Yi, X., Ma, B., Zhang, Y., Liu, L., Wu, J.: Two-step image dehazing with intra-domain and inter-domain adaptation. Neurocomputing  \textbf{485},  1--11 (2022)

\bibitem{restormer}
Zamir, S.W., Arora, A., Khan, S., Hayat, M., Khan, F.S., Yang, M.H.: Restormer: Efficient transformer for high-resolution image restoration. In: Proceedings of the IEEE/CVF Conference on Computer Vision and Pattern Recognition. pp. 5728--5739 (2022)

\bibitem{mirnet}
Zamir, S.W., Arora, A., Khan, S., Hayat, M., Khan, F.S., Yang, M.H., Shao, L.: Learning enriched features for real image restoration and enhancement. In: Computer Vision--ECCV 2020: 16th European Conference, Glasgow, UK, August 23--28, 2020, Proceedings, Part XXV 16. pp. 492--511. Springer (2020)

\bibitem{mprnet}
Zamir, S.W., Arora, A., Khan, S., Hayat, M., Khan, F.S., Yang, M.H., Shao, L.: Multi-stage progressive image restoration. In: Proceedings of the IEEE/CVF conference on computer vision and pattern recognition. pp. 14821--14831 (2021)

\bibitem{didmdn}
Zhang, H., Patel, V.M.: Density-aware single image de-raining using a multi-stream dense network. In: Proceedings of the IEEE conference on computer vision and pattern recognition. pp. 695--704 (2018)

\bibitem{idr}
Zhang, J., Huang, J., Yao, M., Yang, Z., Yu, H., Zhou, M., Zhao, F.: Ingredient-oriented multi-degradation learning for image restoration. In: Proceedings of the IEEE/CVF Conference on Computer Vision and Pattern Recognition. pp. 5825--5835 (2023)

\bibitem{ddmsnet}
Zhang, K., Li, R., Yu, Y., Luo, W., Li, C.: Deep dense multi-scale network for snow removal using semantic and depth priors. IEEE Transactions on Image Processing  \textbf{30},  7419--7431 (2021)

\bibitem{rankersrgan}
Zhang, W., Liu, Y., Dong, C., Qiao, Y.: Ranksrgan: Generative adversarial networks with ranker for image super-resolution. In: Proceedings of the IEEE/CVF International Conference on Computer Vision (ICCV) (October 2019)

\bibitem{wgwsnet}
Zhu, Y., Wang, T., Fu, X., Yang, X., Guo, X., Dai, J., Qiao, Y., Hu, X.: Learning weather-general and weather-specific features for image restoration under multiple adverse weather conditions. In: Proceedings of the IEEE/CVF Conference on Computer Vision and Pattern Recognition. pp. 21747--21758 (2023)

\end{thebibliography}
}



\title{Always Clear Days: Degradation Type and Severity Aware All-In-One Adverse Weather Removal Supplementary Material} 

\titlerunning{Degradation Type and Severity Aware All-In-One Adverse Weather Removal}

\author{Yu-Wei Chen\inst{1}\orcidlink{0000-0001-9127-6536} \and
Soo-Chang Pei\inst{1,2}\orcidlink{0000-0003-2448-4196}}

\authorrunning{Chen and Pei.}

\institute{Graduate Institute of Communication Engineering, National Taiwan University \and Department of Electrical Engineering, National Taiwan University
\email{\{r09942066, peisc\}@ntu.edu.tw}}

\maketitle
\renewcommand\thesection{A\arabic{section}}
\renewcommand{\thefigure}{A\arabic{figure}}

\section{Combination of Multi-weather Removal}
\label{sec:intro}
\begin{figure*}[hptb!]
\subfloat[Input]{\begin{minipage}[b][][b]{0.5\linewidth}
              \includegraphics[width=1.0\linewidth]{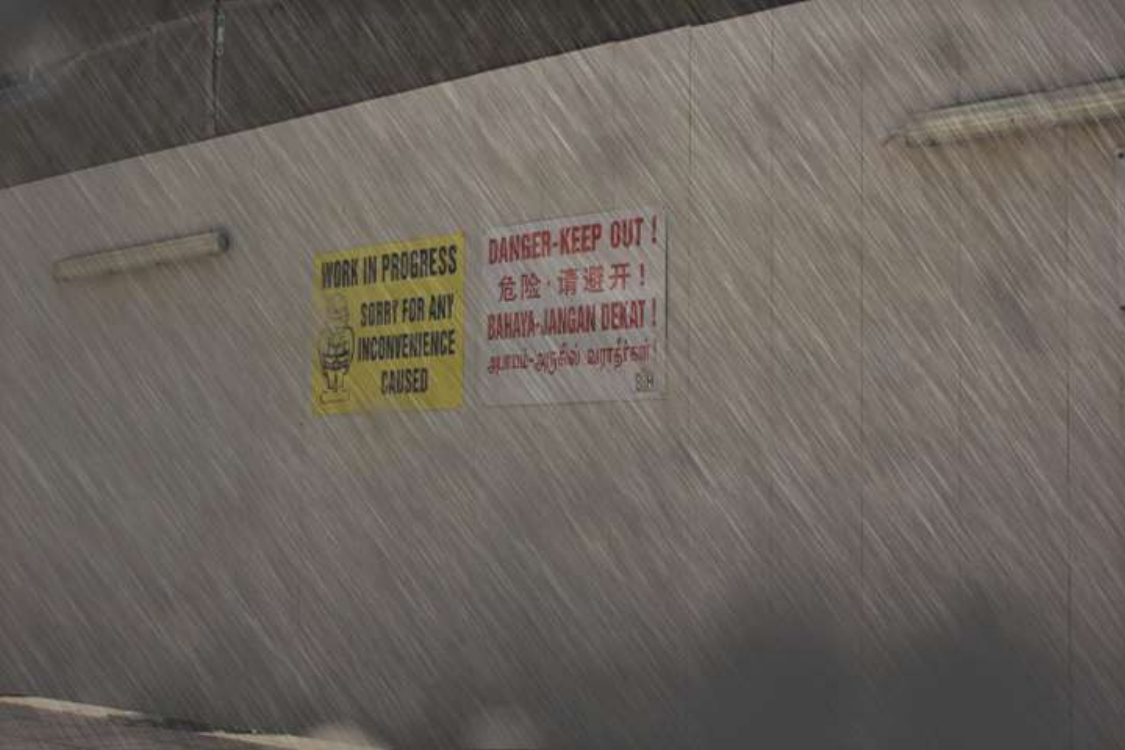}
        \end{minipage}}
    \hfill
    \subfloat[Transweather \cite{transweather}]{\begin{minipage}[b][][b]{0.5\linewidth}
              \includegraphics[width=1.0\linewidth]{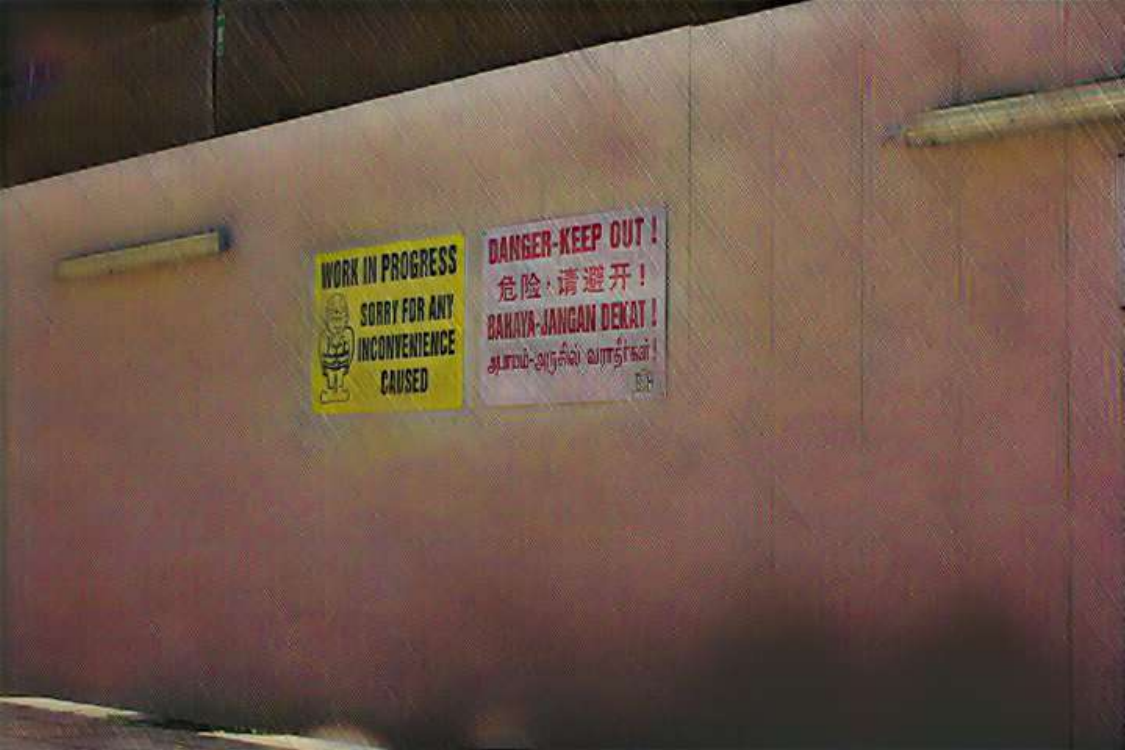}
        \end{minipage}}
    \hfill
  \subfloat[Unified Model \cite{unified}]{\begin{minipage}[b][][b]{0.5\linewidth}
              \includegraphics[width=1.0\linewidth]{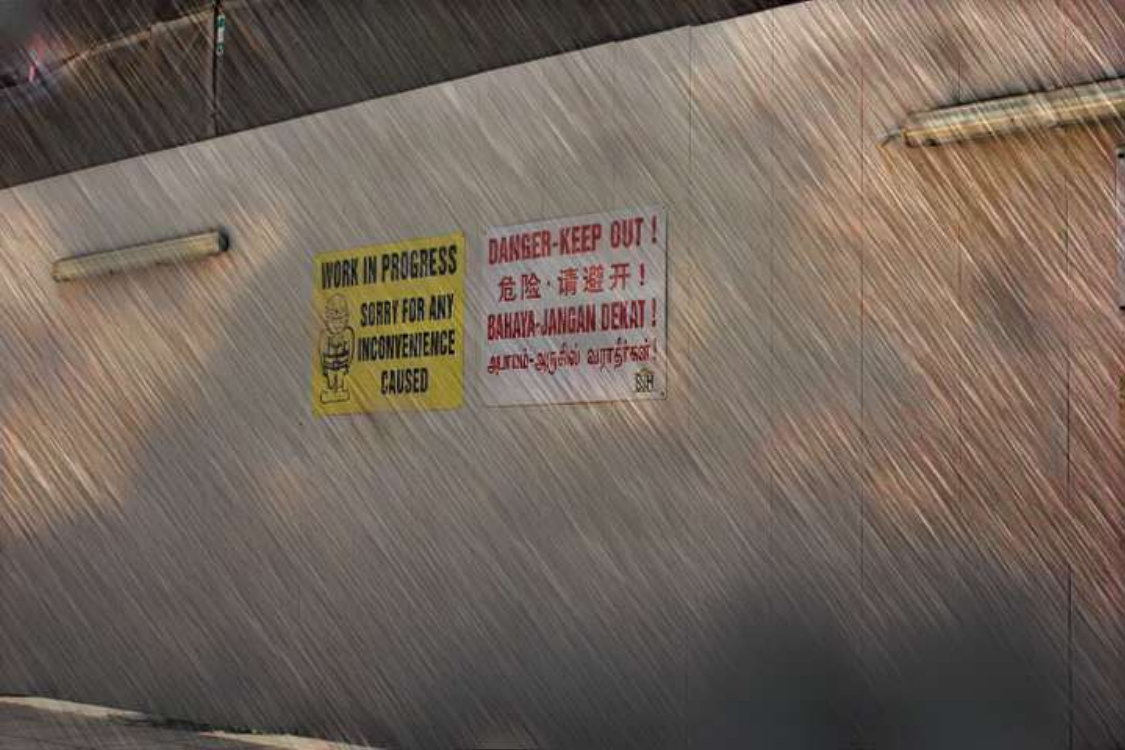}
        \end{minipage}}
    \hfill
    \subfloat[Ours]{\begin{minipage}[b][][b]{0.5\linewidth}
              \includegraphics[width=1.0\linewidth]{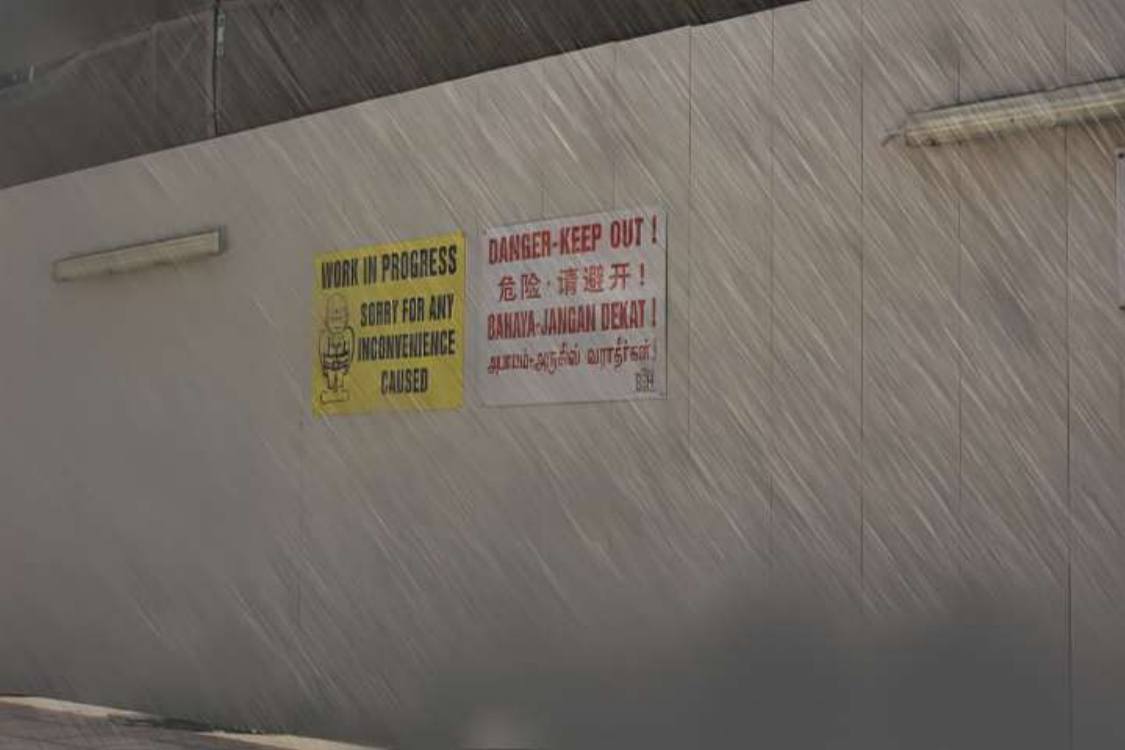}
        \end{minipage}}
    \hfill
    \subfloat[GT]{\begin{minipage}[b][][b]{0.5\linewidth}
              \includegraphics[width=1.0\linewidth]{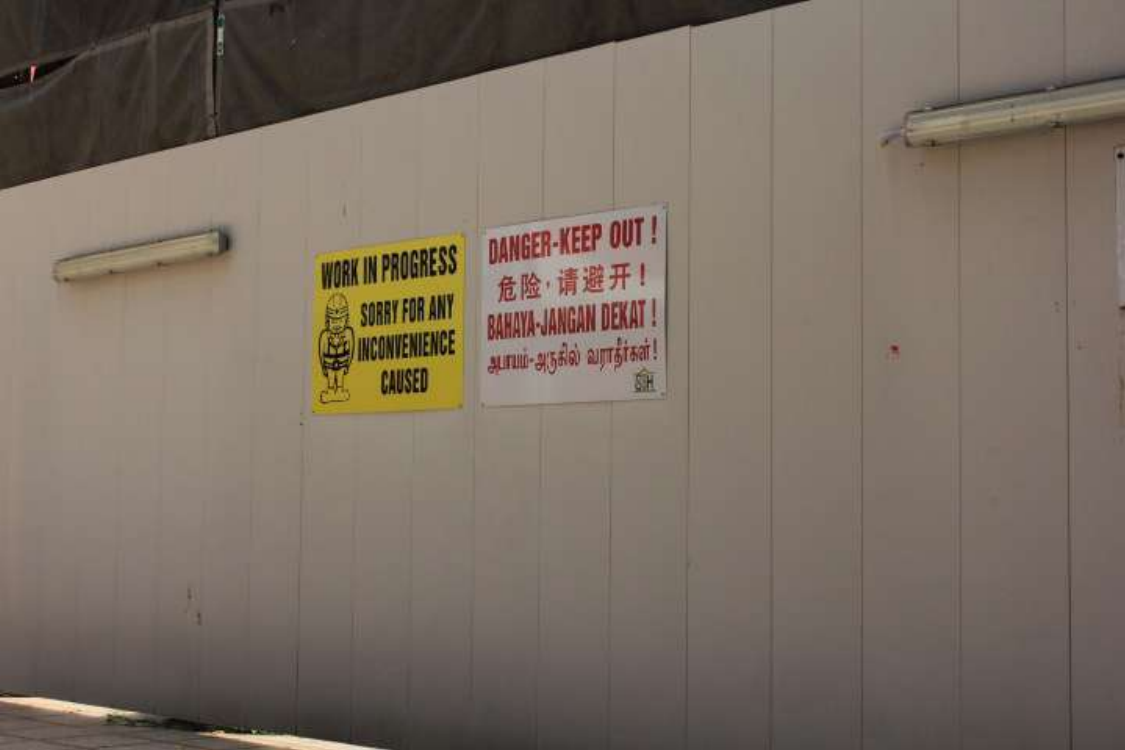}
        \end{minipage}}
    \hfill
   \caption{Failure cases of unseen combined multiple weather input result. Our model won't generate unnatural result.}
   \label{fig:exp_combined_fail}
\end{figure*}

As mentioned in body text, we observe that UtilityIR can perform more stable compared with other state-of-the-art models. For the failure cases shown in Fig. \ref{fig:exp_combined_fail}, UtilityIR can preserve input image appearance, in contrast, other comparison methods produce extremely visual unpleasing result. We also quantitatively validate the observation through report the NIQE \cite{niqe}, which is closer to human perception than PSNR and SSIM. As shown in Table \ref{table:combined}, our method achieve best performance on worse case and standard deviation, which imply the better stability. 

\begin{table}[htbp]
\captionsetup{justification=raggedright,singlelinecheck=true}
\caption{Quantitatively evaluation of unseen combined multiple weather image. Our model can produce better performance with higher stability.}
\begin{center}
\begin{tabular}{|c|c|c|}
\hline
\textbf{Methods} & \textbf{Max (worse case) $\downarrow$} & \textbf{std} $\downarrow$ \\
\hline
Input & 12.173 &  1.712 \\
Transweather \cite{transweather} & 32.249 & 5.847 \\
Unified Model \cite{unified} & 15.249 & 2.654 \\
UtilityIR (Ours)  & \textbf{11.58} & \textbf{1.631} \\
GT  & 8.212 & 0.921 \\
\hline
\end{tabular}
\label{table:combined}
\vspace{-0.5cm}
\end{center}
\end{table}

\section{Progressive Restoration}
\begin{figure*}[hptb!]
\subfloat[Input]{\begin{minipage}[b][][b]{0.5\linewidth}
              \includegraphics[width=1.0\linewidth]{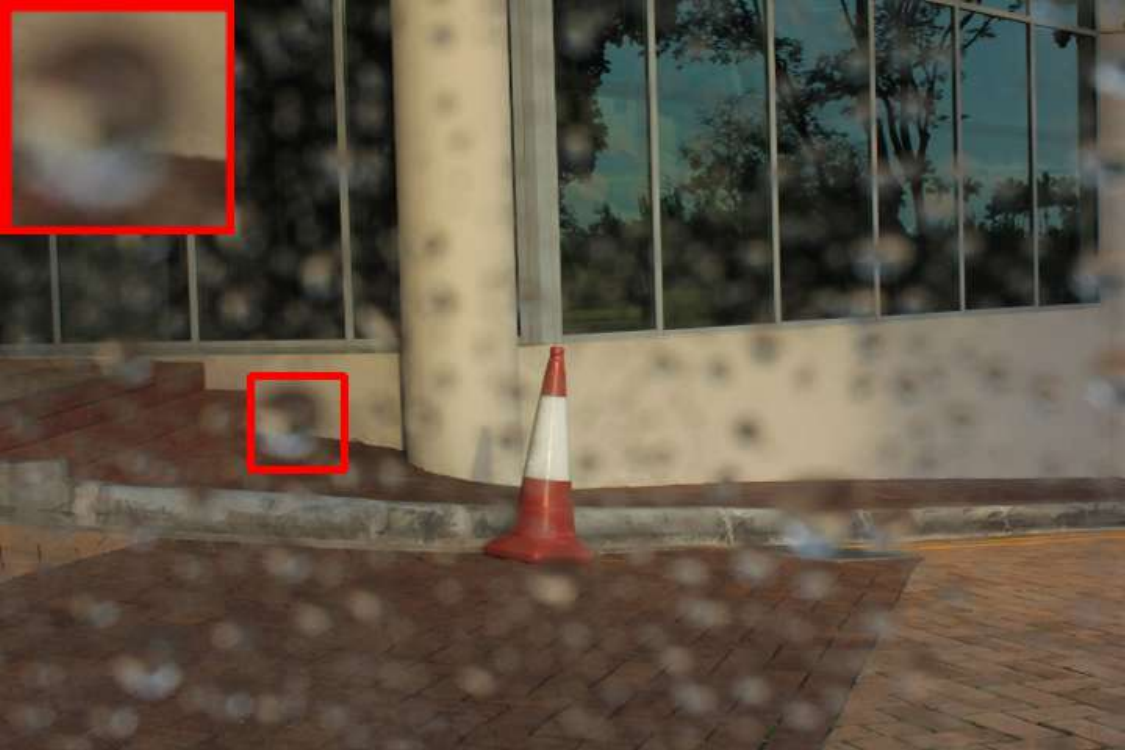}
        \end{minipage}}
    \hfill
    \subfloat[GT]{\begin{minipage}[b][][b]{0.5\linewidth}
              \includegraphics[width=1.0\linewidth]{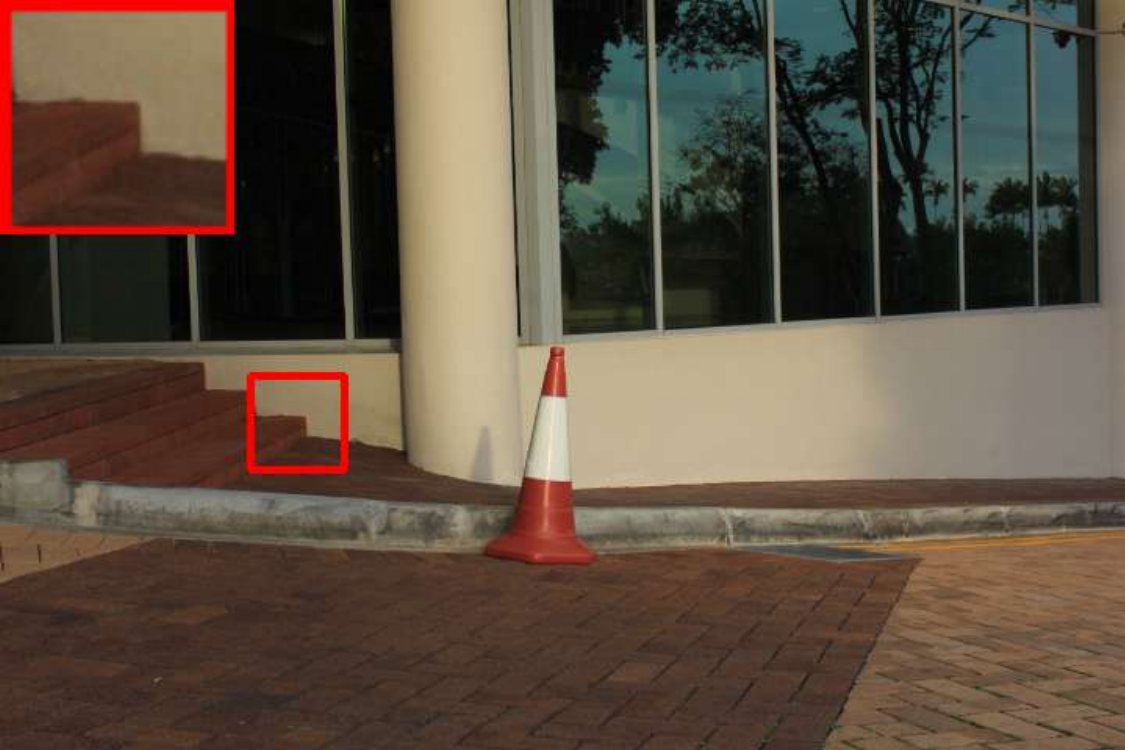}
        \end{minipage}}
    \hfill
  \subfloat[Iteration 1]{\begin{minipage}[b][][b]{0.5\linewidth}
              \includegraphics[width=1.0\linewidth]{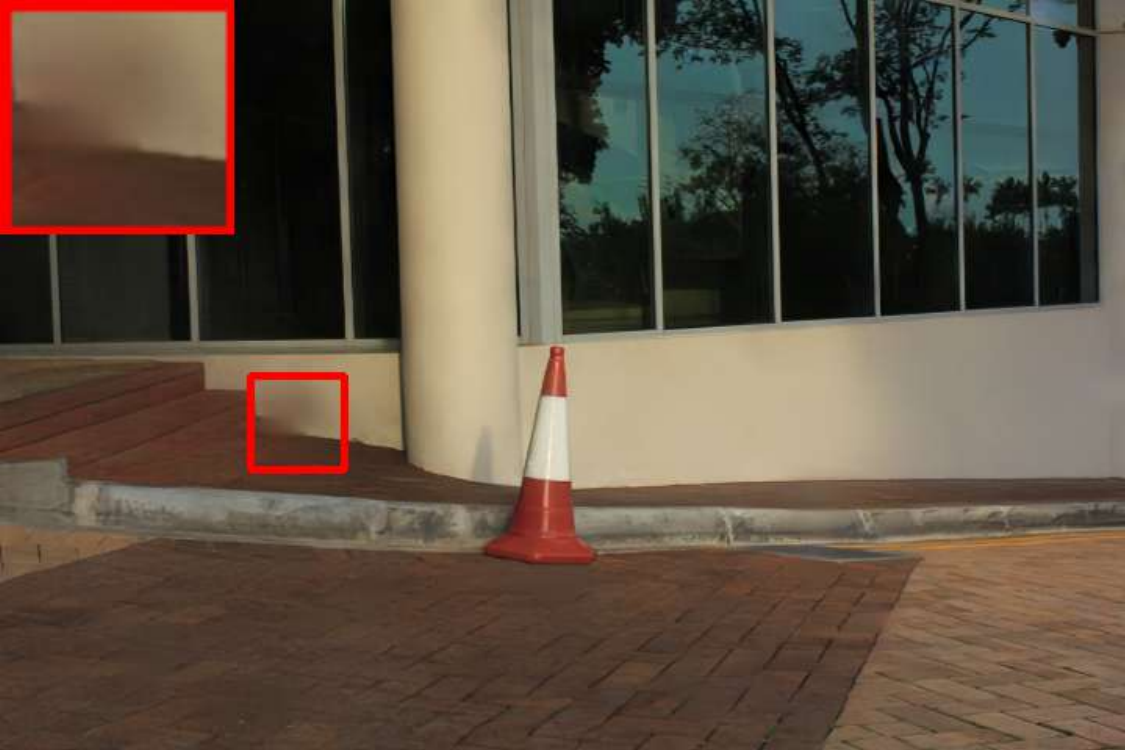}
        \end{minipage}}
    \hfill
    \subfloat[Iteration 2]{\begin{minipage}[b][][b]{0.5\linewidth}
              \includegraphics[width=1.0\linewidth]{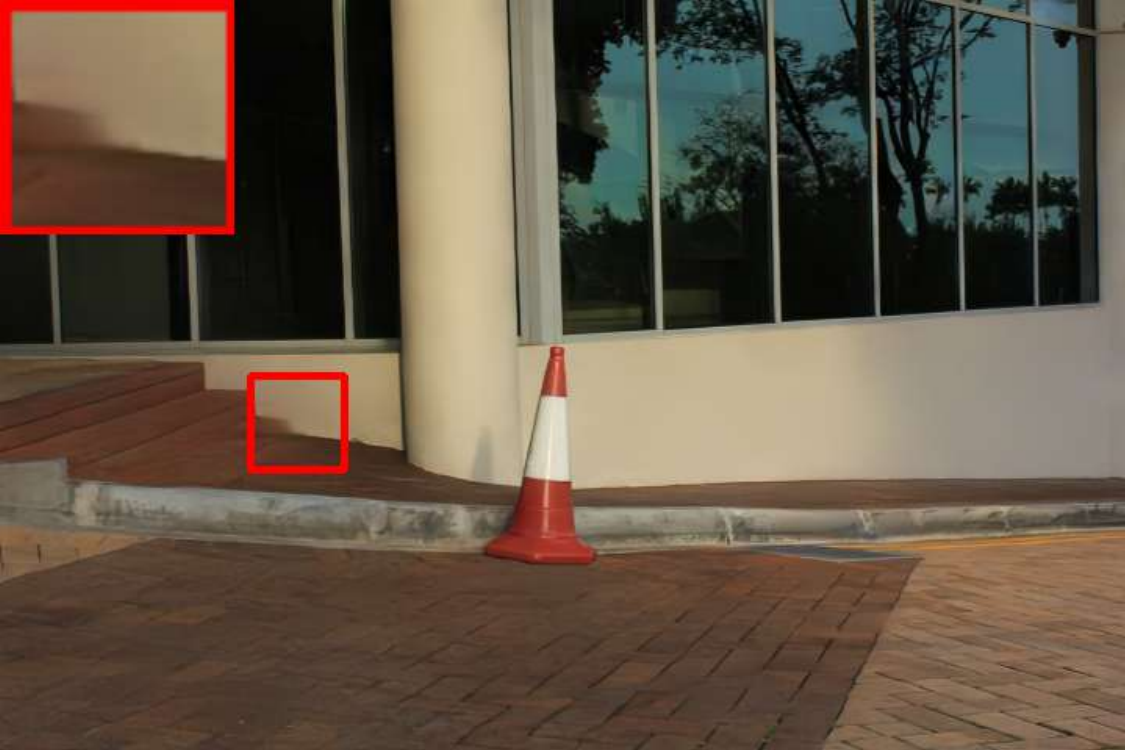}
        \end{minipage}}
    \hfill
   \caption{Visualization of progressive restoration. While iteratively feed image image into model, our model are able to progressive restoration.}
   \label{fig:exp_progressive}
\end{figure*}

We demonstrate the visualization result of progressive restoration in this section. Thanks to severity aware training, as shown in Fig. \ref{fig:exp_progressive}, while iteratively feed the enhanced image into model, the remaining degradation can be progressively removed and obtain more visual pleasing result.  

\section{Limitation}
\begin{figure*}[hptb!]
   \vspace*{-0.4in}
  \subfloat[Input image]{\begin{minipage}[b][][b]{0.33\linewidth}
              \vspace*{\fill}
              \centering
              \includegraphics[width=1.0\linewidth]{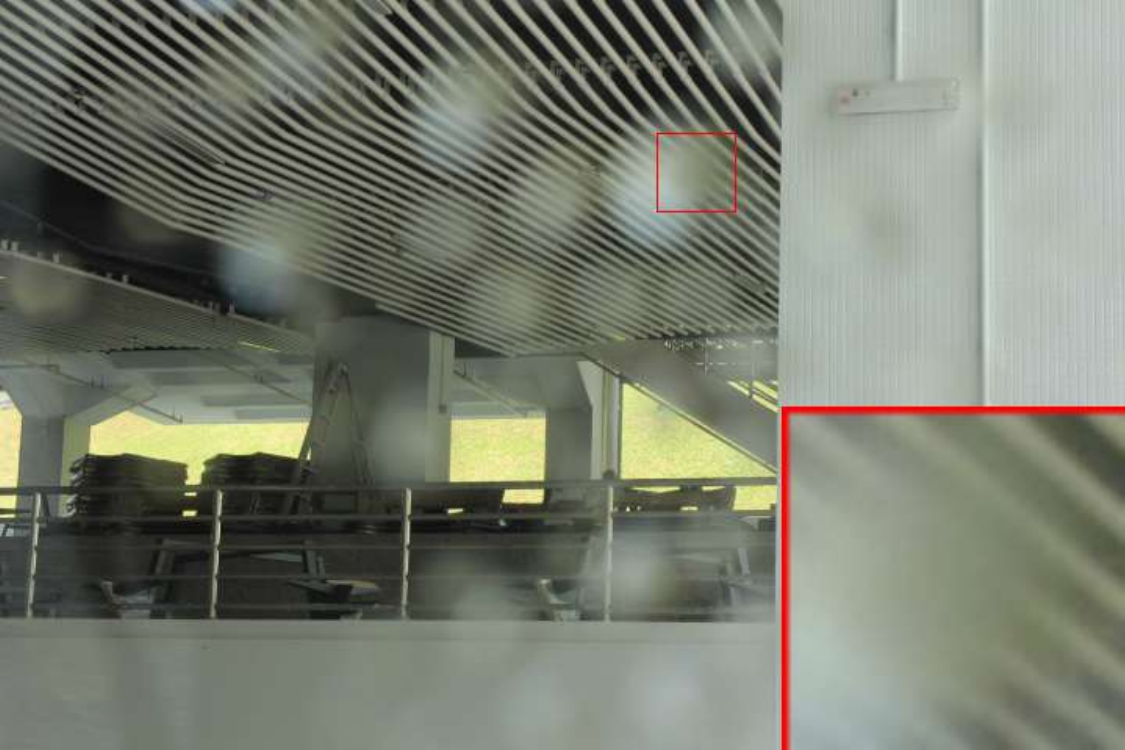} 
        \end{minipage}}
    \hfill
    \subfloat[Transweather \cite{transweather}]{\begin{minipage}[b][][b]{0.33\linewidth}
              \vspace*{\fill}
              \centering
              \includegraphics[width=1.0\linewidth]{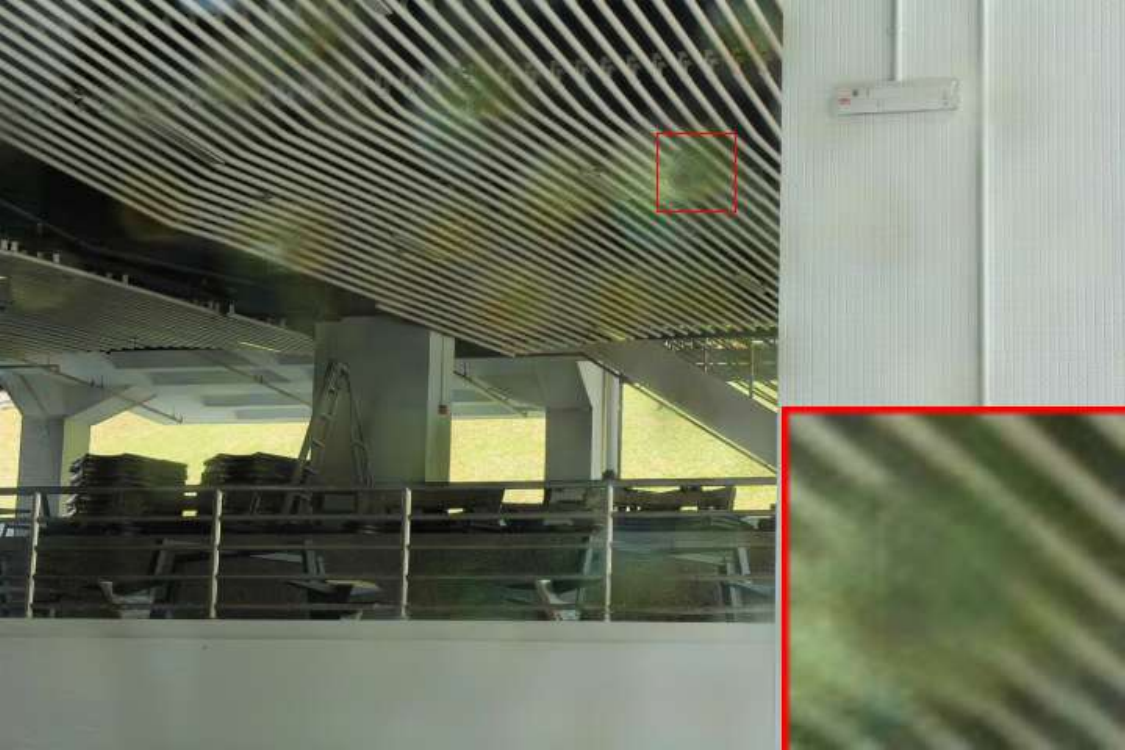} 
        \end{minipage}}
    \hfill
    \subfloat[WGWS-Net \cite{wgwsnet}]{\begin{minipage}[b][][b]{0.33\linewidth}
          \vspace*{\fill}
          \centering
          \includegraphics[width=1.0\linewidth]{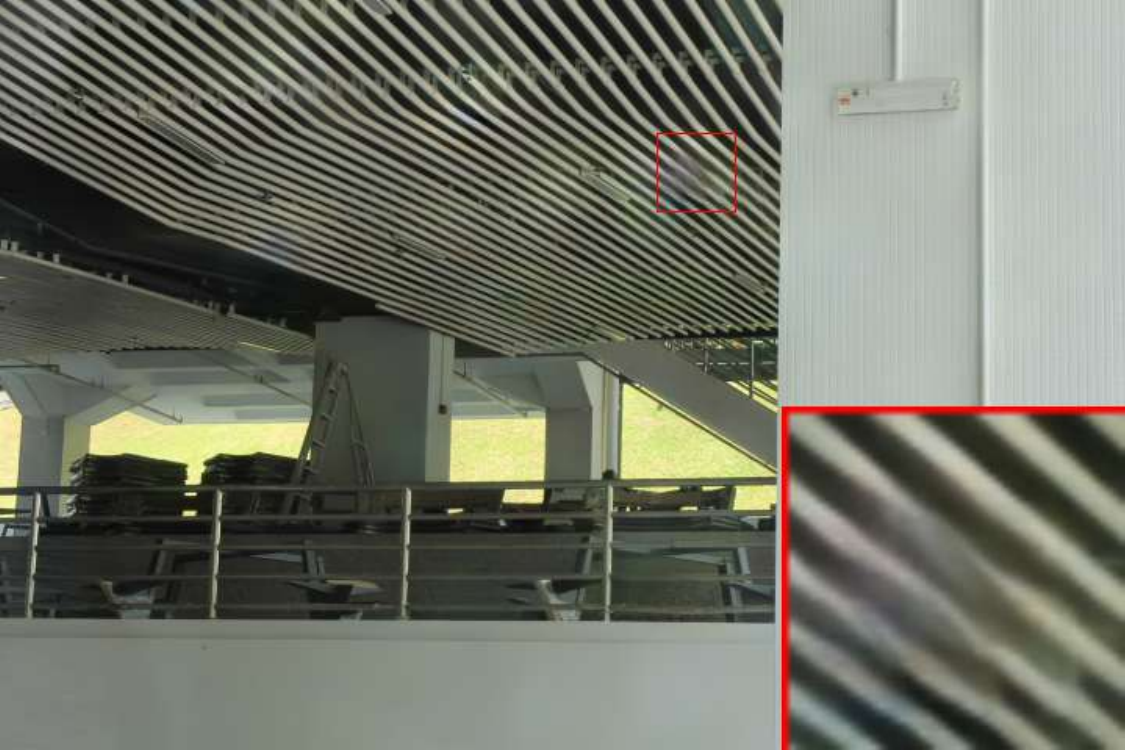} 
        \end{minipage}}
   
    \hfill
    \subfloat[Weatherdiff \cite{weatherdiff}]{\begin{minipage}[b][][b]{0.33\linewidth}
          \vspace*{\fill}
          \centering
          \includegraphics[width=1.0\linewidth]{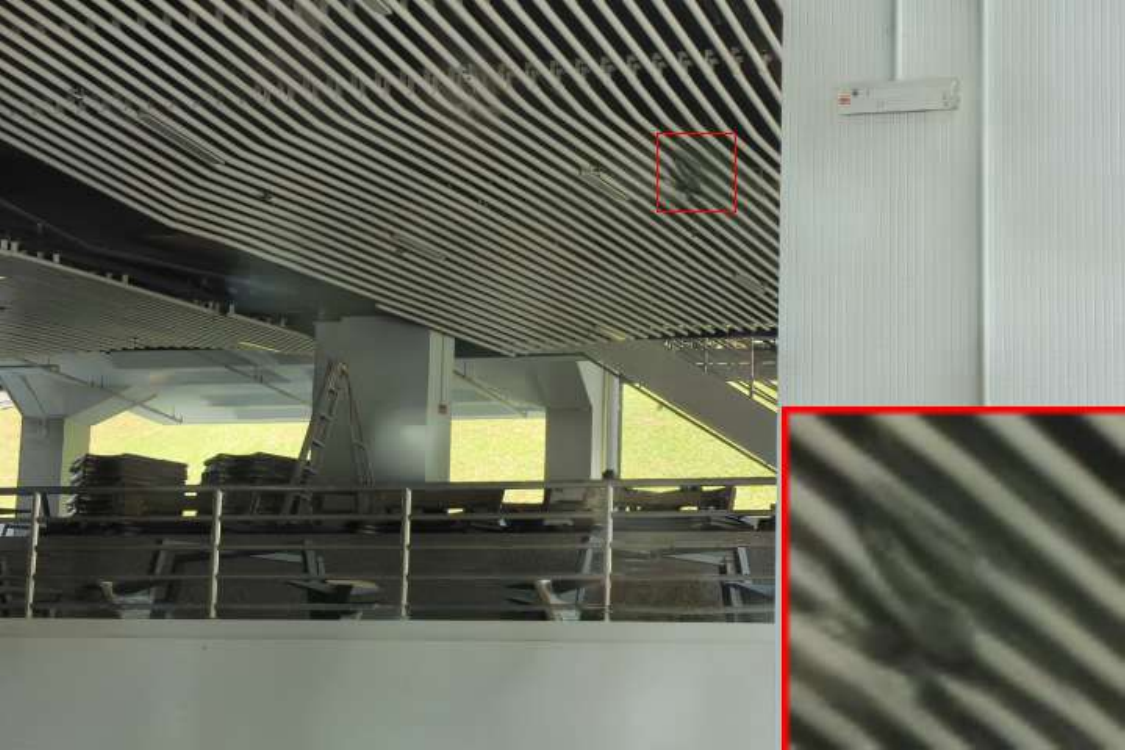} 
        \end{minipage}}
    \hfill
    \subfloat[UtilityIR (ours)]{\begin{minipage}[b][][b]{0.33\linewidth}
          \vspace*{\fill}
          \centering
          \includegraphics[width=1.0\linewidth]{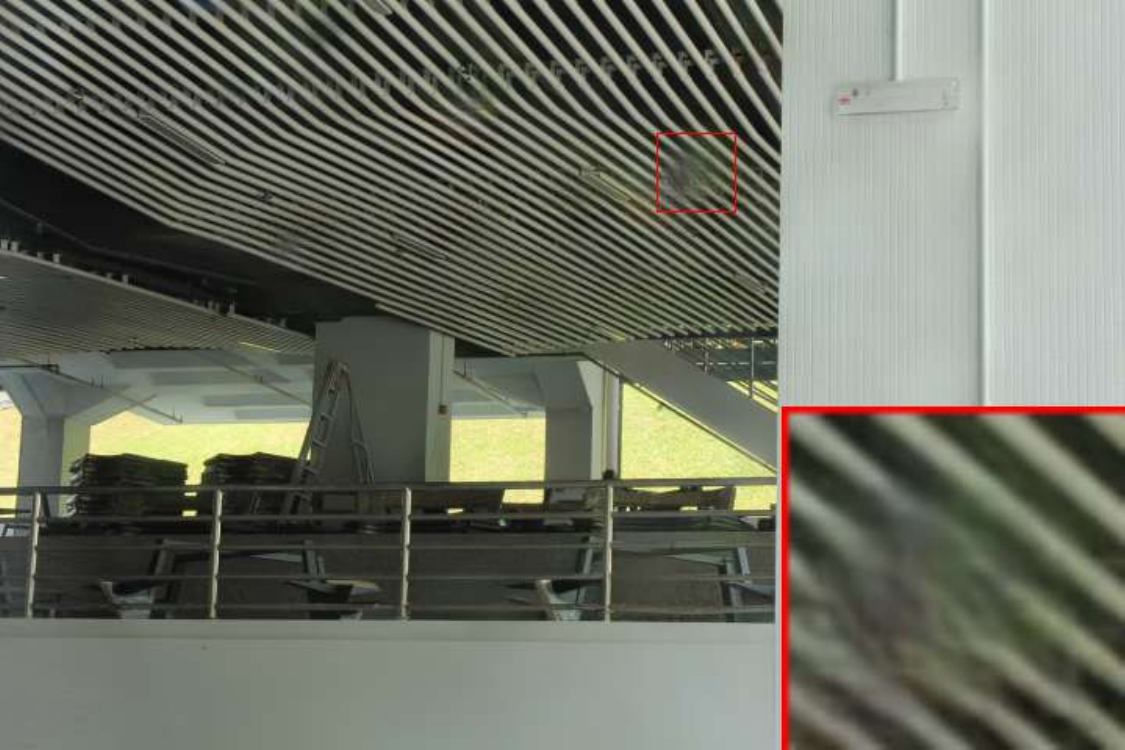} 
        \end{minipage}}
    \hfill
    \subfloat[GT]{\begin{minipage}[b][][b]{0.33\linewidth}
          \vspace*{\fill}
          \centering
          \includegraphics[width=1.0\linewidth]{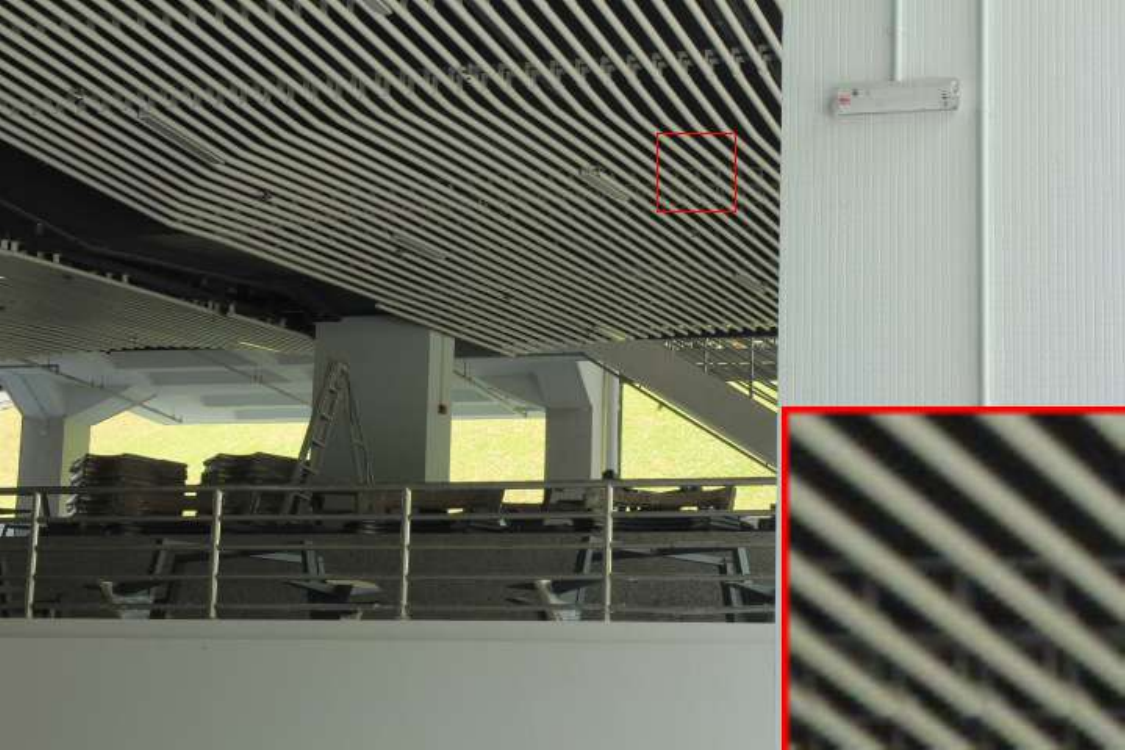} 
        \end{minipage}}
  \caption{Limitation of state-of-the-art methods. All of methods including ours result in blur artifact and fail to restore the high frequency detail with severely occlusion and degraded area.}
  \label{fig:exp_limitation}
  \vspace{-0.4cm}
\end{figure*}

As mentioned in body text, the state-of-the-art models including ours still suffer from blur artifact with severe occlusion on high frequency area as shown in Fig. \ref{fig:exp_limitation}, the same problem shared for almost all methods \cite{transweather, weatherdiff, wgwsnet}.


\begin{figure*}[htbp]
  \centering
    \includegraphics[width=1.0\linewidth]{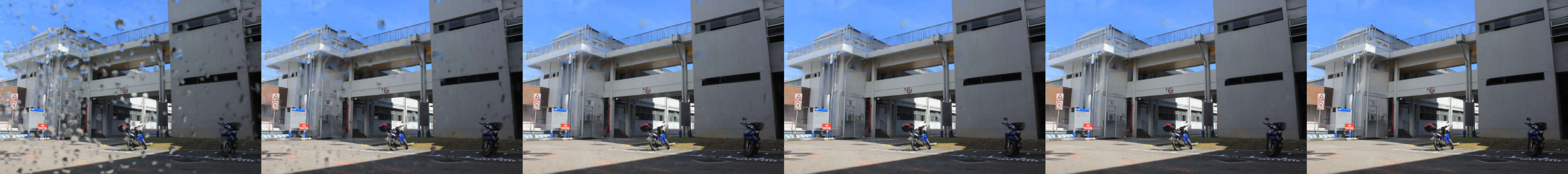}
    \includegraphics[width=1.0\linewidth]{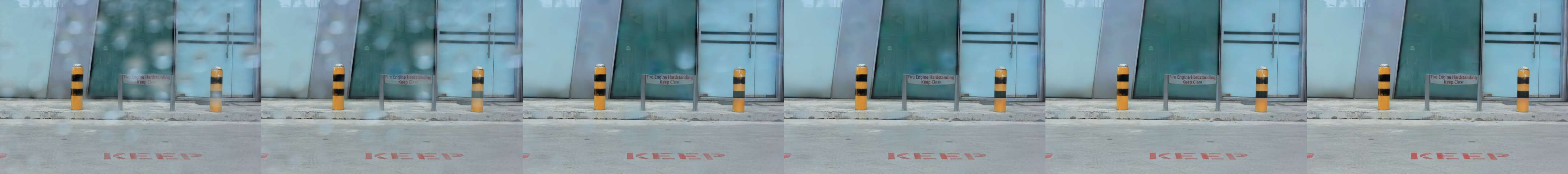}
    \includegraphics[width=1.0\linewidth]{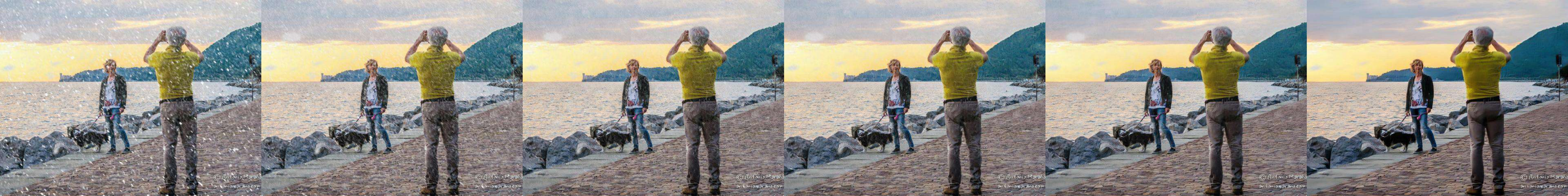}
    \includegraphics[width=1.0\linewidth]{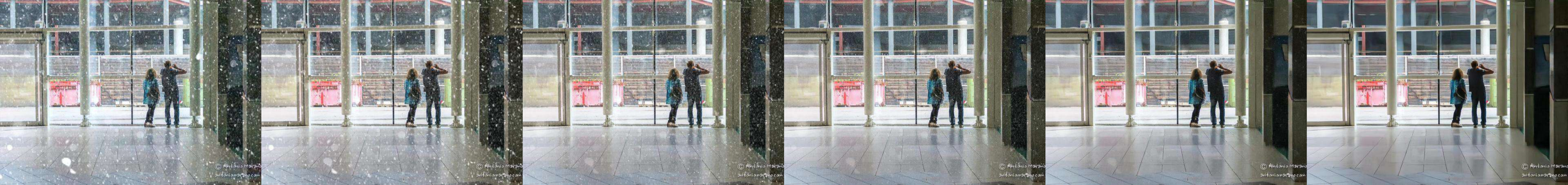}
    \includegraphics[width=1.0\linewidth]{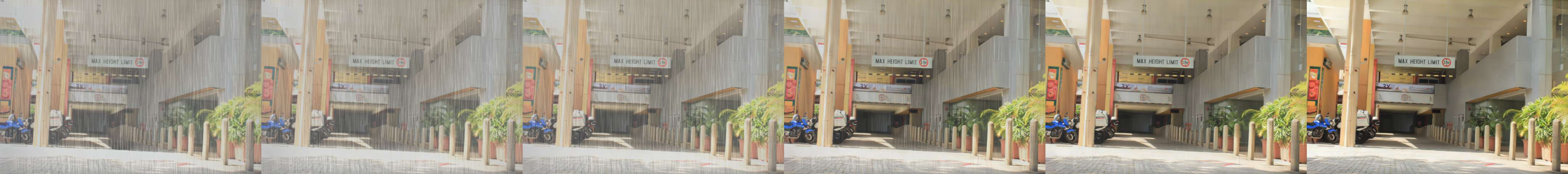}
    \includegraphics[width=1.0\linewidth]{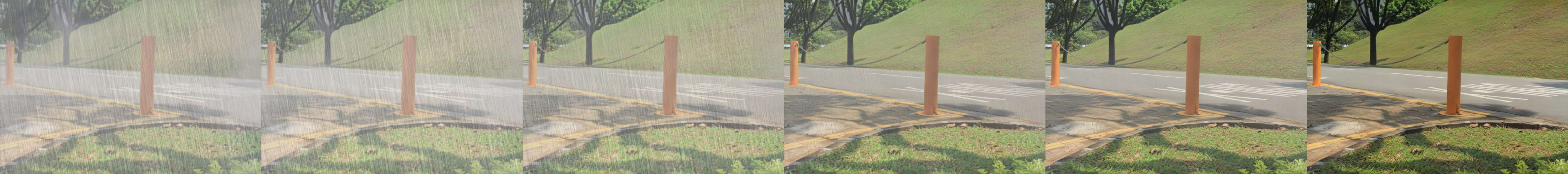}
  \caption{More restoration level modulation results}
   \label{fig:mani}
\end{figure*}
\begin{figure*}
     \centering
        \includegraphics[width=0.6\linewidth]{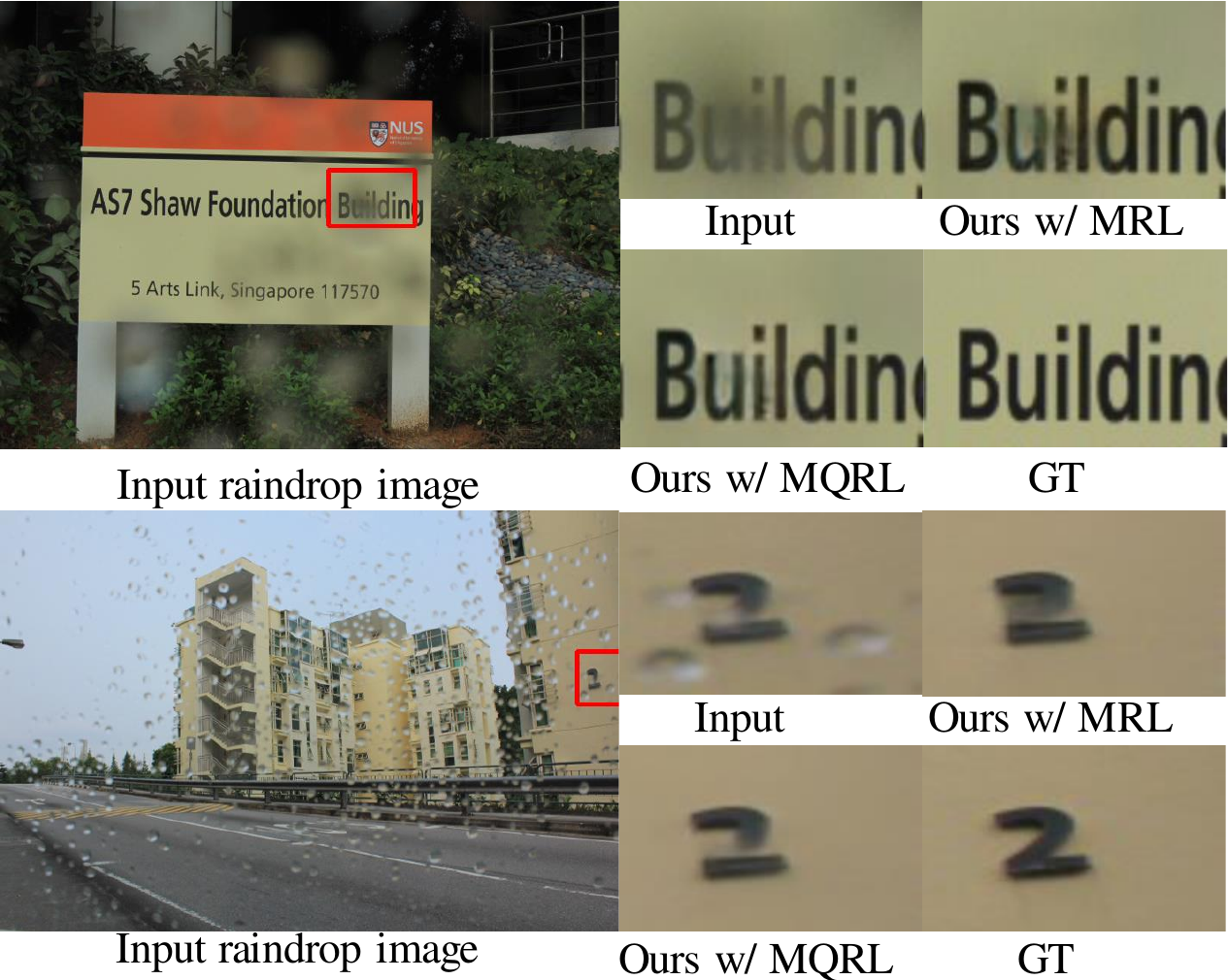}
     \caption{More visual results of the advantage of MQRL over MRL}
     \label{fig:compare}
\end{figure*}

\begin{figure*}[htpb!]
  \subfloat[Input image]{\begin{minipage}[b][3cm][b]{0.21\textwidth}
              \vspace*{\fill}
              \centering
              \includegraphics[width=1.0\linewidth]{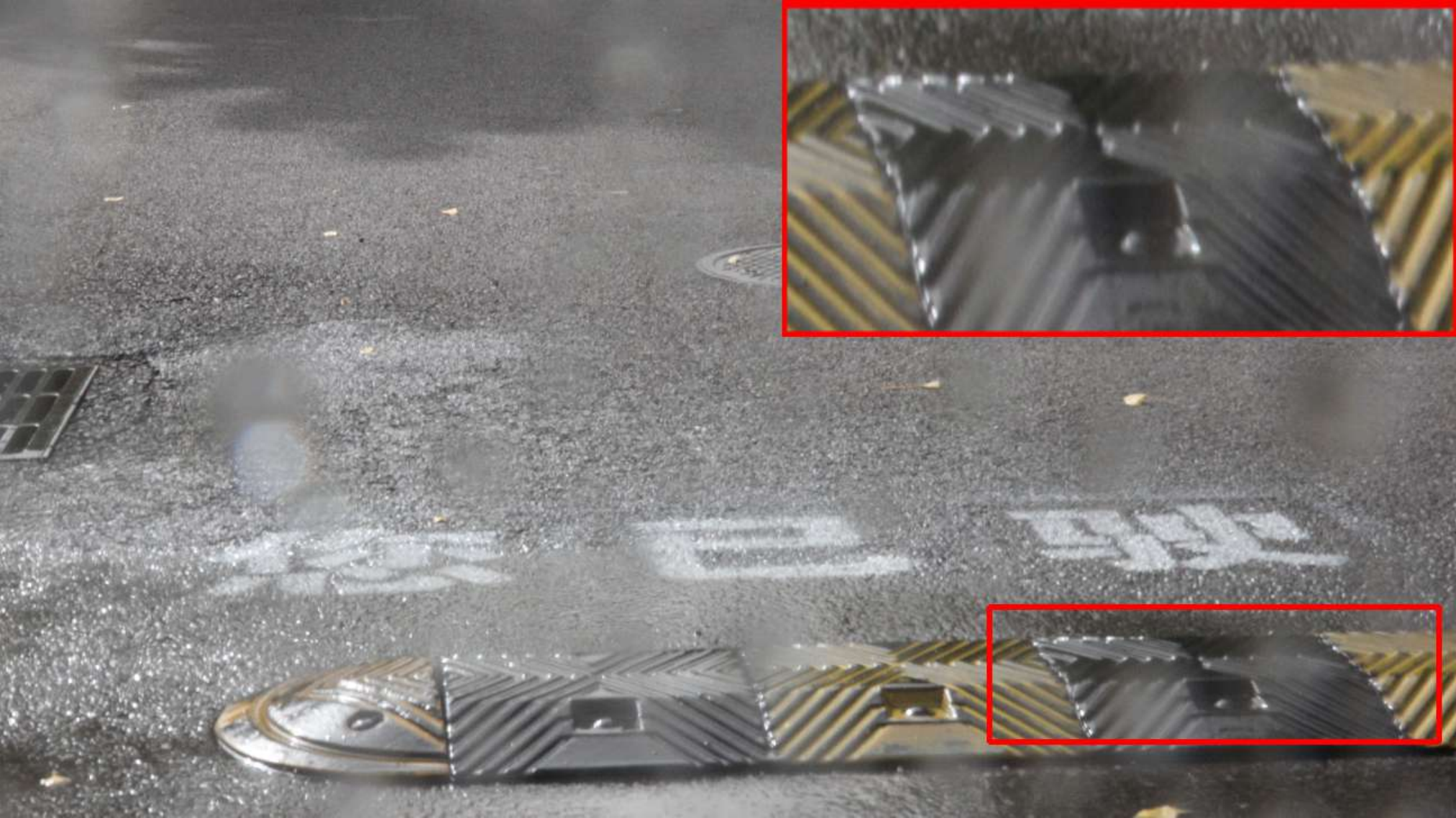} 
              \par\vspace{1mm}
              \includegraphics[width=1.0\linewidth]{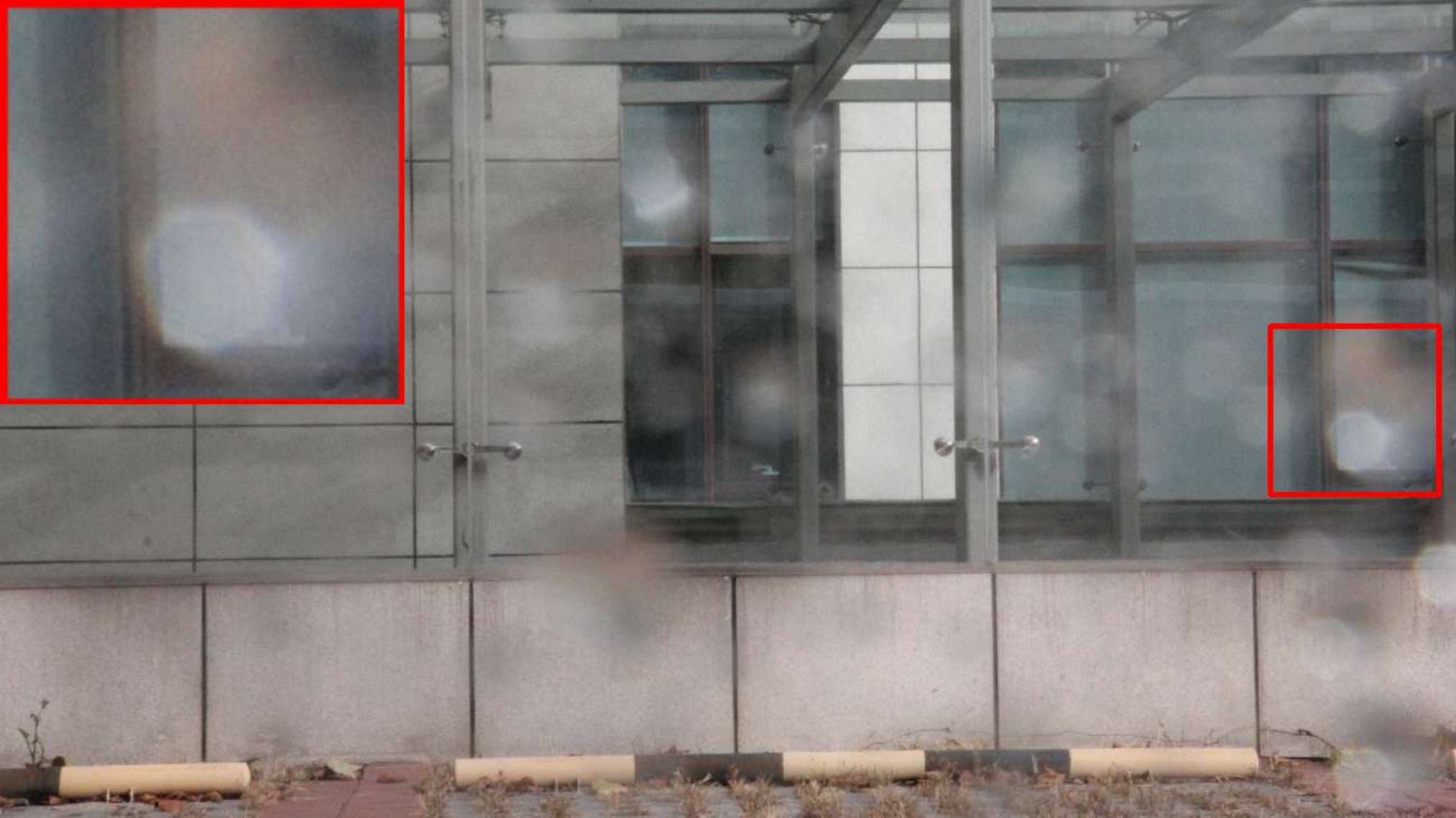} 
        \end{minipage}}
    \hfill
    \hspace{-0.25cm}
    \subfloat[Transweather]{\begin{minipage}[b][3cm][b]{0.21\textwidth}
          \vspace*{\fill}
          \centering
          \includegraphics[width=1.0\linewidth]{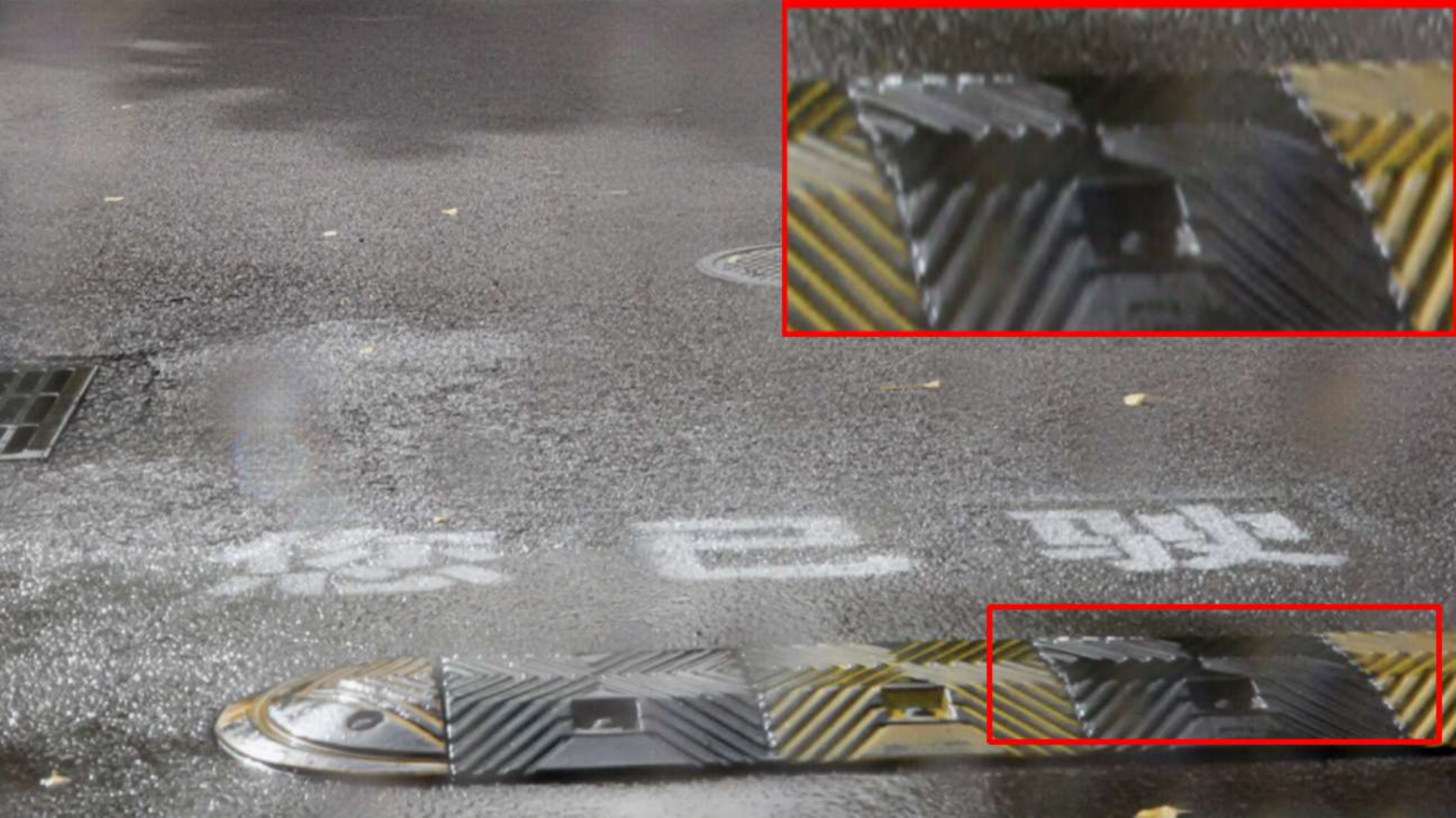} 
          \par\vspace{1mm}
          \includegraphics[width=1.0\linewidth]{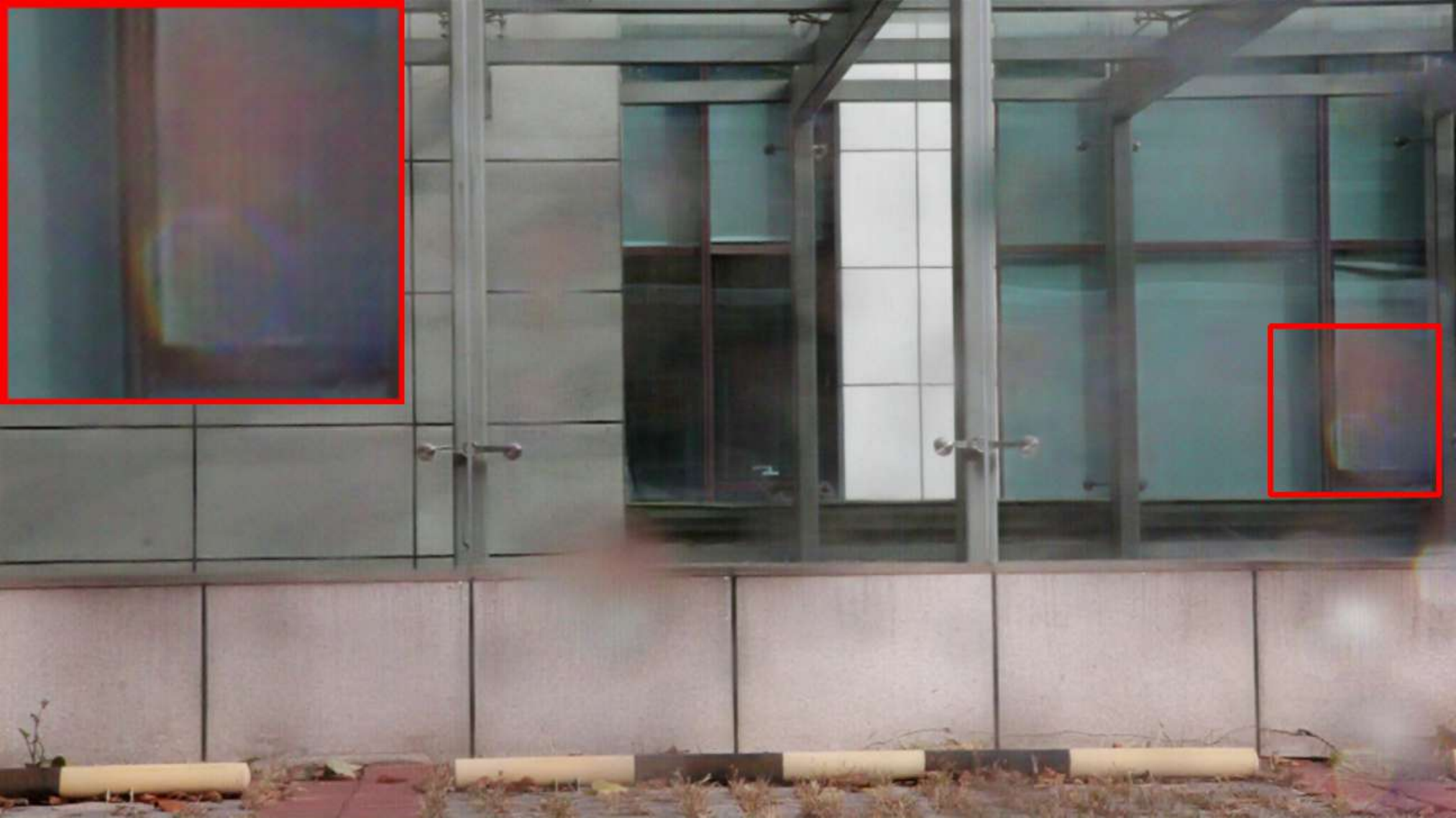} 
        \end{minipage}}
    \hfill
    \hspace{-0.25cm}
    \subfloat[Unified model]{\begin{minipage}[b][3cm][b]{0.21\textwidth}
          \vspace*{\fill}
          \centering
          \includegraphics[width=1.0\linewidth]{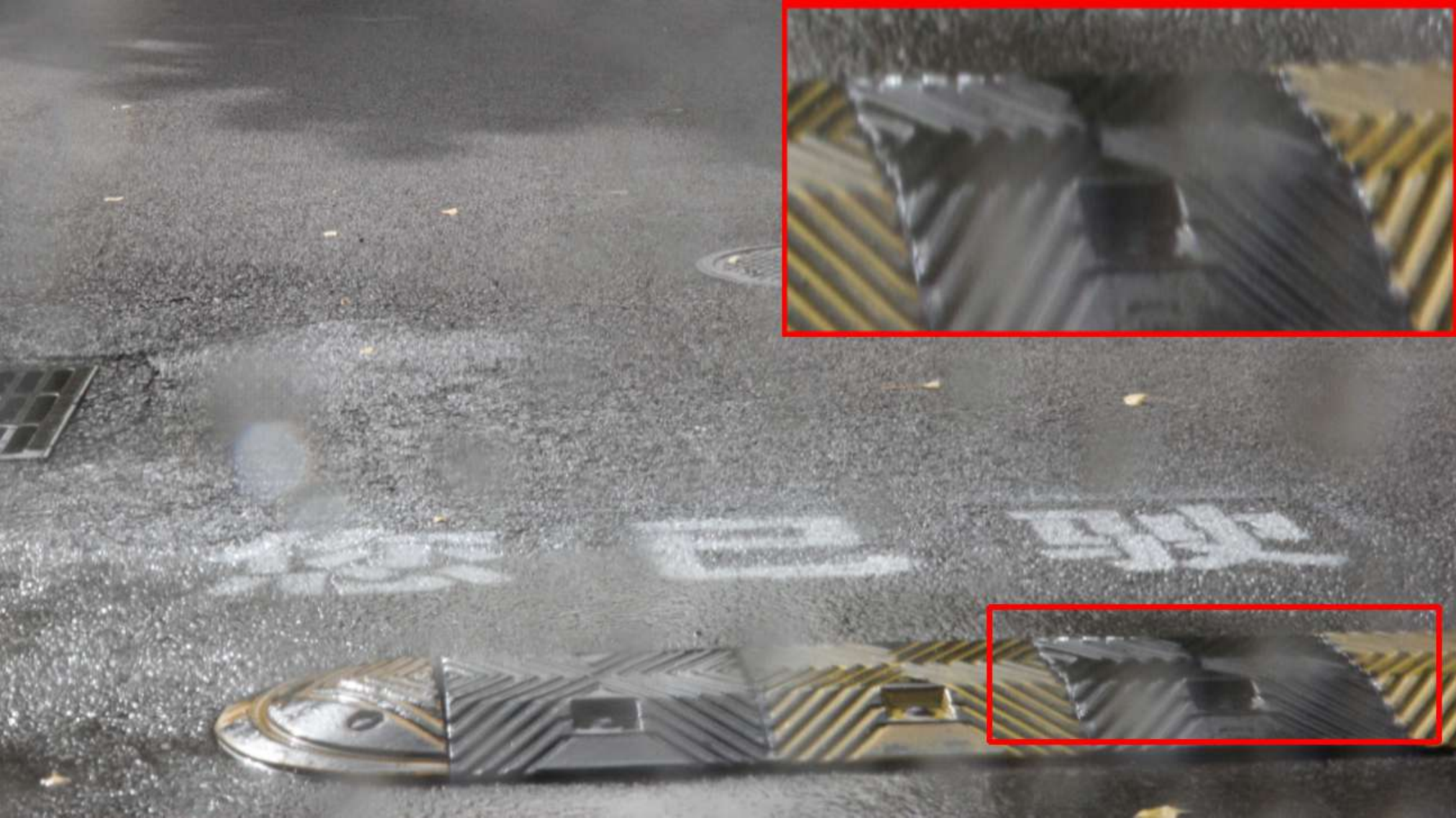} 
          \par\vspace{1mm}
          \includegraphics[width=1.0\linewidth]{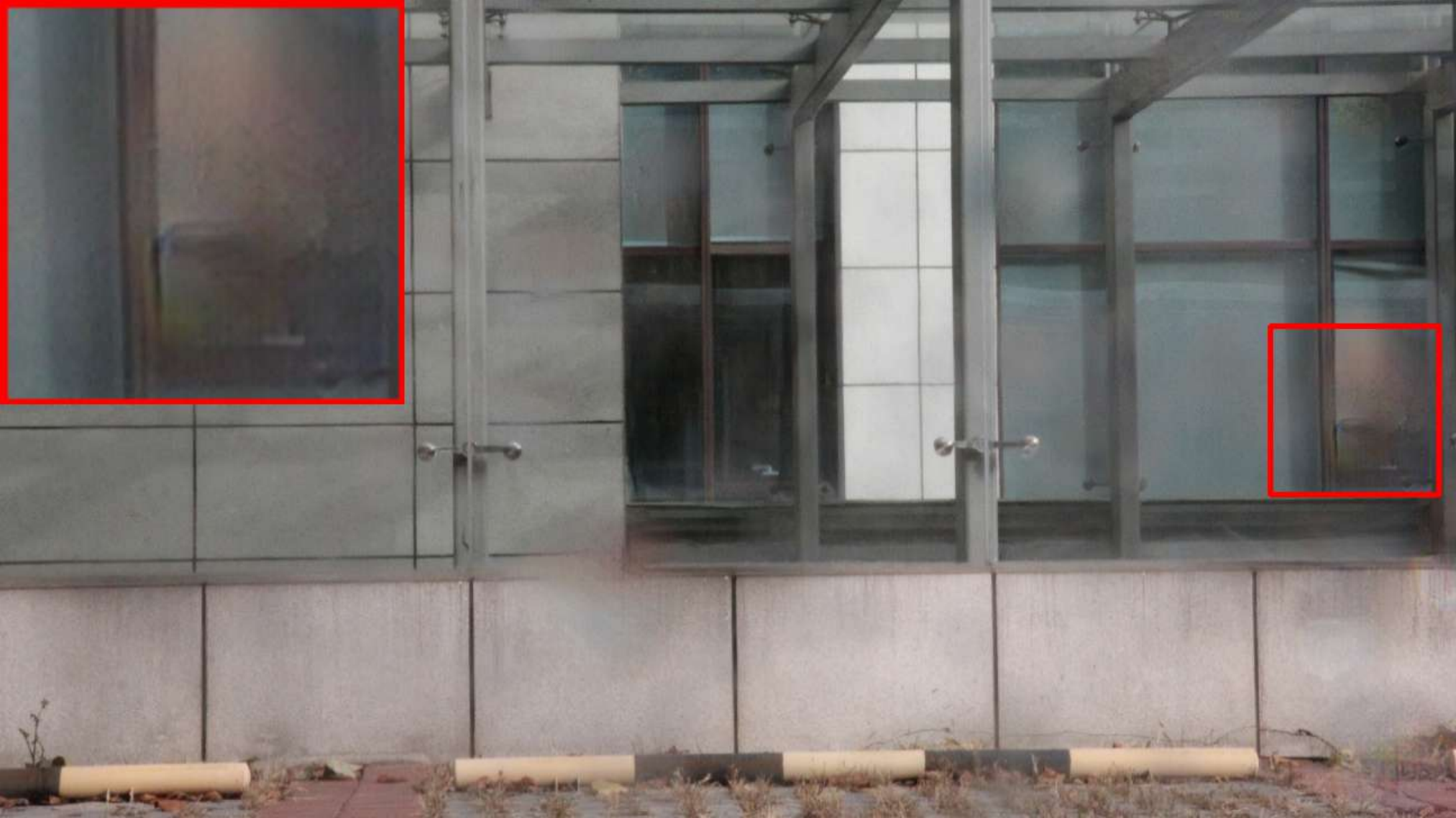} 
        \end{minipage}}
    \hfill
    \hspace{-0.25cm}
    \subfloat[WGWSNet]{\begin{minipage}[b][3cm][b]{0.21\textwidth}
          \vspace*{\fill}
          \centering
          \includegraphics[width=1.0\linewidth]{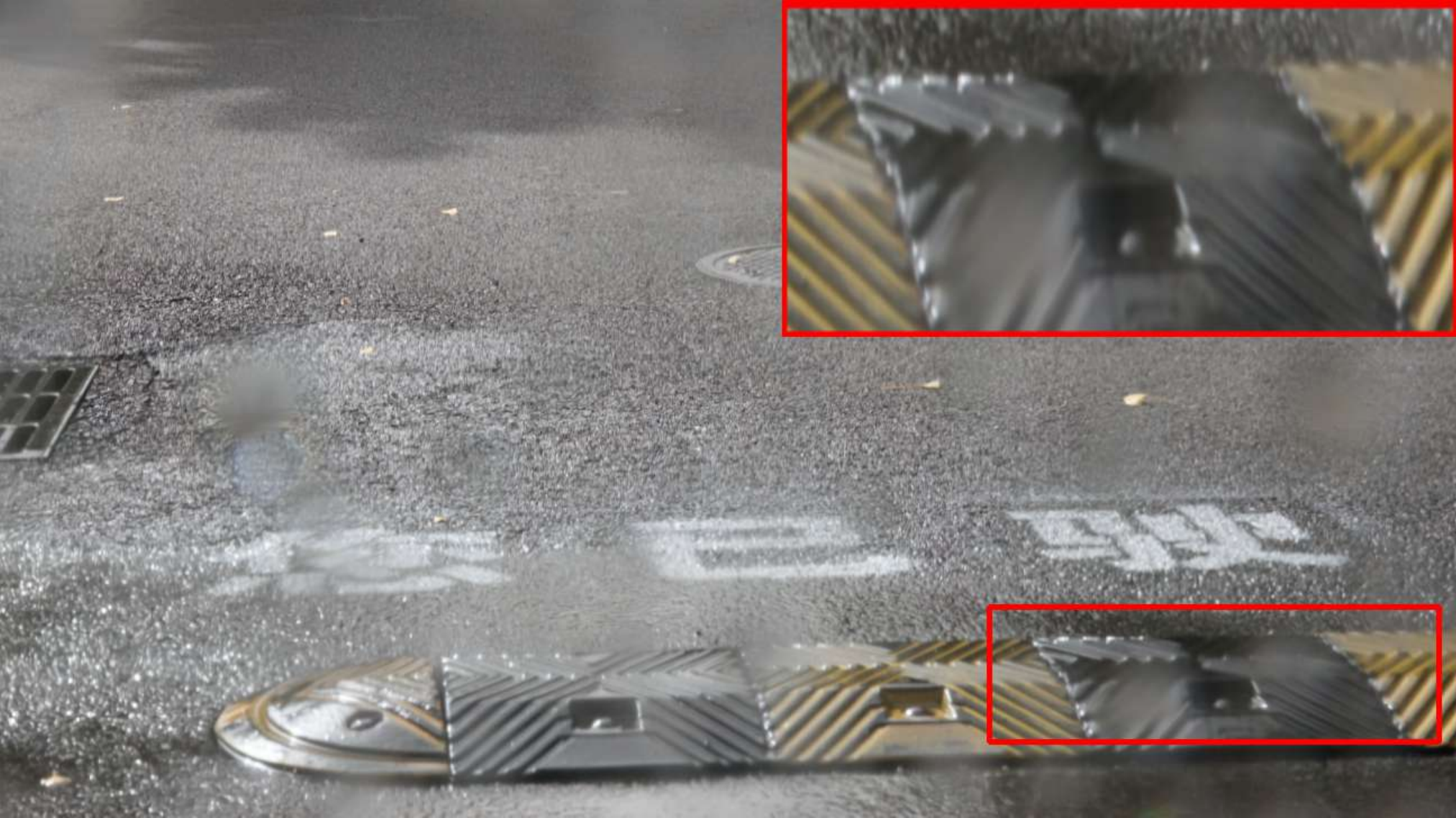} 
          \par\vspace{1mm}
          \includegraphics[width=1.0\linewidth]{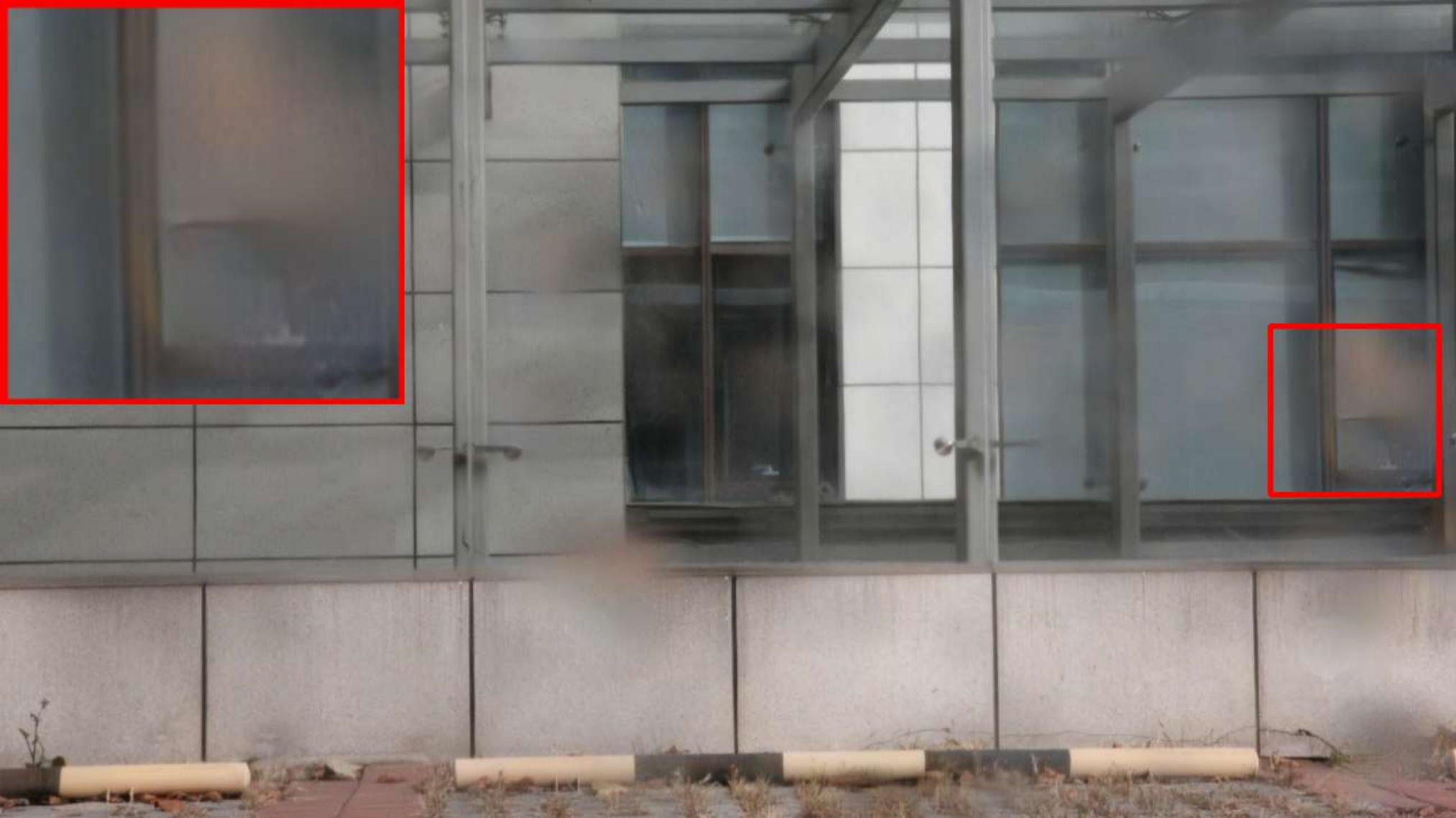} 
        \end{minipage}}
    \hfill
    \hspace{-0.25cm}
    \subfloat[UtilityIR(ours)]{\begin{minipage}[b][3cm][b]{0.21\textwidth}
          \vspace*{\fill}
          \centering
          \includegraphics[width=1.0\linewidth]{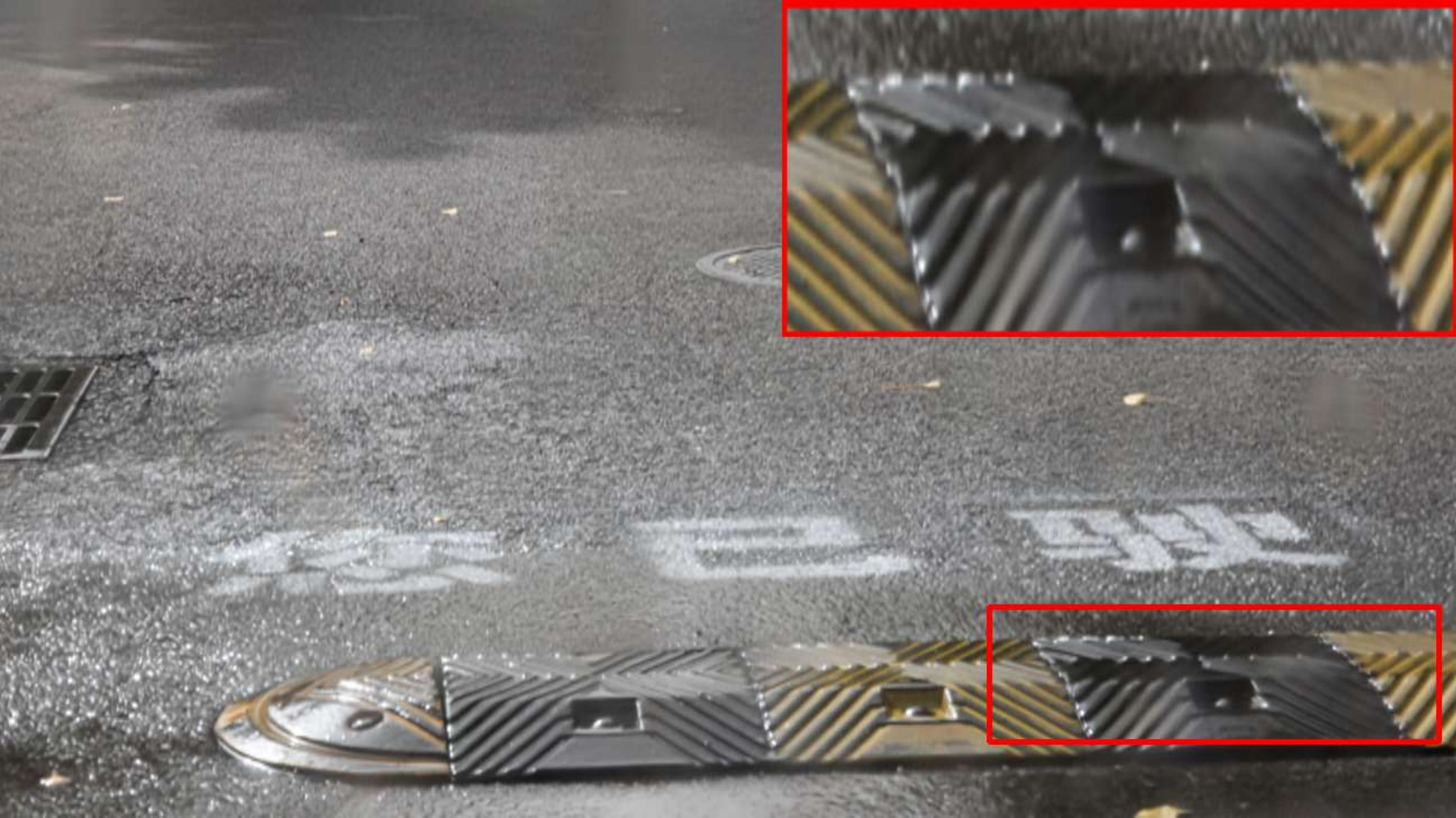} 
          \par\vspace{1mm}
          \includegraphics[width=1.0\linewidth]{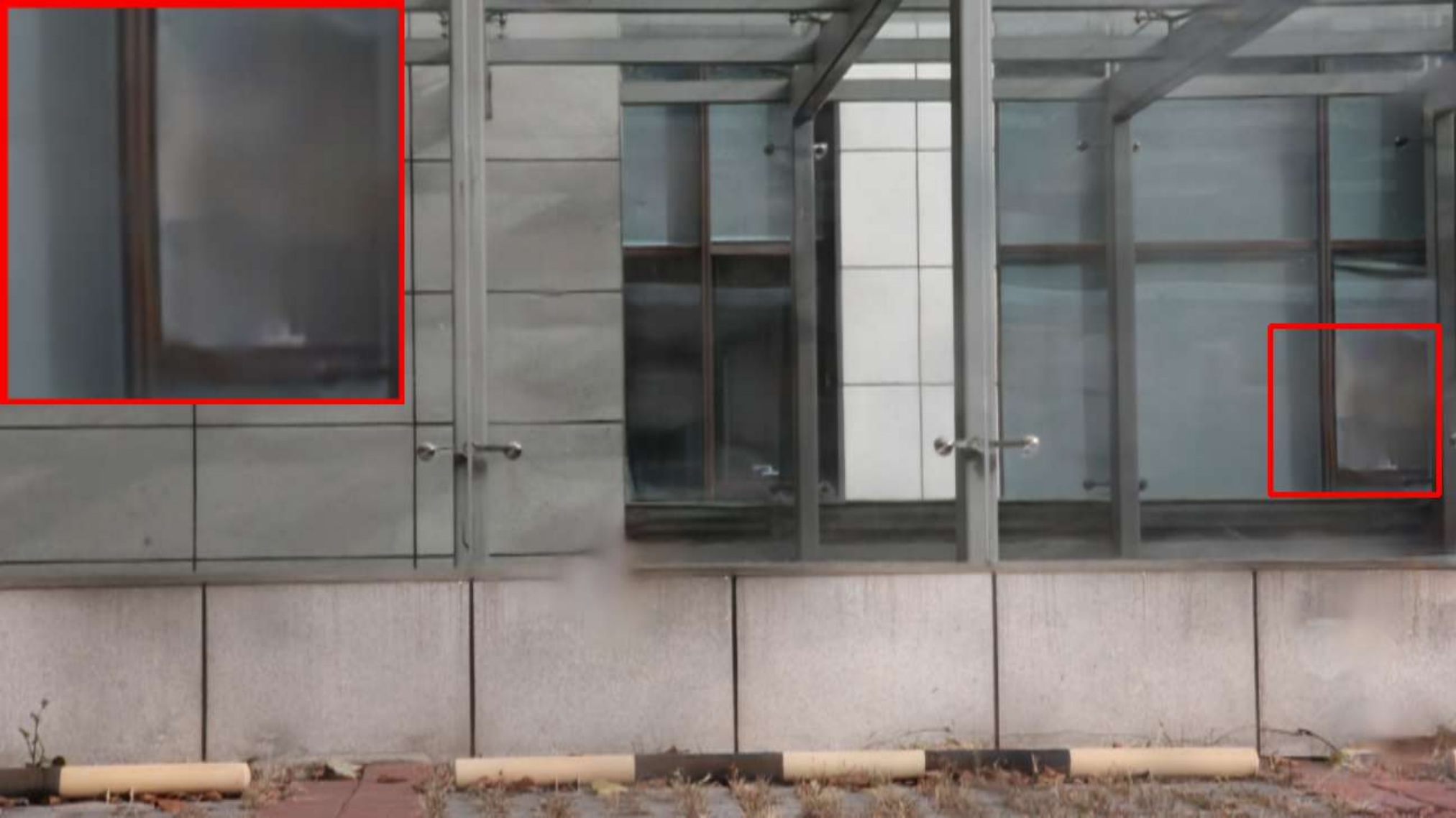} 
        \end{minipage}}
  \caption{Visual results of real-world raindrop images. Zoom in for best view.}
  \label{fig:exp_real_rd}
\end{figure*}
\begin{figure*}[htpb!]
  \subfloat[Input image]{\begin{minipage}[b][3cm][b]{0.21\textwidth}
              \vspace*{\fill}
              \centering
              \includegraphics[width=1.0\linewidth]{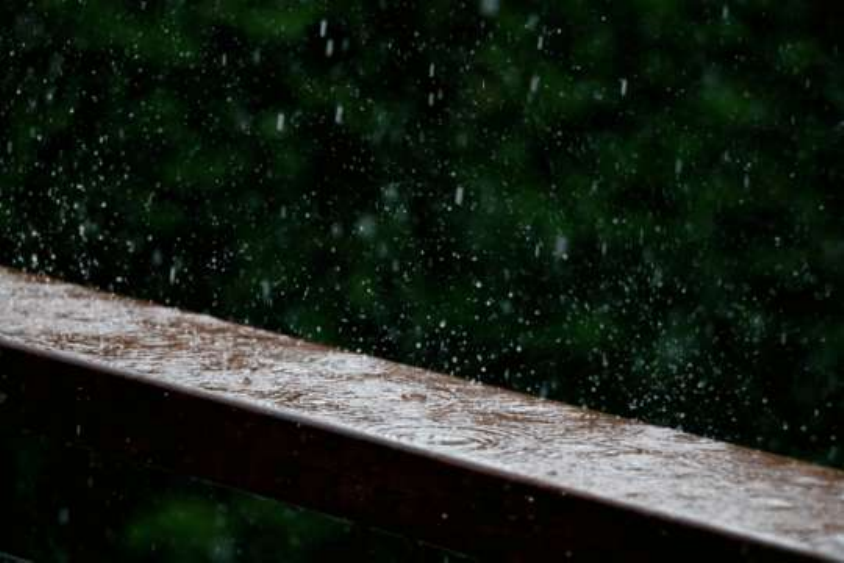} 
              \par\vspace{1mm}
              \includegraphics[width=1.0\linewidth]{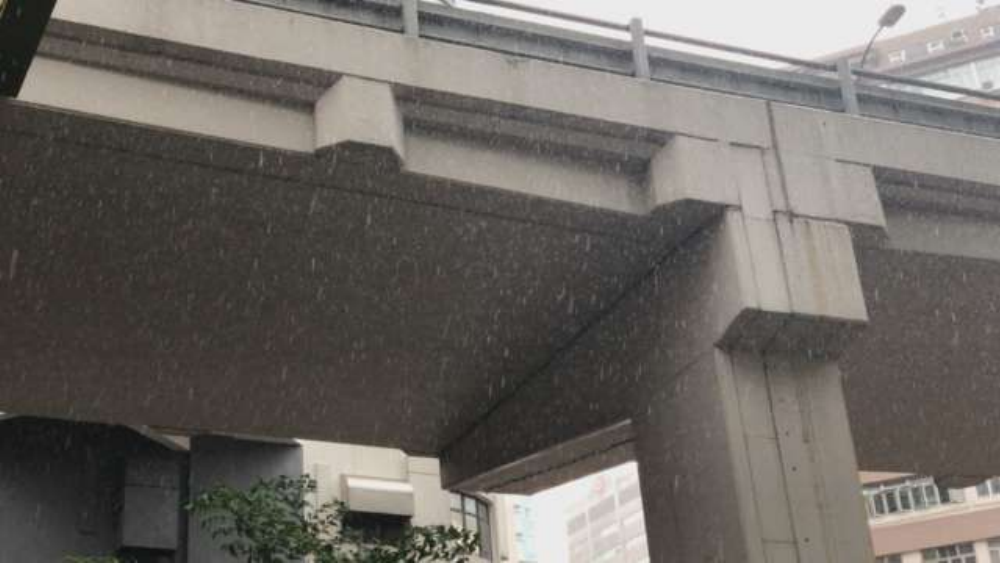} 
        \end{minipage}}
    \hfill
    \hspace{-0.25cm}
    \subfloat[Transweather]{\begin{minipage}[b][3cm][b]{0.21\textwidth}
          \vspace*{\fill}
          \centering
          \includegraphics[width=1.0\linewidth]{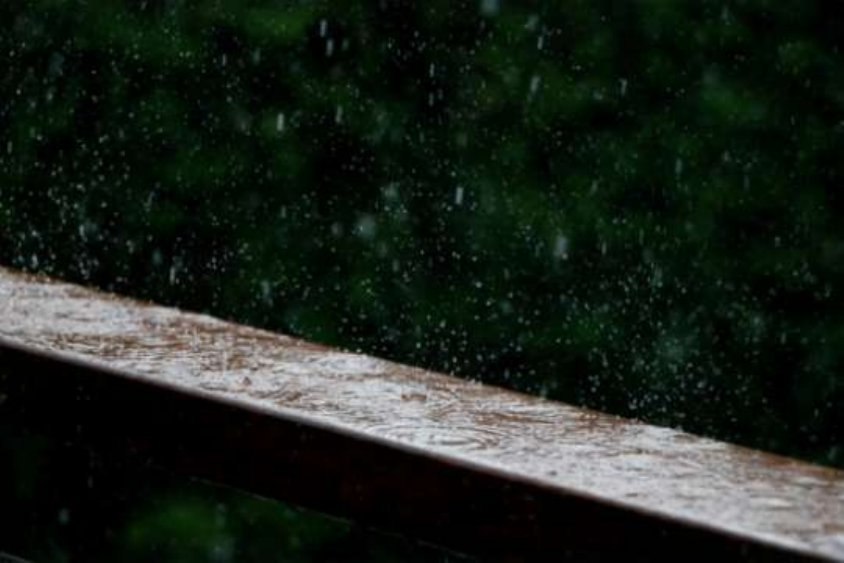} 
          \par\vspace{1mm}
          \includegraphics[width=1.0\linewidth]{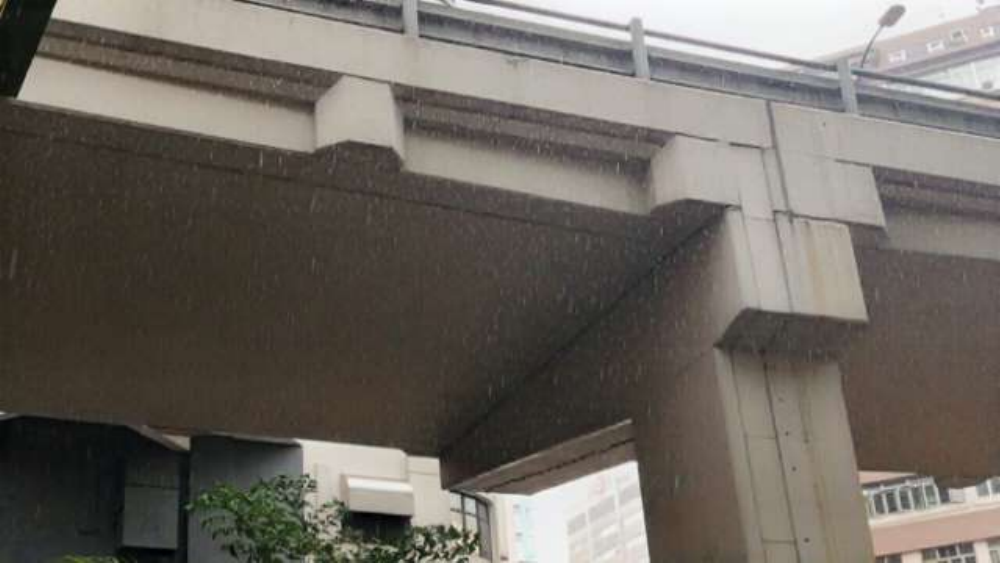} 
        \end{minipage}}
    \hfill
    \hspace{-0.25cm}
    \subfloat[Unified model]{\begin{minipage}[b][3cm][b]{0.21\textwidth}
          \vspace*{\fill}
          \centering
          \includegraphics[width=1.0\linewidth]{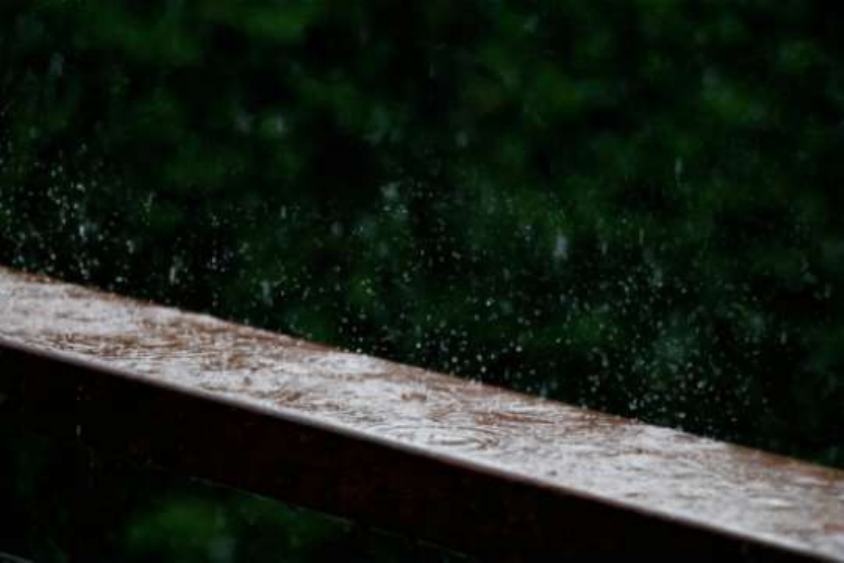} 
          \par\vspace{1mm}
          \includegraphics[width=1.0\linewidth]{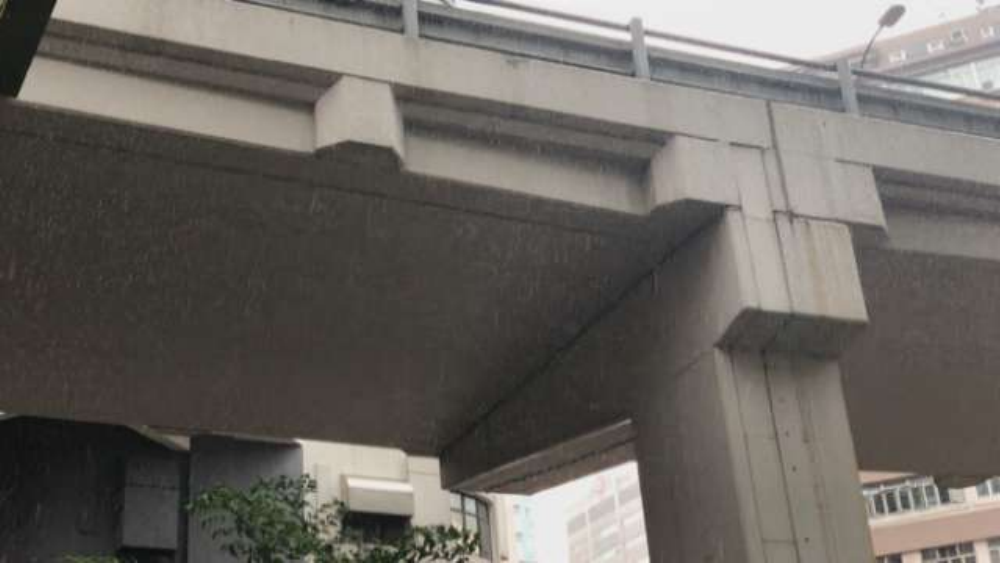} 
        \end{minipage}}
    \hfill
    \hspace{-0.25cm}
    \subfloat[WGWSNet]{\begin{minipage}[b][3cm][b]{0.21\textwidth}
          \vspace*{\fill}
          \centering
          \includegraphics[width=1.0\linewidth]{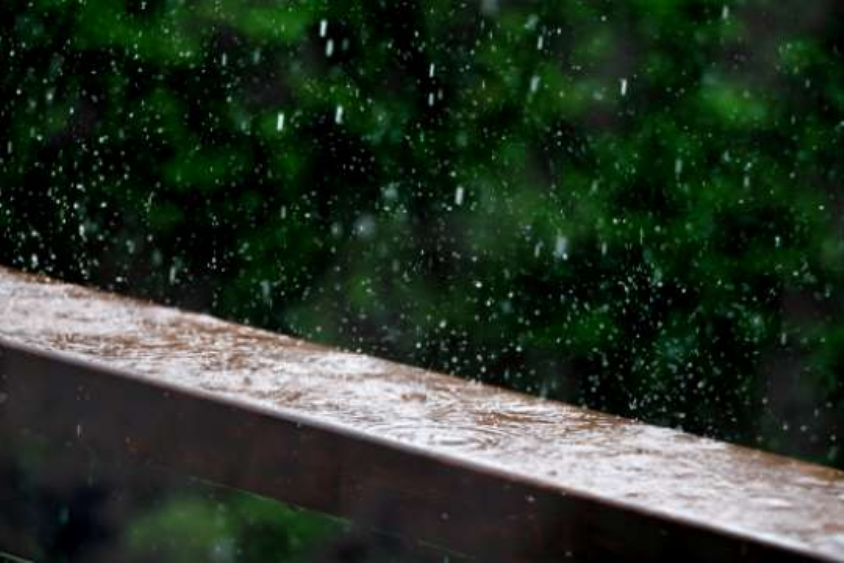} 
          \par\vspace{1mm}
          \includegraphics[width=1.0\linewidth]{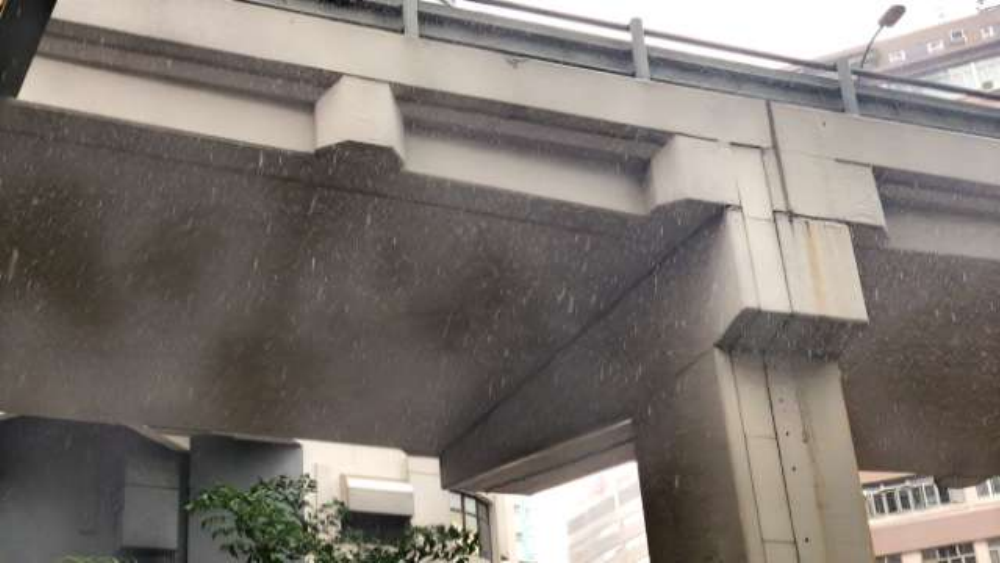} 
        \end{minipage}}
    \hfill
    \hspace{-0.25cm}
    \subfloat[UtilityIR(ours)]{\begin{minipage}[b][3cm][b]{0.21\textwidth}
          \vspace*{\fill}
          \centering
          \includegraphics[width=1.0\linewidth]{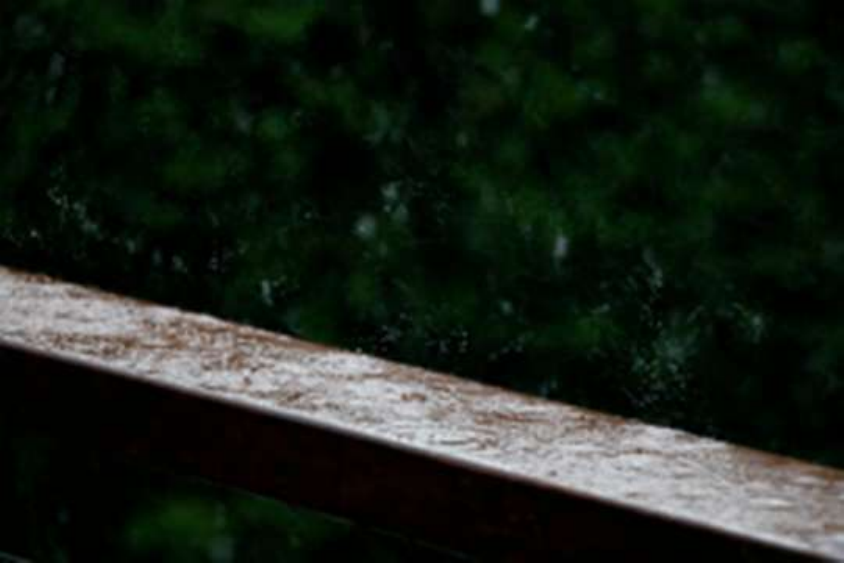} 
          \par\vspace{1mm}
          \includegraphics[width=1.0\linewidth]{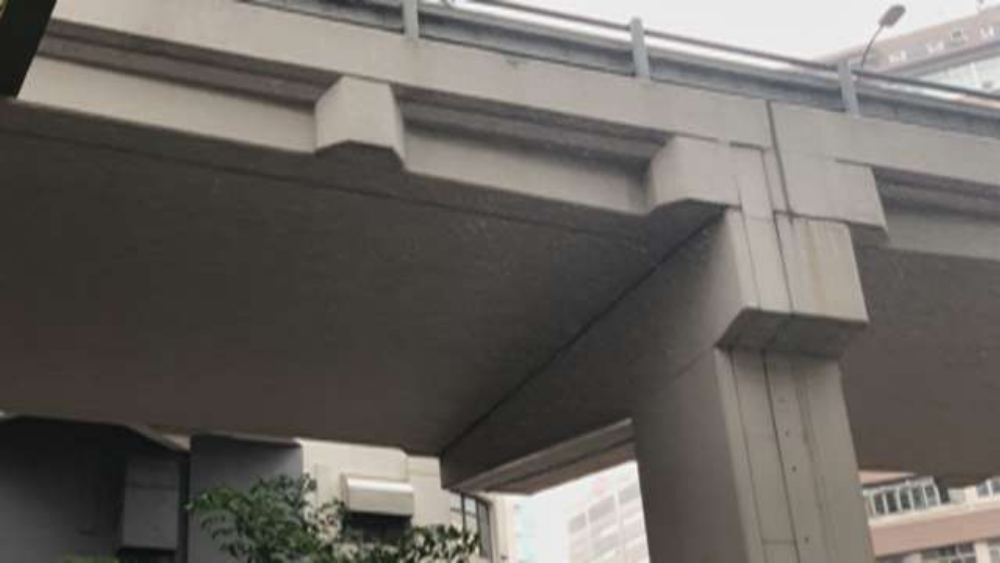} 
        \end{minipage}}
  \caption{Visual result of real-world rainy images. Zoom in for best view.}
  \label{fig:exp_real_rs}
\end{figure*}
\begin{figure*}[htpb!]
  \subfloat[Input image]{\begin{minipage}[b][3cm][b]{0.21\textwidth}
              \vspace*{\fill}
              \centering
              \includegraphics[width=1.0\linewidth]{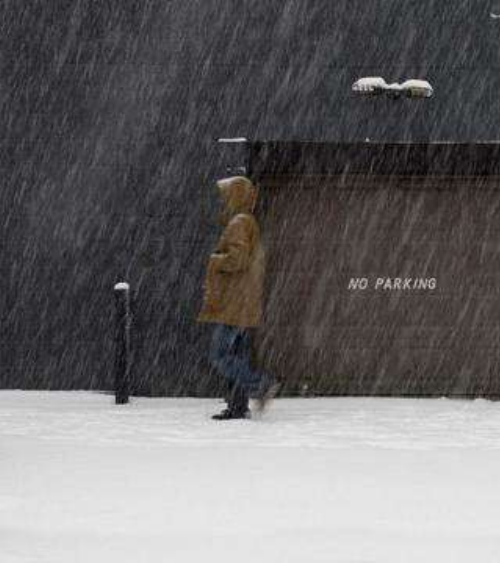} 
              \par\vspace{1mm}
              \includegraphics[width=1.0\linewidth]{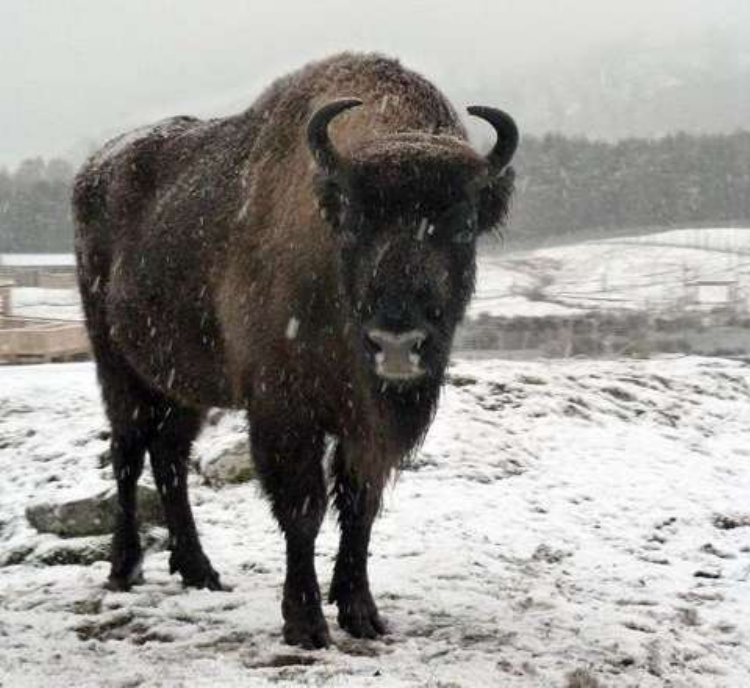} 
        \end{minipage}}
    \hfill
    \hspace{-0.25cm}
    \subfloat[Transweather]{\begin{minipage}[b][3cm][b]{0.21\textwidth}
          \vspace*{\fill}
          \centering
          \includegraphics[width=1.0\linewidth]{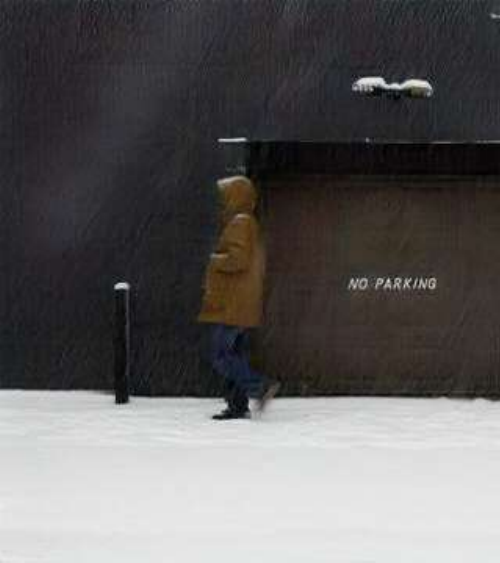} 
          \par\vspace{1mm}
          \includegraphics[width=1.0\linewidth]{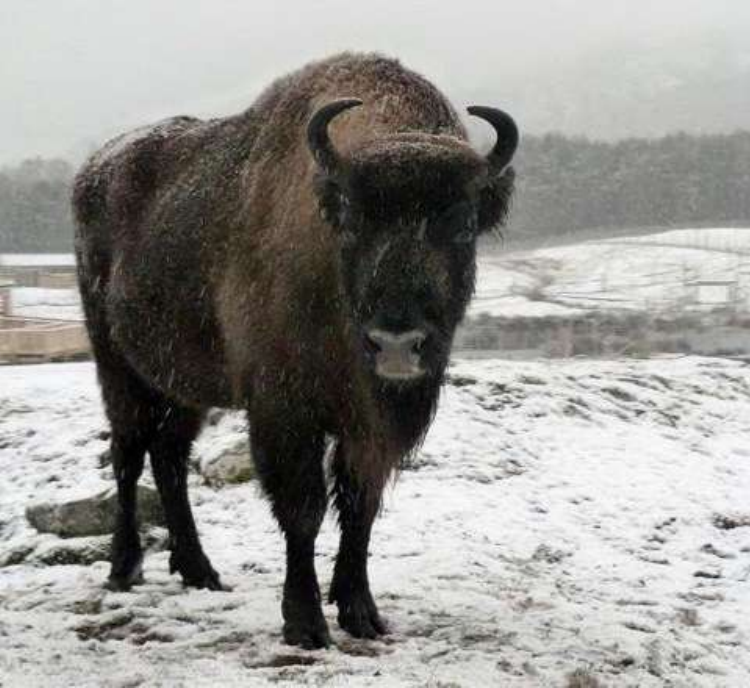} 
        \end{minipage}}
    \hfill
    \hspace{-0.25cm}
    \subfloat[Unified model]{\begin{minipage}[b][3cm][b]{0.21\textwidth}
          \vspace*{\fill}
          \centering
          \includegraphics[width=1.0\linewidth]{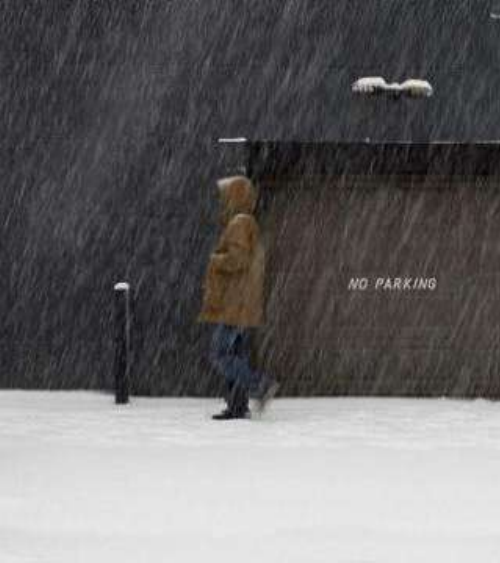} 
          \par\vspace{1mm}
          \includegraphics[width=1.0\linewidth]{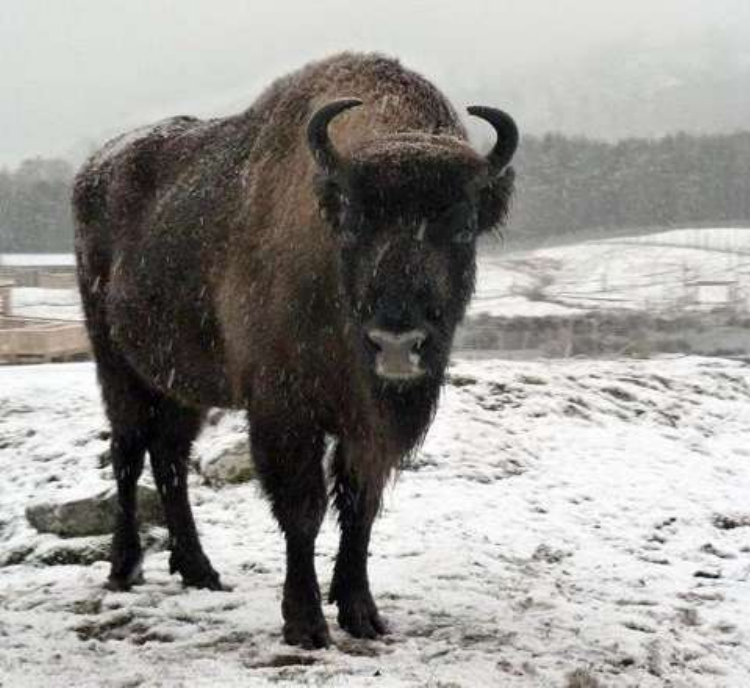} 
        \end{minipage}}
    \hfill
    \hspace{-0.25cm}
    \subfloat[WGWSNet]{\begin{minipage}[b][3cm][b]{0.21\textwidth}
          \vspace*{\fill}
          \centering
          \includegraphics[width=1.0\linewidth]{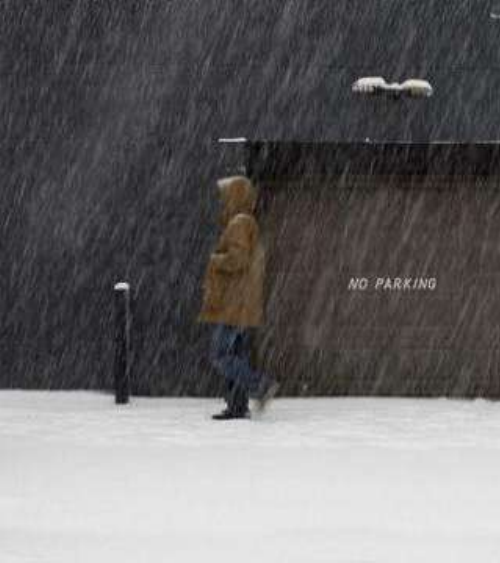} 
          \par\vspace{1mm}
          \includegraphics[width=1.0\linewidth]{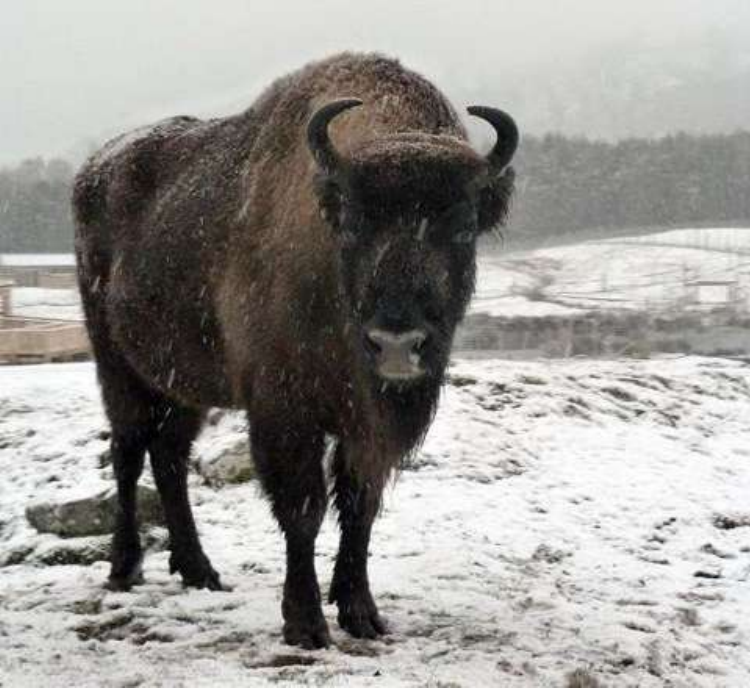} 
        \end{minipage}}
    \hfill
    \hspace{-0.25cm}
    \subfloat[UtilityIR(ours)]{\begin{minipage}[b][3cm][b]{0.21\textwidth}
          \vspace*{\fill}
          \centering
          \includegraphics[width=1.0\linewidth]{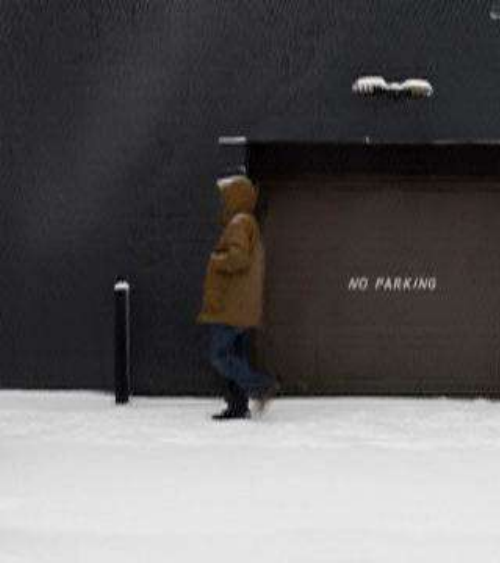} 
          \par\vspace{1mm}
          \includegraphics[width=1.0\linewidth]{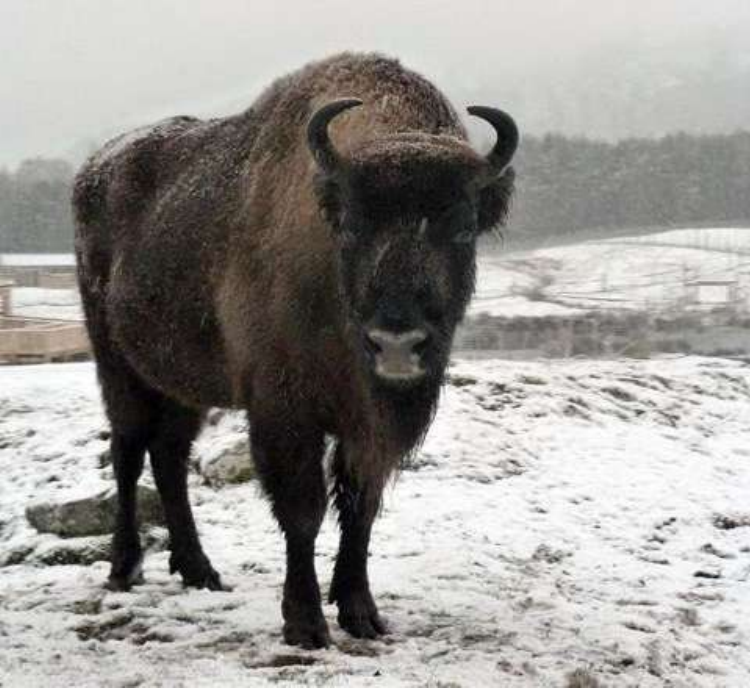} 
        \end{minipage}}
  \caption{Visual result of real-world snow images. Zoom in for best view.}
  \label{fig:exp_real_sw}
\end{figure*}

\section{More Visual Results}
\begin{figure*}[htpb!]
  \subfloat[Input image]{\begin{minipage}[b][3cm][b]{0.21\textwidth}
              \vspace*{\fill}
              \centering
              \includegraphics[width=1.0\linewidth]{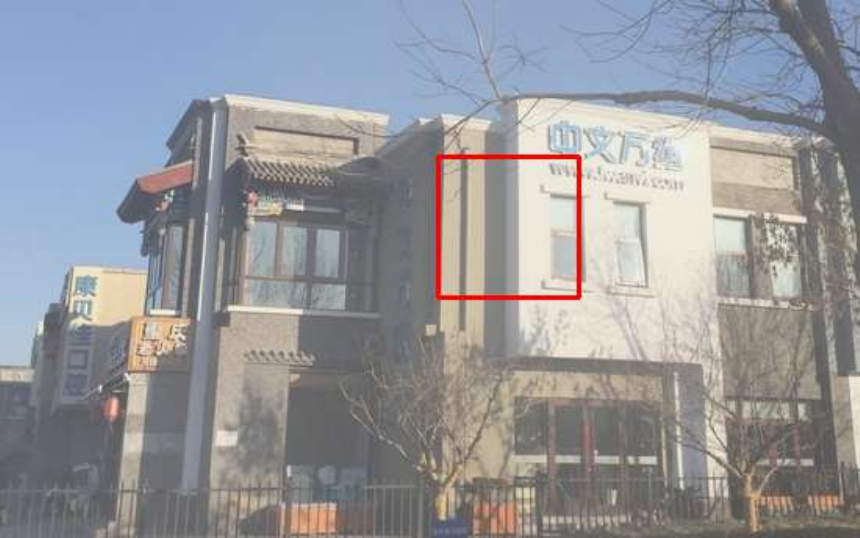} 
              \par\vspace{1mm}
              \includegraphics[width=1.0\linewidth]{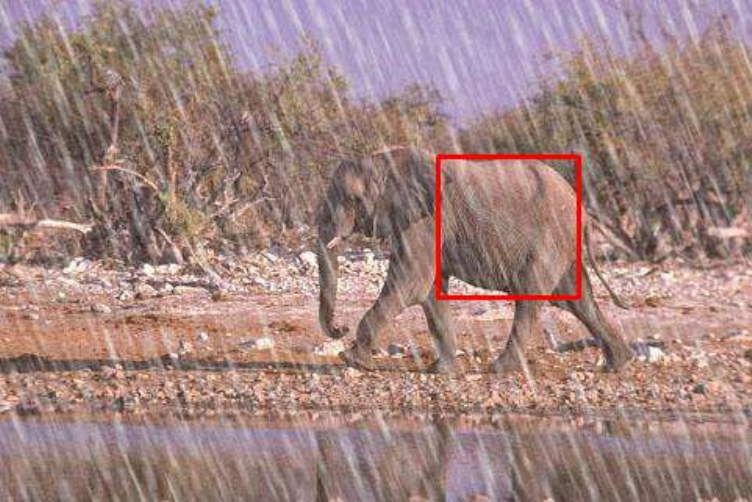} 
        \end{minipage}}
    \hfill
    \hspace{-0.25cm}
    \subfloat[Transweather]{\begin{minipage}[b][3cm][b]{0.21\textwidth}
          \vspace*{\fill}
          \centering
          \includegraphics[width=1.0\linewidth]{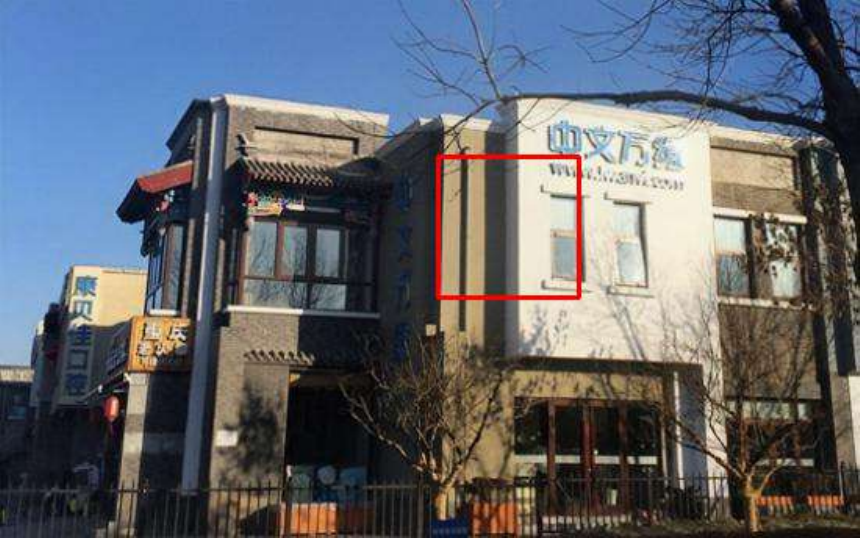} 
          \par\vspace{1mm}
          \includegraphics[width=1.0\linewidth]{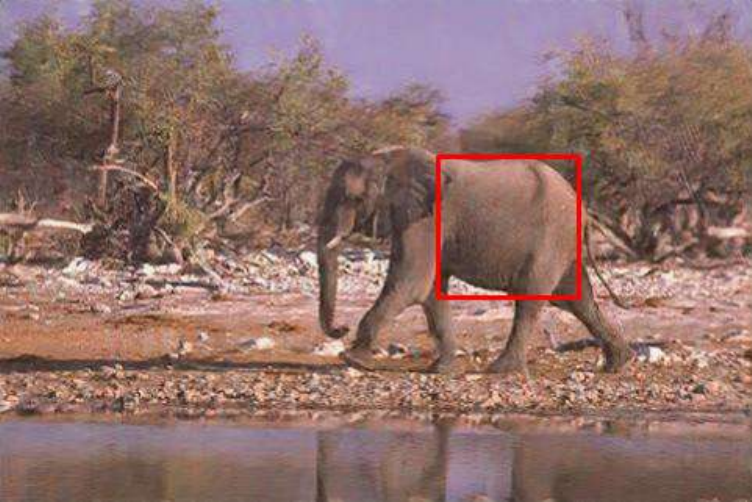} 
        \end{minipage}}
    \hfill
    \hspace{-0.25cm}
    \subfloat[Unified model]{\begin{minipage}[b][3cm][b]{0.21\textwidth}
          \vspace*{\fill}
          \centering
          \includegraphics[width=1.0\linewidth]{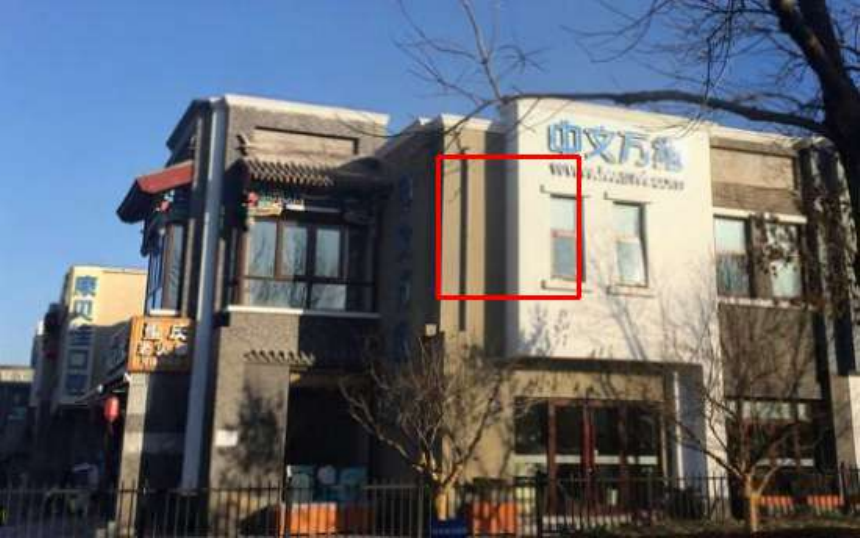} 
          \par\vspace{1mm}
          \includegraphics[width=1.0\linewidth]{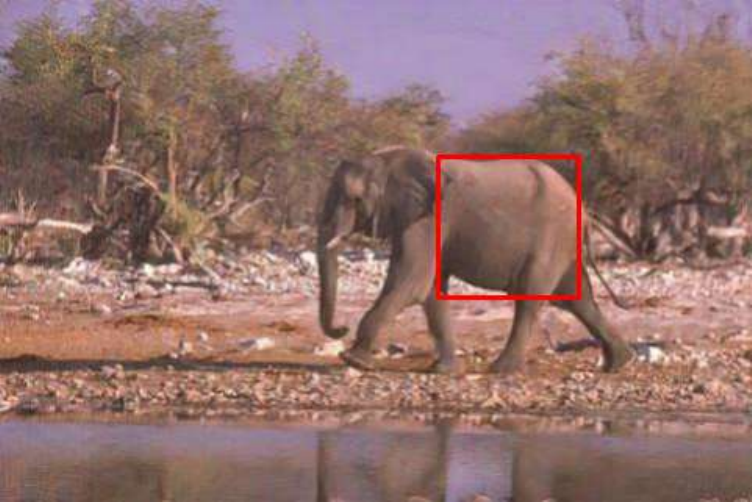} 
        \end{minipage}}
    \hfill
    \hspace{-0.25cm}
    \subfloat[WGWSNet]{\begin{minipage}[b][3cm][b]{0.21\textwidth}
          \vspace*{\fill}
          \centering
          \includegraphics[width=1.0\linewidth]{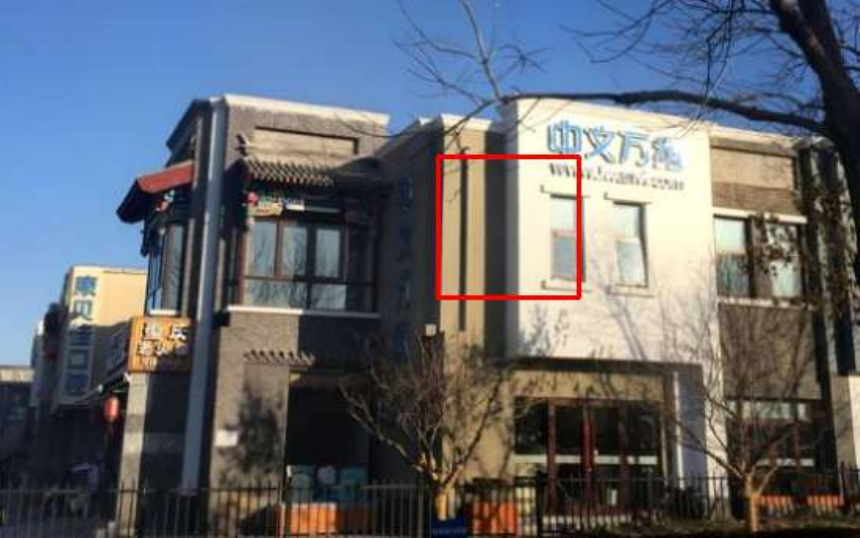} 
          \par\vspace{1mm}
          \includegraphics[width=1.0\linewidth]{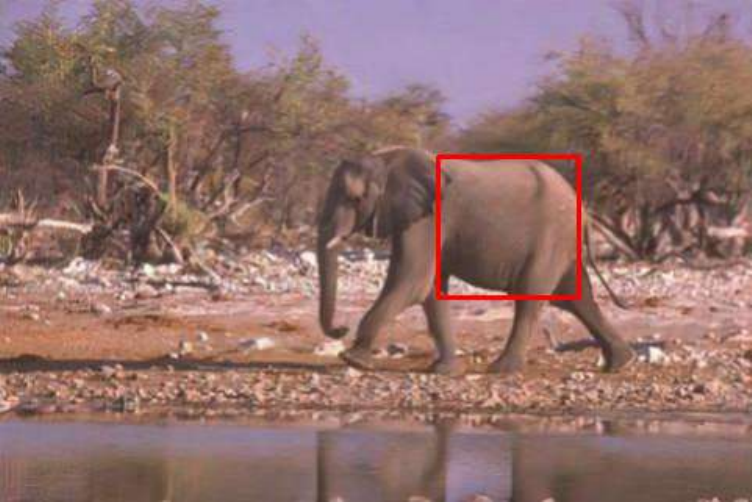} 
        \end{minipage}}
    \hfill
    \hspace{-0.25cm}
    \subfloat[UtilityIR(ours)]{\begin{minipage}[b][3cm][b]{0.21\textwidth}
          \vspace*{\fill}
          \centering
          \includegraphics[width=1.0\linewidth]{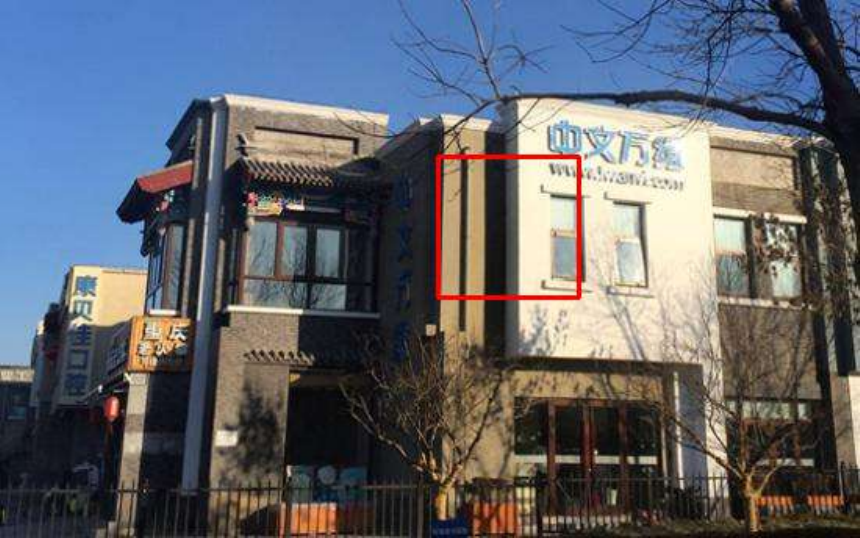} 
          \par\vspace{1mm}
          \includegraphics[width=1.0\linewidth]{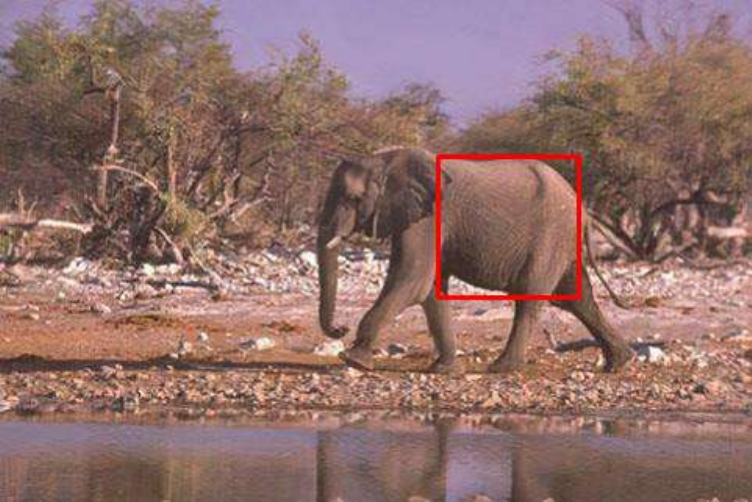} 
        \end{minipage}}
  \caption{Visual result on Rain1400 and RESIDE OTS dataset. Zoom in for best view.}
  \label{fig:setting2}
\end{figure*}

Fig. \ref{fig:mani} depict more restoration level modulation result for different weathers, the results from top to bottom are raindrop, snow, and rain+fog. Fig. \ref{fig:compare} present more visual examples to show the advantage of MQRL over MRL, that model training with MQRL produce clearer and sharper images. Fig. \ref{fig:exp_real_rd}, Fig. \ref{fig:exp_real_rs} and Fig. \ref{fig:exp_real_sw} demonstrate more real-world results on different weathers for raindrop, rain, and snow, as the figures shown, UtilityIR can remove more degradation with less artifacts for diverse weather conditions, and outperform comparison state-of-the-art methods. Fig. \ref{fig:setting2} demonstrate the visual result of \textbf{Setting 2} on rain streak and haze degradation removal. For dehazing, i.e. top row of Fig. \ref{fig:setting2}, UtilityIR can remove more degradation and obtain clearer result compared with other models. For de-rain streak, i.e. bottom row of Fig. \ref{fig:setting2}, UtilityIR can preserve more detail for the red box area in the image, and remove rain streak successfully, while other methods suffer from over-smooth. 

\end{document}